%% file: main.tex
\documentclass[oneside,a4paper,14pt]{extreport}

\usepackage[T5]{fontenc}
\usepackage[utf8]{inputenc}
\DeclareTextSymbolDefault{\DH}{T1}

\usepackage[
	sorting=nty,
	backend=biber]{biblatex}
\usepackage[unicode]{hyperref} 
\usepackage{dblfloatfix}
\makeatletter
\def\blx@maxline{77}
\makeatother

\usepackage[table,xcdraw]{xcolor}
\usepackage{graphicx}
\graphicspath{ {Images/} }
\usepackage[caption=false]{subfig}
\usepackage{float}
\usepackage[flushleft]{threeparttable}
\usepackage{svg}
\usepackage[bottom]{footmisc}

\usepackage{listings}
\usepackage{color}
\definecolor{codegreen}{rgb}{0,0.6,0}
\definecolor{codegray}{rgb}{0.5,0.5,0.5}
\definecolor{codepurple}{rgb}{0.58,0,0.82}
\definecolor{backcolour}{rgb}{0.95,0.95,0.92}
\lstdefinestyle{mystyle}{
    backgroundcolor=\color{backcolour},   
    commentstyle=\color{codegreen},
    keywordstyle=\color{magenta},
    numberstyle=\tiny\color{codegray},
    stringstyle=\color{codepurple},
    basicstyle=\footnotesize,
    breakatwhitespace=false,         
    breaklines=true,                 
    captionpos=b,                    
    keepspaces=true,                 
    numbers=left,                    
    numbersep=5pt,                  
    showspaces=false,                
    showstringspaces=false,
    showtabs=false,                  
    tabsize=2
}
\usepackage{amsmath}
\usepackage{algorithm}
\usepackage{longtable}
\usepackage[noend]{algpseudocode}
\makeatletter
\def\BState{\State\hskip-\ALG@thistlm}
\makeatother

\usepackage{ltablex}
\usepackage{tabularx}
\usepackage{multirow}
\usepackage{array}
\newcolumntype{L}[1]{>{\raggedright\let\newline\\\arraybackslash\hspace{0pt}}m{#1}}
\newcolumntype{C}[1]{>{\centering\let\newline\\\arraybackslash\hspace{0pt}}m{#1}}
\newcolumntype{R}[1]{>{\raggedleft\let\newline\\\arraybackslash\hspace{0pt}}m{#1}}

\renewcommand{\chaptername}{Chương}

\usepackage{titlesec}
\titleformat{\chapter}
    [display] 
    {\normalfont\bfseries\Large}{\chaptername \ \thechapter}{10pt}{\huge}
\titlespacing*{\chapter}{0pt}{-10pt}{40pt} 

\titleformat{\section}
    {\normalfont\bfseries\large}{\thesection}{1em}{}
 
\titleformat{\subsection}
    {\normalfont\bfseries\normalsize}{\thesubsection}{1em}{}
  
\usepackage{setspace}
\usepackage{indentfirst}

\usepackage[
  top=20mm,
  bottom=10mm,
  left=30mm,
  right=20mm,
  footskip = 15mm,
  includefoot]{geometry}
  
\usepackage{tikz}
\usetikzlibrary{calc}

\begin{document}

\input{Title/title.tex}

\pagenumbering{roman} 


\addcontentsline{toc}{chapter}{Lời cảm ơn}
\include{thanks/thanks}

\addcontentsline{toc}{chapter}{Đề cương chi tiết}
\include{Appendix/decuong}

\addcontentsline{toc}{chapter}{Mục lục}
\tableofcontents
\listoffigures
\listoftables

\addcontentsline{toc}{chapter}{Tóm tắt}
\include{Appendix/tomtat}

\clearpage

\pagenumbering{arabic} 

\include{template/chapter1}

\include{template/chapter2}

\include{template/chapter3}

\include{template/chapter4}

\include{template/chapter5}

\include{template/chapter6}

\include{template/chapter7}

\include{template/chapter8}

\addcontentsline{toc}{chapter}{Danh mục công trình của tác giả}
\include{Appendix/publish}

\addcontentsline{toc}{chapter}{Tài liệu tham khảo}
\printbibheading[title={Tài liệu tham khảo}]

\DeclareNameAlias{sortname}{last-first}
\DeclareNameAlias{default}{last-first}

\printbibliography[heading=subbibliography, title={Tiếng Anh}, notkeyword=Viet] 

\addcontentsline{toc}{chapter}{Phụ lục}
\include{Appendix/appendix1}

\end{document}

%% file: Title/title.tex
\begin{titlepage}










\begin{center}

TRƯỜNG ĐẠI HỌC KHOA HỌC TỰ NHIÊN\\
\textbf{KHOA CÔNG NGHỆ THÔNG TIN}\\[2cm]

{\large \bfseries Dương Anh Kiệt - 18120046\\} 
{\large \bfseries Nguyễn Hoàng Lân - 18120051\\[2cm]}


{ \Large \bfseries TĂNG CƯỜNG CHỨNG THỰC MẶT NGƯỜI VỚI MẠNG NHẬN DIỆN KHUÔN MẶT SỬ DỤNG HÀM LARGE-MARGIN COTANGENT LOSS\\[2cm] }

\large KHÓA LUẬN TỐT NGHIỆP CỬ NHÂN\\
\large CHƯƠNG TRÌNH CỬ NHÂN TÀI NĂNG\\[2cm]

\textbf{GIÁO VIÊN HƯỚNG DẪN}\\
TS. Trương Toàn Thịnh

\begin{tikzpicture}[remember picture, overlay]
  \draw[line width = 2pt] ($(current page.north west) + (2cm,-2cm)$) rectangle ($(current page.south east) + (-1.5cm,2cm)$);
\end{tikzpicture}

\vfill
Tp. Hồ Chí Minh, tháng 08/2022

\end{center}

\end{titlepage}

%% file: thanks/thanks.tex
\chapter*{Lời cảm ơn}
\label{thanks}

Đầu tiên chúng em xin gửi lời cảm ơn chân thành đến trường Đại học Khoa học Tự nhiên đã cho chúng em cơ hội được học tập và làm việc tại trường. Nhờ công lao dạy bảo của những thầy cô trong trường và trong Khoa Công nghệ Thông tin, chúng em mới có thể đi đến giai đoạn cuối của chương trình 4 năm Đại học, ngành Công nghệ Thông tin.

Tiếp đến chúng em xin cảm ơn thầy hướng dẫn Trương Toàn Thịnh ở Khoa Công nghệ Thông tin đã giúp đỡ chúng em trong suốt quá trình thực hiện khóa luận tốt nghiệp. Nhờ vào những lời góp ý cùng với những lời khuyên hữu ích của thầy, chúng em có thêm động lực và sự tỉnh táo để hoàn thành khóa luận một cách tốt nhất.
 
Cuối cùng chúng em xin gửi lời cảm ơn sâu sắc đến thầy chủ nhiệm Trần Minh Triết đã gắn bó với chúng em trong suốt quãng thời gian 4 năm học tại Trường Đại học Khoa học Tự nhiên. Nhờ những bài học của thầy, chúng em đã có thể trang bị đủ các kiến thức chuyên môn cùng một số kĩ năng vô cùng cần thiết. Mặc dù thầy không trực tiếp hướng dẫn chúng em trong khóa luận tốt nghiệp nhưng thầy cũng đã dành cho chúng em một số góp ý trong quá trình thực hiện. Nhờ vào đó, chúng em đã tránh được một số sai sót khá nghiêm trọng, giúp chúng em có thêm kinh nghiệm thiết thực trong quá trình làm nghiên cứu sau này.

Qua khóa luận này, chúng em cảm thấy tự tin với bản thân hơn khi đã hoàn thành một đồ án lớn đầu tiên so với những đồ án đã hoàn thành trước đó. Đối với chúng em, khóa luận này đóng vai trò như một bước đệm để giúp chúng em trở nên gần gũi hơn với những dự án thực tế. Đây chắc chắn sẽ là động lực to lớn giúp chúng em đối diện với những thử thách mới ở môi trường làm việc thực tế trong tương lai.

%% file: template/chapter1.tex
\chapter{Giới thiệu}
\label{Chương 1}

\section{Giới thiệu tổng quan và lý do chọn đề tài}

Bài toán \textbf{Nhận diện khuôn mặt (Face Recognition)} là một trong những thách thức lớn đối với lĩnh vực Thị giác Máy tính (Computer Vision) trong vài thập kỷ vừa qua và lĩnh vực Học Sâu (Deep Learning) trong khoảng thời gian gần đây. Bài toán này có ứng dụng quan trọng ở nhiều lĩnh vực đời sống khác nhau, chẳng hạn như: nhận diện danh tính để tìm kiếm bạn bè trên các trang mạng xã hội, tìm kiếm những nạn nhân trong nạn buôn người, và thậm chí truy lùng những tội phạm đang trốn tránh pháp luật; Trong phạm vi trường học, hệ thống nhận diện khuôn mặt có khả năng điểm danh, theo dõi học sinh trong lớp học và trường học, giúp ngăn chặn những ảnh hưởng tiêu cực đến học sinh; Ngoài ra, hệ thống nhận diện khuôn mặt có tiềm năng trong bài toán nhận diện cảm xúc con người, điều này có ảnh hưởng vô cùng to lớn đối với những người khiếm thị, hệ thống có thể giúp họ hiểu rõ hơn trong các tình huống xã hội đồng thời cảnh báo những nguy cơ có thể xảy đến với họ.

Qua một số ứng dụng đã được liệt kê ở trên, bài toán nhận diện khuôn mặt chắc chắn là một bài toán lớn và có giá trị thực tiễn cao. Nhận thấy được tầm quan trọng của bài toán này, nhóm sinh viên chúng em quyết định nghiên cứu và thực hiện đề tài chứng thực danh tính mặt người trong thiết bị thông minh, đây cũng chính là một bài toán có ứng dụng vô cùng hữu ích trong đời sống sinh hoạt của con người hiện nay. Minh chứng rõ ràng nhất chính là hệ thống chứng thực khuôn mặt Face ID của Apple, một hệ thống được tích hợp vào trong hầu hết các điện thoại iPhone. Nhờ tính tiện dụng cùng với độ chính xác, bảo mật cao nên tính năng chứng thực khuôn mặt luôn được người sử dụng điện thoại thông minh ưa dùng.

\section{Mục tiêu của đề tài}
Mục tiêu tổng quát của đề tài chính là đề xuất và phát triển một hệ thống chứng thực mặt người trong mở khóa điện thoại hoặc các ứng dụng trong điện thoại sử dụng nhận diện khuôn mặt. Hệ thống sẽ bao gồm 4 kiến trúc riêng biệt: 
\begin{itemize}
    \item Phát hiện khuôn mặt.
    \item Nhận diện khuôn mặt.
    \item Chống giả mạo khuôn mặt.
    \item Phân loại mắt đóng mở.
\end{itemize}

\noindent Trong đó chúng em nhận định bài toán nhận diện khuôn mặt là quan trọng nhất, việc xác định danh tính thật sự của người đứng trước màn hình một cách chính xác tuyệt đối là điều mà các hệ thống nhận diện khuôn mặt cần phải đạt được. Đi kèm với sự phát triển của bài toán nhận diện khuôn mặt, bài toán chống giả mạo khuôn mặt cũng dần trở nên phổ biến và không kém phần quan trọng. Trong một số trường hợp, bằng cách in một tấm ảnh của người đó ra và đưa ra trước màn hình, ta dễ dàng qua mặt được hầu hết các hệ thống nhận diện khuôn mặt. Nhóm tập trung vào cải tiến hai kiến trúc quan trọng này như sau:

\begin{enumerate}
    \item Chúng em đề xuất Large Margin Cotangent Loss (LMCot), một hàm loss được sử dụng để tăng cường độ hiệu quả nhận diện trong mô hình nhận diện khuôn mặt. Độ hiệu quả của LMCot vượt qua hàm suy hao thông dụng nhất và được chứng minh qua nhiều thí nghiệm trên một số tập dữ liệu và các cuộc thi có liên quan.
    
    \item Chúng em đề xuất Double Loss, một hàm loss được sử dụng để tăng cường độ hiệu quả xác thực mặt người là giả hoặc thật trong mô hình chống giả mạo khuôn mặt.
\end{enumerate}
 
Mục tiêu của chúng em chính là đề xuất và phát triển hai hàm suy hao mới: LMCot và Double Loss. Sau đó áp dụng chúng vào quy trình chứng thực khuôn mặt. 

Chúng em phát triển một phần mềm ứng dụng được triển khai trên nền tảng Android để minh họa cho hệ thống đã đề xuất. Phần mềm ứng dụng sẽ tập trung biểu diễn các quy trình của một hệ thống chứng thực đã đề xuất một cách cơ bản và đầy đủ nhất.

\pagebreak

\section{Giới thiệu khái quát nội dung khóa luận}
\noindent Nội dung khóa luận được chúng em trình bày lần lượt như sau:
\begin{enumerate}
    \item Tiến hành khảo sát một số công nghệ nhận diện khuôn mặt đã được tích hợp vào điện thoại thông minh hiện nay.
    
    \item Giới thiệu một số công nghệ, framework, IDE đã được sử dụng trong suốt quá trình nghiên cứu.
    
    \item Nhóm đi sâu vào việc phân tích tường minh các chức năng được đề xuất sử dụng cho hệ thống chứng thực. Các quy trình trong hệ thống được đề xuất và được phân tích một cách khái quát. Các ý tưởng về hai hàm suy hao được nghiên cứu và phát triển cũng được phân tích tại chương này.
    
    \item Trình bày chi tiết các mô hình được sử dụng trong hệ thống, bao gồm dữ liệu, kiến trúc. 
    
    \item Trình bày kiến trúc hệ thống chứng thực trên Android và đặc tả các chức năng được cài đặt trong ứng dụng.
    
    \item Thực hiện đánh giá và so sánh một số mô hình đề xuất cùng với các công trình liên quan. Đồng thời kết luận những nghiên cứu, những đóng góp cho đề tài.
    
    \item Cuối cùng, chúng em phân tích hướng phát triển của đề tài trong tương lai.
\end{enumerate}



%% file: template/chapter2.tex
\chapter{Khảo sát công nghệ Nhận diện khuôn mặt tích hợp trong các dòng điện thoại}
\label{Chương 2}
\section{Khảo sát dòng điện thoại iPhone}
Face ID\footnote{https://en.wikipedia.org/wiki/Face\_ID} thuộc dòng điện thoại iPhone là một công nghệ nhận diện khuôn mặt ra mắt trên iPhone X vào năm 2017. Công nghệ này thay thế hệ thống nhận diện vân tay Touch ID của Apple cho các iPhone mới nhất của công ty, bao gồm iPhone 13 mini, 13, Pro và 13 Pro Max và có khả năng nó cũng sẽ được tìm thấy trên các iPhone tương lai.

Face ID sử dụng hệ thống camera TrueDepth, bao gồm cảm biến, camera và máy chiếu dạng chấm (Dot Projector) ở trên cùng của màn hình iPhone để tạo ra một bản đồ 3D chi tiết trên khuôn mặt của người dùng. Mỗi lần người dùng nhìn vào điện thoại của họ, hệ thống này tiến hành kiểm tra xác thực an toàn, cho phép thiết bị được mở khóa hoặc thanh toán được thực hiện nhanh chóng và trực quan nếu hệ thống nhận ra họ. Những khảo sát sau đây được dịch và diễn đạt lại từ bài báo trên tạp chí trực tuyến \textbf{Pocket-lint}\footnote{https://www.pocket-lint.com/phones/news/apple/142207-what-is-apple-face-id-and-how-does-it-work}.

\subsection{Những tính năng của Face ID}
Nhờ vào một số yếu tố về phần cứng cùng với những công nghệ hiện đại như hệ thống camera TrueDepth, chip sinh học, mạng nơ-ron,... Face ID trở thành một trong những công nghệ hiện đại bậc nhất, hỗ trợ các tính năng quan trọng một cách đầy đủ, chi tiết và an toàn.

Khi ta có sự khác biệt đáng kể về ngoại hình của mình, chẳng hạn như cạo râu, Face ID sẽ xác nhận danh tính của ta bằng cách sử dụng mật khẩu trước khi nó cập nhật dữ liệu khuôn mặt của ta.

Face ID còn được thiết kế để hoạt động với mũ, khăn choàng, kính áp tròng và hầu hết kính râm. Vào thời điểm năm 2017 khi hệ thống ra mắt trên iPhone X, nó không hoạt động được khi người dùng đeo khẩu trang. Cho đến năm 2020, khi tình hình đại dịch COVID-19 diễn biến phức tạp, một bản cập nhật phần mềm\footnote{https://www.apple.com/pt/newsroom/2021/04/ios-14-5-offers-unlock-iphone-with-apple-watch-diverse-siri-voices-and-more/} đến với iOS 14.5 cho phép Face ID hoạt động khi ta đeo khẩu trang, nhưng đồng thời phải đeo Apple Watch để mở khóa.

\subsection{Hệ thống camera TrueDepth}
Mỗi lần ta nhìn vào chiếc điện thoại (iPhone X hoặc các phiên bản mới hơn), máy chiếu chấm chiếu hơn 30.000 điểm hồng ngoại vô hình lên khuôn mặt của ta. Flood Illuminator sẽ sử dụng ánh sáng hồng ngoại vô hình (ánh sáng IR) để phát hiện khuôn mặt của ta, ngay cả trong bóng tối. Tiếp theo, camera IR đọc mẫu chấm và nắm được bản đồ không gian bao gồm các đặc điểm khuôn mặt xung quanh mắt, mũi và miệng của ta. Sau đó đưa các đặc điểm đó qua mạng nơ-ron để tạo ra một mô hình toán học cho khuôn mặt.

\subsection{Mạng nơ-ron}
Hệ thống trong điện thoại sẽ kiểm tra hình chụp khuôn mặt của ta so với hình đã thiết lập và lưu trữ trên thiết bị của mình để xem liệu nó có khớp hay không, và nếu đúng, nó sẽ mở khóa điện thoại hoặc cho phép thanh toán (trên Apple Pay).

Tất cả những bước này xảy ra trong thời gian thực và vô hình. Apple cho biết hãng đã làm việc với hàng ngàn người trên khắp thế giới và tiếp nhận hàng tỉ hình ảnh, và cùng với đó, hãng đã phát triển nhiều mạng nơ-ron để hình thành công nghệ Face ID của mình.

\subsection{Bionic Neural Engine}
Để xử lý tất cả dữ liệu cần thiết cho Face ID, thông qua việc học máy, Apple đã phải phát triển nhân nơ-ron (neural engine) A11 Bionic. Con chip này đã được nâng cấp thành nhân nơ-ron A12 Bionic trên các dòng điện thoại iPhone XS, XS Max và XR, mang lại nhiều cải tiến hơn cho công nghệ Face ID, tiếp theo là A13 trên các mẫu iPhone 11, A14 trên các mẫu iPhone 12 và A15 trên các mẫu iPhone 13.

Mô hình chung, các chip này là phần cứng chuyên biệt được xây dựng cho một bộ thuật toán học máy. Chúng có thể xử lý hàng trăm tỷ hoạt động mỗi giây và do đó có thể được sử dụng cho công nghệ nhận diện khuôn mặt trong thời gian thực.

Với tư cách là người dùng, Face ID rất đơn giản và hầu như lúc nào cũng vận hành được. Nó nhanh hơn 30\% so với khi nó mới ra mắt và nhiều góc độ hơn cũng được hỗ trợ, cho phép mở khóa thiết bị khi nó phẳng trên bàn miễn là ta ở gần. Trong khi khi Face ID lần đầu ra mắt, ta phải nhấc iPhone lên và đưa lên trước mặt.

\subsection{Bảo mật}
Khi Face ID được phát hành, Apple cho biết họ đã nỗ lực rất nhiều để đảm bảo Face ID không xảy ra vấn đề bởi những bức ảnh giả mạo. Họ thậm chí còn làm việc với các nhà sản xuất mặt nạ chuyên nghiệp và các nghệ sĩ trang điểm ở Hollywood để đào tạo mạng nơ-ron của họ và nhờ đó bảo vệ Face ID khỏi những nỗ lực nhằm vượt qua lỗ hổng hệ thống. Họ đã đề cập trên trang web của họ rằng không thể sao chép dữ liệu nhận diện khuôn mặt bằng cách sử dụng ảnh kỹ thuật số 2D, mặt nạ hoặc các kỹ thuật tương tự khác.

Dữ liệu khuôn mặt cũng được bảo vệ bởi một gói bảo mật trong chip Bionic, và tất cả các xử lý được thực hiện trên các chip, cho dù đó là A11, A12, A13, A14 hoặc A15. Điều đó có nghĩa là dữ liệu khuôn mặt của ta không được gửi đến máy chủ.

Apple còn tuyên bố rằng Face ID an toàn gấp 20 lần Touch ID. Trong khi cơ hội của ai đó mở khóa iPhone của ta bằng cách sử dụng một dấu vân tay giả là 1 trong 50 nghìn, con số này tăng theo cấp số nhân lên thành 1 trong 1 triệu khi nói đến Face ID.

Là một lớp bảo mật bổ sung, Face ID của Apple đi kèm với một tính năng nhận diện sự chú ý\footnote{https://support.apple.com/en-us/HT208108}, đảm bảo rằng iPhone của ta không thể được mở khóa cho đến khi mắt của ta mở hoàn toàn và ta đang chú ý đến thiết bị này. Do đó, không thể có ai đó sử dụng Face ID để mở khóa iPhone/iPad Pro của ta trong khi ta đang ngủ.

\subsection{Các khảo sát người dùng về Face ID}
Vào một cuộc khảo sát\footnote{https://www.sellcell.com/blog/apple-2021-new-products-survey/} về sản phẩm mới từ Apple năm 2021, khoảng 3040 người dùng Apple đã được khảo sát rằng họ có mong muốn công nghệ Face ID được tích hợp vào các thiết bị MacBook và iMac ở các phiên bản tiếp theo hay không?
\begin{figure}[H]
    \centering
    \includegraphics[width =  0.98\textwidth]{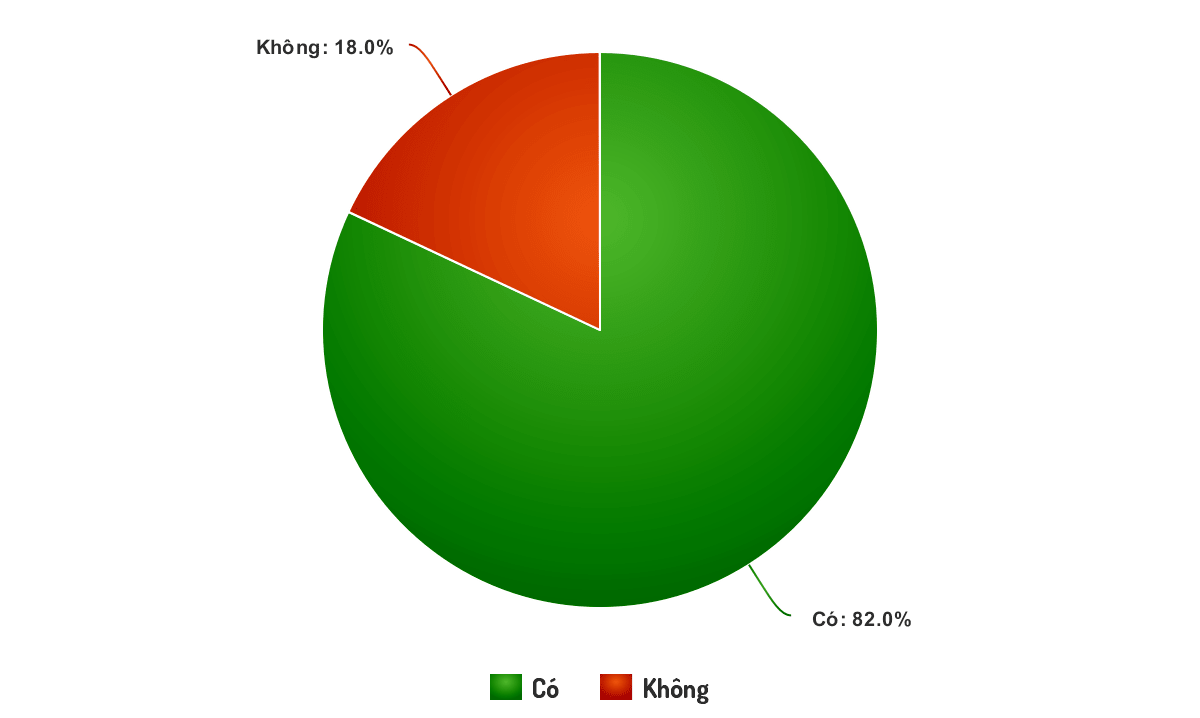}
    \caption{Biểu đồ tròn thể hiện câu trả lời của người dùng vào năm 2021 về việc tích hợp công nghệ Face ID trong tương lai.}
    \label{fig:chart2.1}
\end{figure}
\noindent Theo số liệu Hình \ref{fig:chart2.1}, có đến 82\% người dùng muốn công nghệ Face ID của Apple được tích hợp vào các máy MacBook và iMac trong tương lai vào năm 2021.

Vào một khảo sát trực tuyến được thực hiện từ ngày 2 đến ngày 7 tháng 12 năm 2020 với sự tham gia của hơn 2000 người dùng iPhone, từ 18 tuổi trở lên, tại Hoa Kỳ. Mục tiêu của cuộc khảo sát chính là thu thập phản hồi từ người dùng iPhone về sự thất vọng/vấn đề Face ID trong khi đeo khẩu trang. Ngoài ra, cũng để hỏi liệu họ có muốn xem Touch ID trở lại hay không và nhiều hơn nữa.  Dữ liệu cho 4 biểu đồ tròn bên dưới được trích từ tạp chí trực tuyển SellCell \footnote{https://www.sellcell.com/blog/report-79-of-iphone-users-want-touch-id-to-make-a-comeback-in-the-future-iphones/}.

\begin{figure}[H]
    \centering
    \includegraphics[width = 0.9 \textwidth]{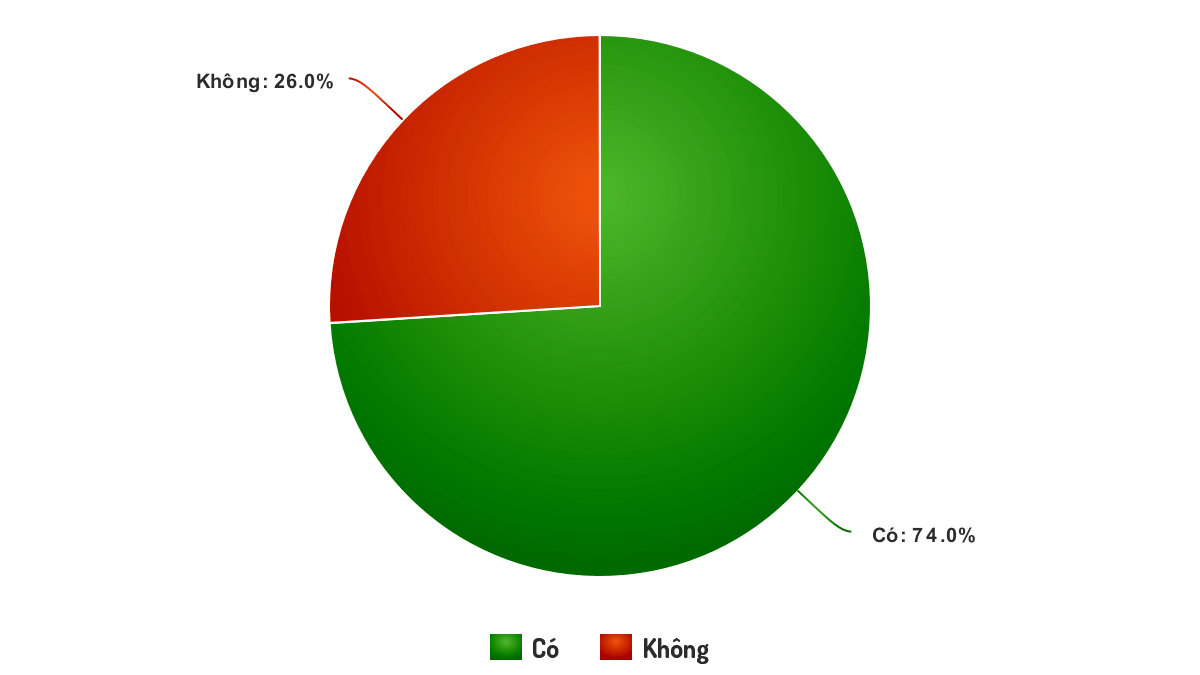}
    \caption{Biểu đồ tròn thể hiện câu trả lời cho người dùng đeo khẩu trang trong mở khóa iPhone.}
    \label{fig:chart2.2}
\end{figure}
\noindent Theo số liệu Hình \ref{fig:chart2.2}, ta thấy có đến 74\% người dùng cảm thấy khó khăn trong việc mở khóa iPhone hoặc cho phép các ứng dụng và thanh toán.

\begin{figure}[H]
    \centering
    \includegraphics[width = 0.9 \textwidth]{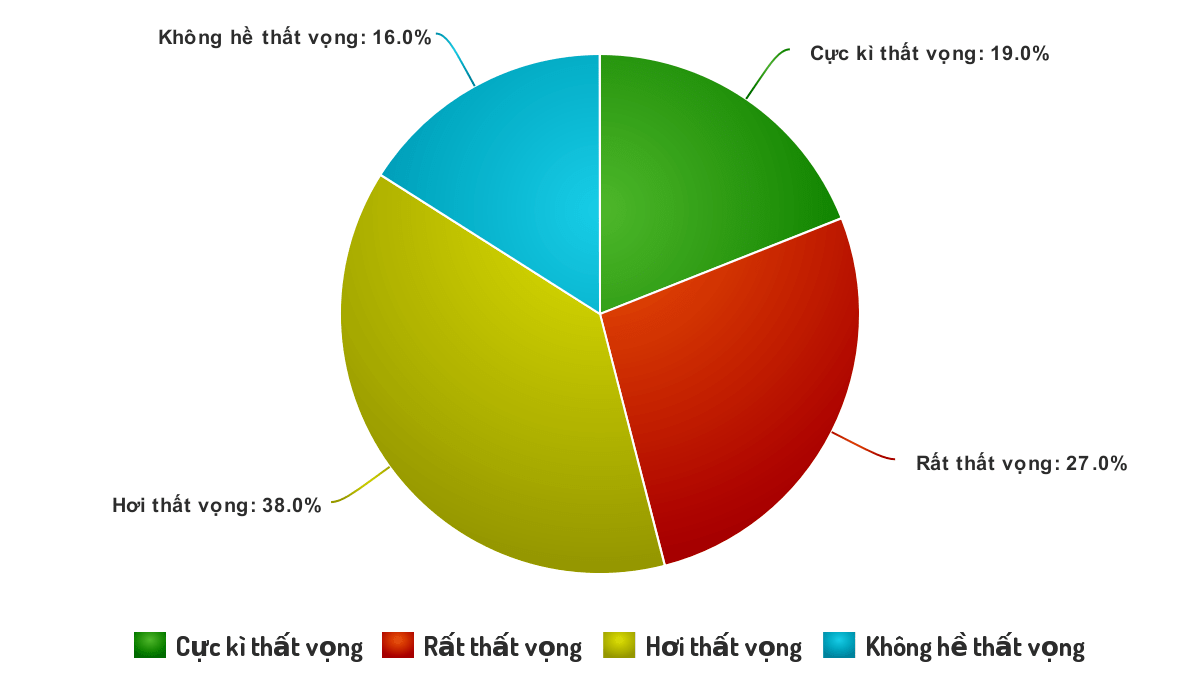}
    \caption{Biểu đồ tròn thể hiện các mức độ thất vọng của người dùng Face ID.}
    \label{fig:chart2.3}
\end{figure}
\noindent Theo số liệu Hình \ref{fig:chart2.3}, có đến tổng cộng 84\% người dùng thật vọng khi Face ID không hoạt động tốt như bình thường khi họ đeo khẩu trang.

\begin{figure}[H]
    \centering
    \includegraphics[width = 0.9 \textwidth]{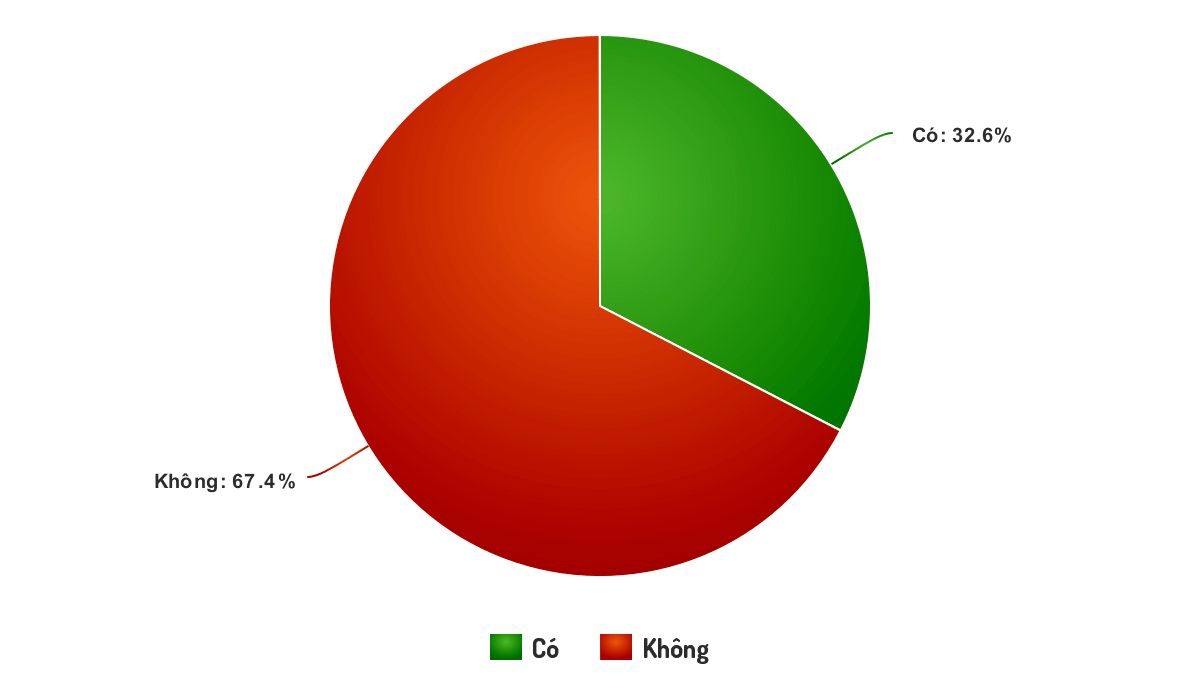}
    \caption{Biểu đồ tròn thể hiện câu trả lời cho việc tháo khẩu trang nơi công cộng của người dùng Face ID.}
    \label{fig:chart2.4}
\end{figure}
\noindent Theo số liệu Hình \ref{fig:chart2.4}, chỉ có 32.6\% người dùng iPhone đã từng tháo khẩu trang ở nơi công cộng để thực hiện nhận diện khuôn mặt bằng Face ID.

\begin{figure}[H]
    \centering
    \includegraphics[width = 0.85 \textwidth]{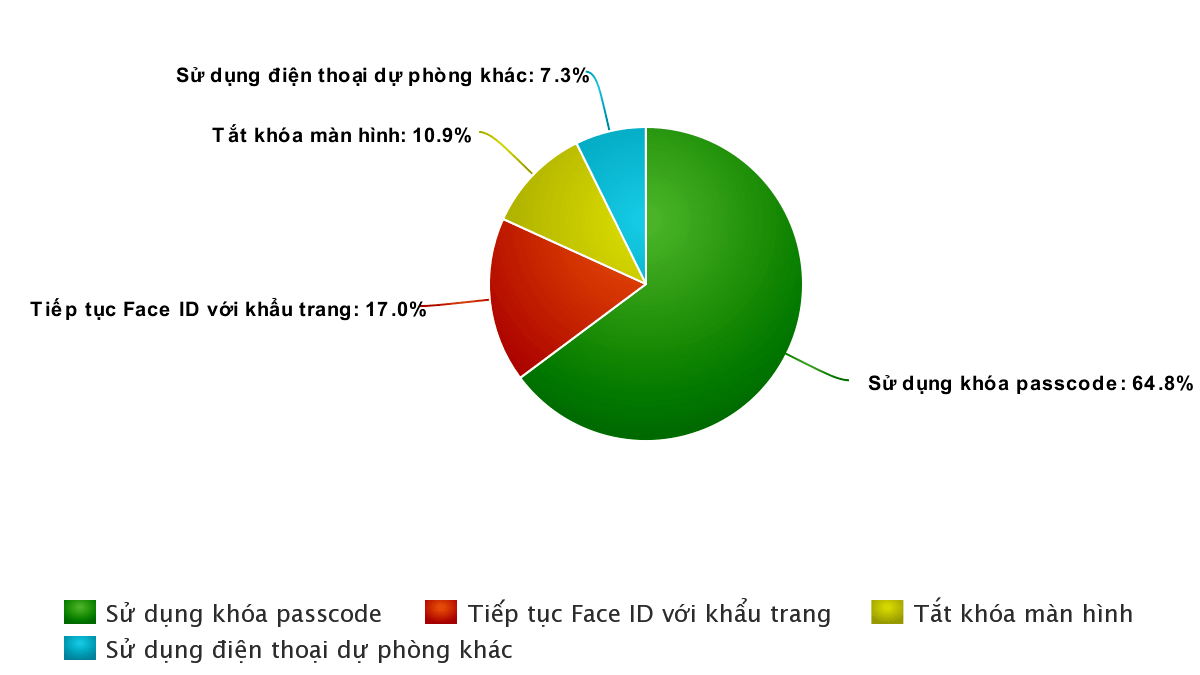}
    \caption{Biểu đồ tròn thể hiện các cách giải quyết tạm thời đối với các trường hợp thực hiện Face ID thất bại.}
    \label{fig:chart2.5}
\end{figure}
\noindent Theo số liệu Hình \ref{fig:chart2.5}, chỉ có 17\% người dùng đã tiếp tục sử dụng Face ID khi đeo khẩu trang. Có đến tổng cộng 83\% người dùng sử dụng những phương pháp khác để mở khóa điện thoại.

\section{Khảo sát dòng điện thoại Google Pixel}
Face Unlock\footnote{https://en.wikipedia.org/wiki/Pixel\_4} thuộc dòng điện thoại Google Pixel là một công nghệ nhận diện khuôn mặt ra mắt trên Google Pixel 4 vào năm 2019. Ở phiên bản này, Google đã loại bỏ tính năng cảm biến vân tay và chỉ dựa vào việc sử dụng các đặc điểm trên khuôn mặt để xác định danh tính. Phụ thuộc vào danh tính của mình, chúng ta mới có thể mở khóa hoặc truy cập vào các ứng dụng. Sau này bởi vì nhiều người dùng cảm thấy không hài lòng với sự thay đổi đó, Google đã phải cập nhật lại các tính năng như mở khóa vân tay và khóa mật khẩu, như trong phiên bản Pixel 6.

Face Unlock cũng sử dụng ánh sáng hồng ngoại để tạo bản đồ 3D về các đặc điểm trên khuôn mặt, tương tự công nghệ Face ID của Apple.  Những khảo sát sau đây được dịch và diễn đạt lại từ hai bài báo trên tạp chí trực tuyến \textbf{Android Central}\footnote{https://www.androidcentral.com/pixel-4-face-unlock-actually-work} \footnote{https://www.androidcentral.com/pixel-4-face-unlock-vs-apples-face-id}.

\subsection{Cách hoạt động}
Về cơ bản, ở mặt trước của Pixel 4, có một chip Radar thu nhỏ được gọi là Soli có thể nhận diện khuôn mặt của người dùng và mở khóa. Soli sẽ quét khuôn mặt của ta theo bất kỳ hướng nào, ngay sau khi ta nhấc máy. Vì vậy, về mặt lý thuyết, điều này sẽ giúp ta giảm bớt quá trình rườm rà khi nâng thiết bị lên hết cỡ, hoặc tạo tư thế theo một cách nhất định và sau đó chờ thiết bị mở khóa.

Google không chỉ sử dụng chip Soli để nhận diện khuôn mặt. Phần trán dày của màn hình Pixel 4 chứa một loạt chip và cảm biến. Vì vậy, bên cạnh Soli, cổng âm thanh và camera selfie, điện thoại còn có cảm biến ánh sáng xung quanh, cảm biến độ gần, máy chiếu chấm, Flood Illuminator và một cặp camera IR. Ba thiết bị cuối giống trong iPhone X, như đã giới thiệu ở trên.

\subsection{Điểm khác nhau giữa Face Unlock và Face ID}
Face Unlock có sử dụng thêm một camera IR và cặp camera IR này được đặt cách xa nhau. Những lợi ích của camera bổ sung này là:
\begin{itemize}
    \item Chụp một hình ảnh ở hai góc độ.
    
    \item Chúng có thể hoạt động trong góc lệch 180 độ, nghĩa là ngay cả khi Pixel 4 được đặt trên bàn, người dùng vẫn có thể mở khóa nó.
    
    \item Trong cả hai trường hợp nói trên, nó sẽ bao phủ nhiều diện tích hơn, ánh xạ nhiều dữ liệu hơn và do đó mở khóa nhanh hơn và đáng tin cậy.
    
    \item Một khả năng khác có thể là người dùng không cần phải căng mặt ra để thu hút sự chú ý của điện thoại.

\end{itemize}

\subsection{Bảo mật}
Ta cũng có thể sử dụng Face Unlock để thanh toán an toàn và xác thực ứng dụng bằng khuôn mặt. Theo Google, dữ liệu khuôn mặt của người dùng được mã hóa và bảo mật an toàn trong điện thoại. Giống như Face ID, tất cả quá trình xử lý sẽ được thực hiện trong thiết bị. Dữ liệu sẽ được lưu trữ an toàn trong chip bảo mật Titan M của Pixel chứ không phải trên đám mây hoặc trên máy chủ của Google.

Khi Face ID ra mắt, nó đã bị ảnh hưởng bởi các vấn đề như nhận diện khuôn mặt không rõ ràng và chậm, đặc biệt là với những người da màu. Chúng ta cần hiểu rằng Face ID và Face Unlock phụ thuộc chủ yếu vào các thuật toán. Chúng trải qua quá trình huấn luyện tiến bộ và theo thời gian, chúng sẽ phát triển, ngay cả Google cũng đề cập đến trong thông báo ra mắt của Soli.

Tại phiên bản Pixel 4 và Pixel 4 XL, Google đã liệt kê những trường hợp người dùng cần lưu ý như\footnote{https://support.google.com/pixelphone/answer/9517039}:
\begin{itemize}
    \item Nhìn vào điện thoại có thể gây mở khóa ngay cả khi người dùng không có ý định đó.
    
    \item Face Unlock có thể không hoạt động nếu người dùng híp mắt, sử dụng chúng dưới ánh nắng gay gắt và giữ diện thoại quá gần mặt.

    \item Điện thoại có thể được mở khóa bởi một người trông rất giống chúng ta, chẳng hạn như anh em sinh đôi.

    \item Người khác cũng có thể mở khóa điện thoại của chúng ta nếu điện thoại được đặt gần mặt mình, ngay cả khi đang nhắm mắt.
\end{itemize}
Trong đó có những trường hợp Google cần phải khắc phục nhằm nâng cao khả năng bảo mật của hệ thống cho người dùng.

Ngạc nhiên thay, tại phiên bản Pixel 6, Google chưa quyết định triển khai Face Unlock trên đó. Kể cả nếu có, nó sẽ thiếu camera hồng ngoại, máy chiếu chấm, cảm biến\footnote{https://9to5google.com/2022/04/27/pixel-6-pro-face-unlock-works/}. Điều đó cho thấy Google có thể sẽ triển khai một số phương pháp tiếp cận khác cho bài toán nhận diện khuôn mặt này trong tương lai.

%% file: template/chapter3.tex
\chapter{Các vấn đề liên quan}
\label{Chapter3}

\noindent Chúng em giới thiệu một số công nghệ, framework và IDE được sử dụng trong đề tài.

\section{Mạng nơ-ron}
Theo định nghĩa từ \textbf{IBM}\footnote{https://www.ibm.com/cloud/learn/neural-networks}, học sâu là một tập hợp con của khái niệm Học máy, chúng sử dụng các hàm toán học để ánh xạ đầu vào với đầu ra. Các hàm này có thể trích xuất thông tin từ dữ liệu, cho phép chúng hình thành mối quan hệ giữa đầu vào và đầu ra.

\begin{figure}[H]
    \centering
    \includegraphics[width=0.68\textwidth]{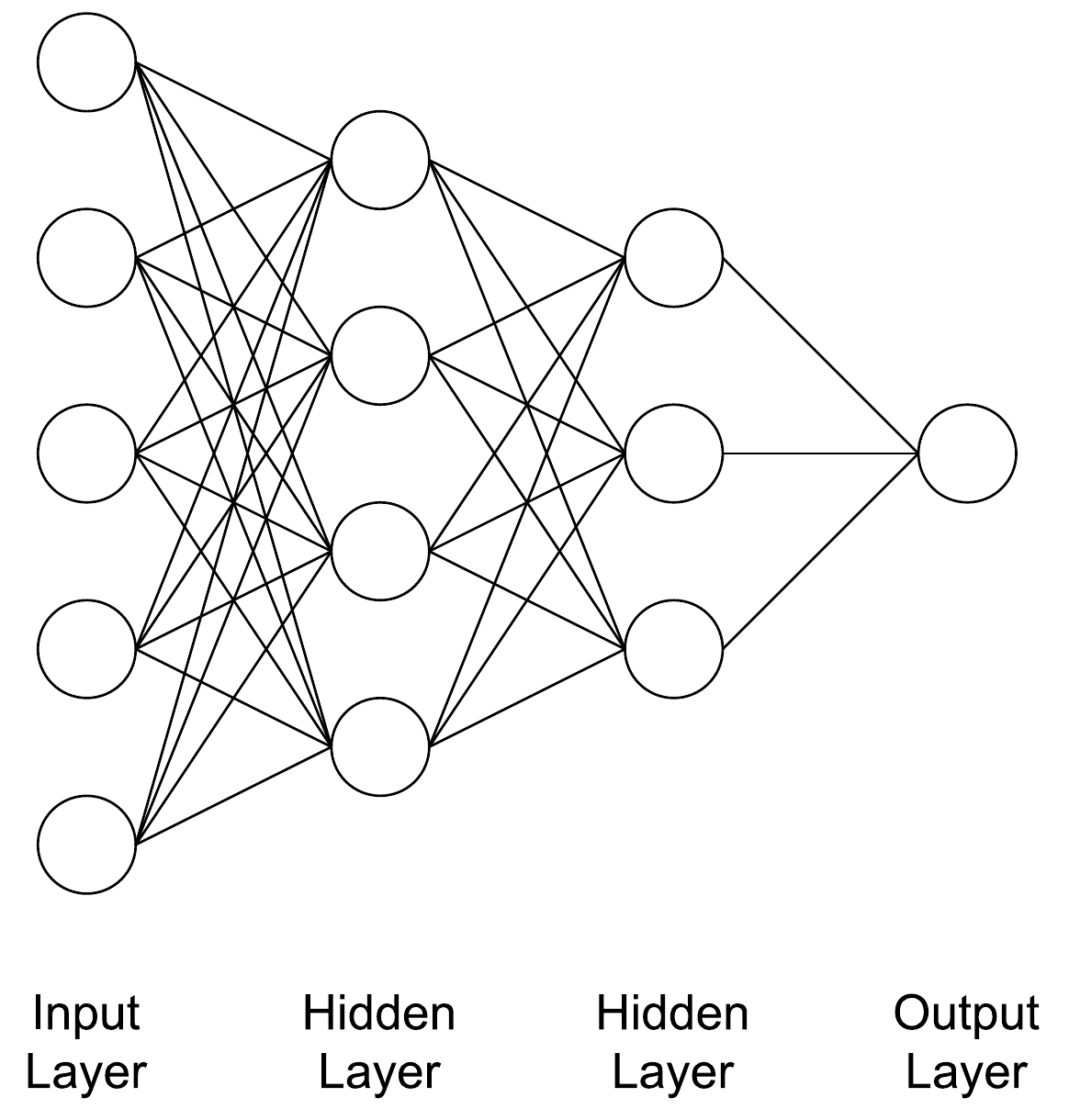}
    \caption{Ví dụ minh họa về một kiến trúc mạng nơ-ron với hai lớp ẩn.}
    \label{fig:chart3.1}
\end{figure}

\noindent Khái niệm về mạng nơ-ron từ đó ra đời, minh họa như Hình \ref{fig:chart3.1}. Mạng nơ-ron được phát triển bởi hai ý tưởng chính:
\begin{itemize}
    \item Bộ não con người đã chứng minh rằng những hành vi thông minh là có thể xảy ra, dựa vào đó, chúng ta tin rằng có thể xây dựng một hệ thống thông minh.
    \item Một quan điểm khác là để hiểu hoạt động của bộ não con người và các nguyên tắc làm nền tảng cho trí thông minh của nó là xây dựng một mô hình toán học có thể làm sáng tỏ các câu hỏi khoa học cơ bản.
\end{itemize}

Về bản chất, mạng nơ-ron cho phép chúng ta học cấu trúc của dữ liệu hoặc thông tin và giúp chúng ta hiểu nó bằng cách huấn luyện trên các tác vụ như hồi quy, phân cụm, phân loại, vân vân.

Sức mạnh của mạng nơ-ron đã được chứng minh ở nhiều tác vụ khác nhau, vượt xa các phương pháp học máy truyền thống. Với quy trình chứng thực khuôn mặt, đây hiện tại là phương pháp tiếp cận hiệu quả nhất mà chúng em nghĩ đến.

\section{Frameworks}
\subsection{TensorFlow}
TensorFlow\footnote{https://www.tensorflow.org/} là một nền tảng mã nguồn mở dành cho học máy được phát triển bởi Google, giúp dễ dàng quá trình thu thập dữ liệu, đào tạo mô hình, phục vụ dự đoán và tinh chỉnh kết quả trong tương lai.

TensorFlow cho phép tạo biểu đồ luồng dữ liệu — cấu trúc và mô tả cách dữ liệu di chuyển qua biểu đồ, hoặc một loạt các nút xử lý. Mỗi nút trong biểu đồ đại diện cho một phép toán và mỗi kết nối giữa các nút là một mảng dữ liệu đa chiều, hoặc tensor.

Ngoài ra, các ứng dụng TensorFlow có thể chạy trên hầu hết mọi thiết bị: máy cục bộ, trên dịch vụ đám mây, thiết bị iOS và Android, CPU, GPU, TPU. Hơn thế, các mô hình kết quả được tạo bởi TensorFlow có thể được triển khai trên hầu hết mọi thiết bị nơi chúng sẽ được sử dụng để thực hiện các dự đoán.

Đây sẽ là framework có vai trò chính trong việc huấn luyện các tác vụ được mô tả trong hệ thống chứng thực của chúng em.

\subsection{Pytorch}
Pytorch\footnote{https://pytorch.org/} cũng là một nền tảng mã nguồn mở dành cho học máy nhưng được phát triển bởi Meta. Framework này đã trở nên ngày càng phổ biến cùng với TensorFlow và được những lập trình viên chuyên ngôn ngữ Python thường ưa chuộng sử dụng nhất. Cả hai hiển nhiên đều có điểm mạnh điểm yếu khác nhau, nhưng về công dụng và mục tiêu, cơ bản cả hai framework đều giống nhau.

\subsection{TensorFlow Lite}
Các mô hình TensorFlow được huấn luyện có thể được triển khai trên thiết bị di động, chẳng hạn như trên nền tảng iOS hoặc Android. TensorFlow Lite\footnote{https://www.tensorflow.org/lite} (TFLite) giúp tối ưu hóa các mô hình TensorFlow để chạy tốt trên các thiết bị như vậy, bằng cách cho phép cân bằng giữa kích thước và độ chính xác của mô hình. Một mô hình nhỏ hơn (khoảng 10 MB, hoặc thậm chí 100 MB) thường kém chính xác hơn, nhưng sự mất mát về độ chính xác là nhỏ và nhiều hơn được bù đắp bởi tốc độ và hiệu quả về năng lượng của mô hình.

Để xây dựng hệ thống trên thiết bị di động nền tảng Android, chúng em sử dụng các mô hình đã được huấn luyện trên các framework TensorFlow, Pytorch và chuyển đổi sang định dạng TFLite (.tflite) - định dạng mô hình học máy nhỏ hơn và hiệu quả hơn trên thiết bị di động\footnote{https://www.tensorflow.org/lite/android}. TensorFlow Lite cung cấp môi trường thực thi được xây dựng sẵn và có thể tùy chỉnh để chạy các mô hình trên Android một cách nhanh chóng và hiệu quả, bao gồm các tùy chọn để tăng tốc phần cứng. Với thư viện TensorFlow Lite Android Support được thiết kế để giúp xử lý đầu vào và đầu ra của các mô hình TFLite, trình biên dịch TFLite sẽ dễ sử dụng hơn.

\section{COLABORATORY}
COLABORATORY\footnote{https://colab.research.google.com/}, hoặc Google Colab là một trình biên dịch dạng trình duyệt cho phép chạy các đoạn code Python trong các ô (cell), phù hợp với công việc huấn luyện mô hình học máy, học sâu, phân tích dữ liệu. Colab không cần yêu cầu cài đặt hay cấu hình đồ họa máy tính, mọi thứ có thể chạy thông qua trình duyệt, ta có thể sử dụng tài nguyên máy tính từ CPU tốc độ cao và cả GPU lẫn TPU đều được cung cấp từ dịch vụ đám mây của Google. 

Google Colab chạy trên một máy ảo đám mây (Cloud Virtual Machine). Hệ điều hành được sử dụng là Ubuntu phiên bản 18.04.3 LTS. Ngoài ra các phiên bản CUDA (nền tảng tính toán song song trên GPU) sẽ là 10.2 đối với bản Google Colab Free hoặc 11.2 đối với bản Google Colab Pro.

Do hạn chế về kinh phí phần cứng nên Google Colab chính là công cụ mà chúng em tin tưởng để thực hiện các tính toán và huấn luyện các mô hình học sâu.

\section{Android Studio}
Android Studio\footnote{https://developer.android.com/studio/intro} là môi trường phát triển tích hợp (IDE) chính thức để phát triển ứng dụng Android, dựa trên IntelliJ IDEA. Dựa trên các trình soạn thảo mã và công cụ phát triển mạnh mẽ của IntelliJ, Android Studio còn cung cấp thêm nhiều tính năng giúp ta nâng cao năng suất khi xây dựng ứng dụng Android.

Lý do chúng em sử dụng Android Studio trong phát triển phần mềm hệ thống chứng thực chính là môi trường phát triển, đặc biệt trong thời gian gần đây Android Studio hỗ trợ nhà phát triển xây dựng, kiểm thử trên các thiết bị thực qua cổng USB và cả mạng WiFi, giúp nâng cao hiệu quả công việc và tiết kiệm về thời gian.

%% file: template/chapter4.tex
\chapter{Mô hình hệ thống chứng thực}
\label{Chapter4}

\section{Nhóm chức năng đề xuất}
\subsection{Phát hiện khuôn mặt}
Đây là bài toán mở đầu luôn được sử dụng trong tất cả các công nghệ liên quan đến khuôn mặt, nhiệm vụ chính là phát hiện khuôn mặt có trong ảnh hoặc frame trong video. Một bức ảnh có thể chứa rất nhiều thông tin nền không liên quan, những thông tin đó thường rất đa dạng do vậy khả năng gây nhiễu rất cao, đặc biệt đối với các dạng bài toán như nhận diện khuôn mặt.

Để xây dựng một hệ thống chứng thực, điều đầu tiên chính là tạo nên một nền móng vững chắc ở công đoạn phát hiện khuôn mặt. Phát hiện khuôn mặt thực chất chính là một nhánh con của một dạng bài toán lớn là phát hiện vật thể (object detection), khi mà chúng được xây dựng và phát triển dựa trên một số lượng dữ liệu lớn về khuôn mặt.

\begin{figure}[H]
    \centering
    \includegraphics[width = 0.8        \textwidth]{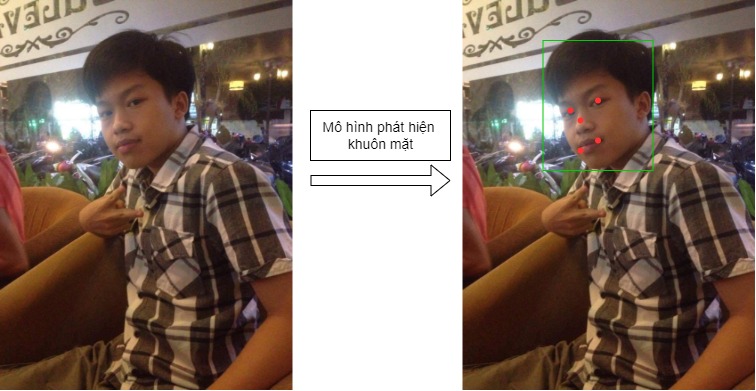}
    \caption{Ví dụ minh họa đầu vào và đầu ra của mô hình phát hiện khuôn mặt.}
    \label{fig:chart4.1}

\end{figure}
Theo Hình \ref{fig:chart4.1}, sau khi đưa một ảnh chân dung qua mô hình phát hiện khuôn mặt. Ngoài thông tin về 4 tọa độ nhằm xác định đối tượng xuất hiện trên bức ảnh, đó chính là cửa sổ (bounding box). Với những phương pháp học sâu hiện đại nhất bấy giờ, chúng ta đã có thể tìm ra được chính xác những thông tin chi tiết của khuôn mặt, chẳng hạn như: miền bao của khuôn mặt, mắt, mũi và miệng. Điều đó giúp ích cho chúng ta rất nhiều trong việc mở rộng bài toán.





\subsection{Nhận diện khuôn mặt}
Đây là bài toán cốt lõi trong một quy trình chứng thực khuôn mặt. Nhiệm vụ chính của bài toán chính là định danh người dùng để xem khuôn mặt của ảnh hiện tại có đúng với thông tin khuôn mặt đã lưu trước đó trong hệ thống hay không. 

Đây cũng chính là dạng bài toán thu nhỏ của Nhận diện đặc điểm (pattern recognition). Đối với con người, chúng ta nhận biết những đặc điểm của một vật bởi thông tin có ý nghĩa nhất định mà chúng mang lại, và được xử lý bởi não bộ. Đối với máy tính, thông tin như bức ảnh sẽ biểu diễn dưới dạng các ma trận pixel để làm đầu vào, sau đó sẽ đưa qua một mô hình học sâu để lấy được những vectơ đặc trưng. Cuối cùng ta sẽ sử dụng kỹ thuật embedding để đưa những vectơ đặc trưng về một không gian vectơ có số chiều nhỏ hơn và cố định, thường là 128 chiều.

Các embedding vectơ hoạt động rất tốt trong việc tìm các giá trị gần nhau (nearest neighbor) trong một cụm, và dựa trên khoảng cách giữa các embedding vectơ mà nó sẽ được phân vào cùng một cụm.

\begin{figure}[H]
    \centering
    \includegraphics[width =                                                                         \textwidth]{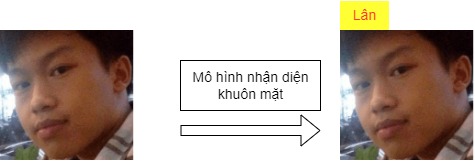}
    \caption{Ví dụ minh họa đầu vào và đầu ra của mô hình nhận diện khuôn mặt.}
    \label{fig:chart4.2}
\end{figure}

Theo Hình \ref{fig:chart4.2}, ta đưa ảnh khuôn mặt được cắt từ ảnh gốc qua mô hình nhận diện khuôn mặt để xác định danh tính của khuôn mặt.

\subsection{Chống giả mạo khuôn mặt}
Đối với một quy trình chứng thực, độ bảo mật vẫn là một trong những ưu tiên hàng đầu và đóng vai trò quan trọng trong thời đại hiện nay, khi mà có nhiều cách thức tấn công ra đời nhằm khai thác các lỗ hổng hệ thống. Ta biết rằng các mô hình học sâu thường nhạy cảm với các trường hợp bất thường, nhất là khi chúng ta làm việc với ảnh hoặc video. Có rất nhiều yếu tố mà con người có thể tác động mạnh mẽ đến chúng, bằng cách trực tiếp hoặc gián tiếp. Do đó, trong những năm gần đây, bài toán chống giả mạo mặt người đã thu hút nhiều sự chú ý, trở thành một đề tài quan trọng khi làm việc cùng với các dạng bài toán về khuôn mặt.

Chống giả mạo khuôn mặt có thể hiểu là bài toán phân loại khuôn mặt ở trước camera là mặt thật hoặc mặt giả. Một số cách thức tấn công giả mạo bao gồm tấn công 2D và 3D:
\begin{itemize}
    \item Tấn công 2D: Sử dụng màn hình điện thoại, laptop có xuất hiện mặt người. Hình mặt người in lên giấy, bìa, hoặc được cắt ra và đặt trước khuôn mặt.
    \item Tấn công 3D: Sử dụng mặt nạ, hình nộm mặt người. Kiểu tấn công này thường kém phổ biến hơn tấn công 2D.
\end{itemize}

\pagebreak

\noindent Một số phương pháp đã được áp dụng để giải quyết bài toán này bao gồm\footnote{https://viblo.asia/p/tong-quan-ve-face-anti-spoofing-bai-toan-chong-gia-mao-khuon-mat-1Je5E6oYKnL}:

\begin{itemize}
    \item Local Binary Pattern (LBP): Phương pháp này tính toán những biểu diễn cục bộ, được xây dựng bằng cách so sánh từng pixel với các pixel lân cận xung quanh của nó.
    
    \item Phát hiện nháy mắt: Phương pháp này được sử dụng để chụp được khoảnh khắc mí mắt sụp xuống. 
    
    \item Mạng học sâu: Phương pháp này coi bài toán chống giả mạo thành bài toán phân loại mặt thật và mặt giả. Sử dụng một mạng CNN để trích xuất đặc trưng với đầu ra là điểm thật giả. Ta so sánh điểm thật giả và điểm ngưỡng để tìm được kết quả.
\end{itemize}

\begin{figure}[H]
    \centering
    \includegraphics[width = 0.8                                                                                     \textwidth]{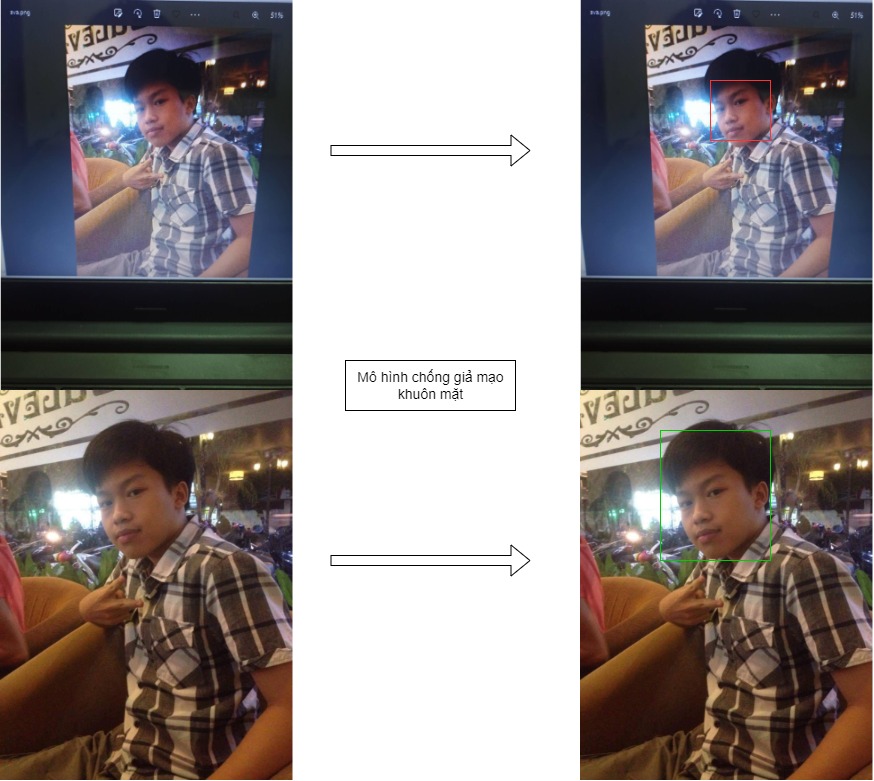}
    \caption{Ví dụ minh họa đầu vào và đầu ra của mô hình chống giả mạo khuôn mặt.}
    \label{fig:chart4.3}
\end{figure}
Theo Hình \ref{fig:chart4.3}, hộp hình chữ nhật màu đỏ tượng trưng cho kết quả ảnh khuôn mặt được mô hình dự đoán là ảnh giả mạo. Ngược lại, hộp hình chữ nhật màu xanh lục tượng trưng cho ảnh thật.

\subsection{Phân loại mắt đóng mở}
Lấy cảm hứng từ hệ thống Face ID của Apple. Họ đã chỉ ra rằng có trường hợp người khác có thể sử dụng điện thoại của ta để mở khóa trong lúc ta đang ngủ. Vì lí do đó, hệ thống còn có cơ chế phân biệt mắt đang đóng hoặc mở. Khi mắt đóng, dù khuôn mặt có trùng khớp với khuôn mặt đã đăng ký trong hệ thống, kết quả vẫn sẽ không cho phép người dùng đi đến tác vụ tiếp theo, như là mở khóa hoặc thanh toán.  

\begin{figure}[H]
    \centering
    \includegraphics[width =  0.6      \textwidth]{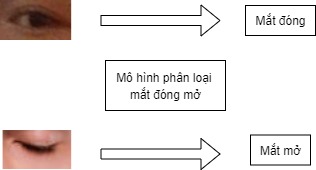}
    \caption{Ví dụ minh họa đầu vào và đầu ra của mô hình phân loại mắt đóng mở.}
    \label{fig:chart4.4}

\end{figure}
Theo Hình \ref{fig:chart4.4}, với thông tin tròng mắt đã thu được từ hệ thống phát hiện khuôn mặt, ta có thể cắt lần lượt hai ảnh, mỗi ảnh chứa một con mắt theo bề ngang và bề rộng nhất định. Sau đó ta đưa qua một mô hình học sâu để huấn luyện, đây là bài toán phân loại với hai nhãn \textit{mở} và \textit{đóng}.

\section{Kiến trúc đề xuất xây dựng hệ thống}
Trong một hệ thống chứng thực khuôn mặt, nhóm nhận định có tổng cộng hai quy trình con cơ bản, lần lượt là: quy trình đăng ký khuôn mặt và quy trình chứng thực khuôn mặt. Từ đó, chúng em giới thiệu một quy trình chứng thực đơn giản, hiệu quả và ít tốn kém chi phí như sau:

\subsection{Quy trình đăng ký khuôn mặt}
\begin{figure}[H]
    \centering
    \includegraphics[width = 0.9              \textwidth]{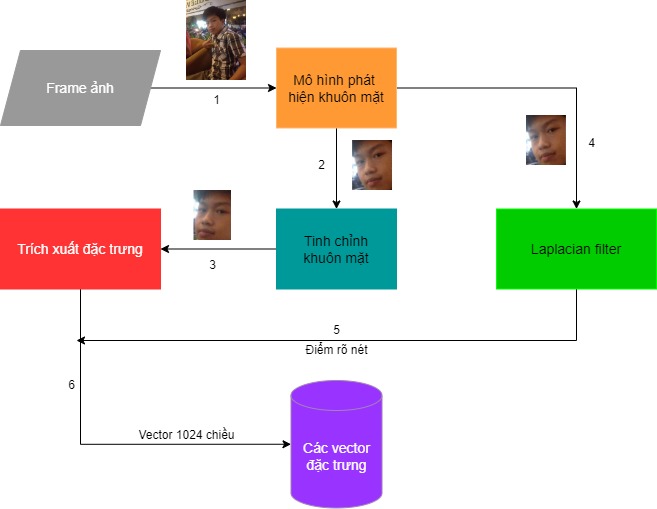}
    \label{fig:chart4.5}
    \caption{Kiến trúc mô tả quy trình đăng ký khuôn mặt.}
\end{figure}
\begin{figure}[H]
    \centering
    \includegraphics[width = \textwidth]{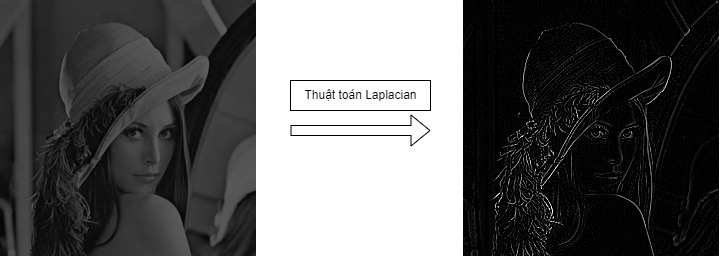}
    \caption{Ví dụ minh họa thuật toán Laplacian.}
    \label{fig:laplacian}
\end{figure}

Quy trình được mô tả chi tiết hơn qua các bước sau:
\begin{itemize}
    \item Frame ảnh lấy từ luồng dữ liệu của camera trước được đưa qua mô hình phát hiện khuôn mặt. Hệ thống sẽ chỉ lấy duy nhất khuôn mặt có kích thước lớn nhất, nếu như có nhiều khuôn mặt cùng xuất hiện trong một bức ảnh. 
    \item Những khuôn mặt được cắt từ những tấm hình đó sẽ được tinh chỉnh lại, trước khi đi qua mô hình nhận diện khuôn mặt để trích xuất các vectơ đặc trưng. Mỗi vectơ đặc trưng đại diện cho thông tin khuôn mặt của mỗi tấm ảnh. 
    \item Trước khi lưu lại dữ liệu vectơ đặc trưng vào cơ sở dữ liệu, hệ thống còn sử dụng thuật toán Laplacian như Hình \ref{fig:laplacian}, để đánh giá độ rõ của ảnh khuôn mặt. Trong trường hợp người dùng chụp những tấm hình mờ, không đủ điều kiện môi trường như ánh sáng, hệ thống sẽ thông báo và không ghi nhận lại các vectơ đặc trưng được trích xuât từ những trường hợp đó.
\end{itemize}

\begin{figure}[H]
    \centering
    \includegraphics[width = 0.3 \textwidth]{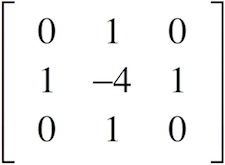}
    \caption{Ví dụ về một kernel Laplacian.}
    \label{fig:laplacian-kernel}
\end{figure}

Thuật toán Laplacian thực chất là phép tích chập giữa một ảnh và một kernel Laplacian, kernel đó là một ma trận 3x3 như Hình \ref{fig:laplacian-kernel}. Nhiệm vụ thông dụng nhất của thuật toán chính là tìm những điểm góc cạnh có trong ảnh, như ở ví dụ trên ta tìm được những điểm đặc trưng trên khuôn mặt với thuật toán. Tại bước này trong quy trình, sau khi tính được những điểm góc cạnh với ảnh đầu vào là ảnh khuôn mặt đã được tinh chỉnh, chúng sẽ được so sánh với một giá trị ngưỡng ngay tại pixel đó (càng gần giá trị 255, điểm càng rõ). Đây được coi là một phương pháp tính độ rõ của ảnh, ảnh sẽ được coi là không hợp lệ nếu độ rõ không đạt.

\subsection{Quy trình chứng thực khuôn mặt}
\begin{figure}[H]
    \centering
    \includegraphics[width = 0.95 \textwidth]{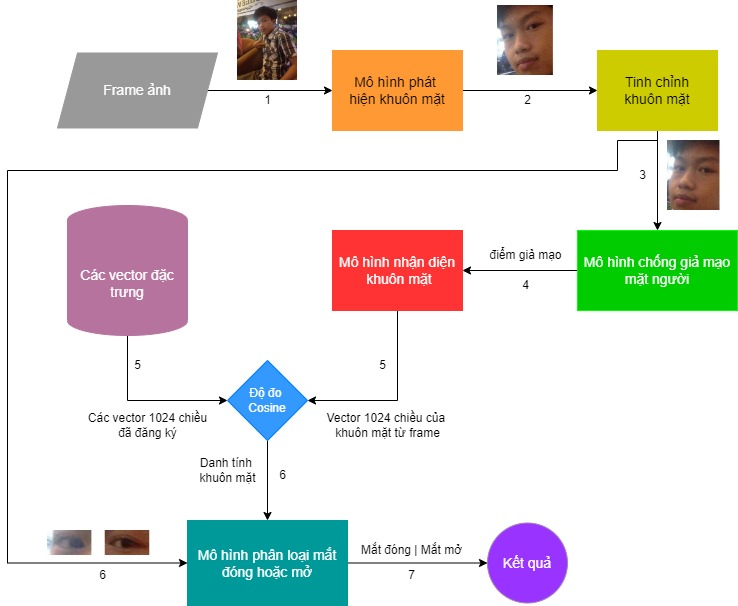}
    \label{fig:chart4.6}
    \caption{Kiến trúc mô tả quy trình chứng thực khuôn mặt.}
\end{figure}
Quy trình được mô tả chi tiết hơn qua các bước sau:
\begin{enumerate}
    \item 
    Frame ảnh lấy từ luồng dữ liệu của camera trước và được đưa qua mô hình phát hiện khuôn mặt, mô hình mà nhóm sử dụng là mạng MTCNN~\cite{DBLP:journals/corr/ZhangZL016}. Đầu vào của mô hình là một ảnh và đầu ra sẽ là cửa sổ (bounding box) chứa khuôn mặt và 5 điểm khuôn mặt lần lượt là: hai mắt, một mũi, hai khóe miệng. Nếu không phát hiện được khuôn mặt thì sẽ tiếp tục thực hiện lại bước phát hiện khuôn mặt ở những frame tiếp theo.
    \item
    Ảnh chứa khuôn mặt sẽ được tinh chỉnh bởi kỹ thuật biến đổi (transformation) như xoay ảnh, canh giữa, scale. Ảnh đã tinh chỉnh sẽ được đưa qua mô hình chống giả mạo khuôn mặt, khi đó ảnh sẽ được quyết định là giả hoặc thật.
    \item
    Nếu ảnh được xác nhận là thật thì ta sẽ tiếp tục đưa ảnh qua mô hình nhận diện khuôn mặt để rút trích vectơ đặc trưng 1024 chiều. Tiếp theo sẽ so sánh lần lượt với các vectơ đặc trưng từ cơ sở dữ liệu - chính là các vectơ được rút trích từ các ảnh khuôn mặt đã đăng ký. Cuối cùng đo độ tương tự cosine giữa từng đôi một các vectơ đặc trưng, nếu độ tương tự cosine lớn hơn ngưỡng cho trước, ta lấy định danh từ độ tương tự cosine lớn nhất.
    \item
    Khi định danh trùng khớp, tiến hành cắt hai ảnh chứa hai mắt từ ảnh khuôn mặt. Sau đó sẽ đưa ảnh chứa mắt qua một mô hình phân loại mắt đóng mở. Nếu mắt được dự đoán là đóng thì ta sẽ quay về giai đoạn đầu tiên. Lí do là vì điều kiện đủ để một hệ thống Face ID xác nhận danh tính chính là đôi mắt. Người dùng phải ở trong trạng thái có nhận thức (mắt mở) và chú ý đến thiết bị (attention-aware).
\end{enumerate}

\section{Nghiên cứu đề xuất các hàm loss}
\subsection{Large Margin Cotangent Loss (LMCot)} \label{sec:LMCot}
\subsubsection{Ý tưởng ban đầu}
Ta bắt đầu với hàm suy hao Softmax truyền thống, một hàm rất phổ biến được sử dụng trong các mô hình phân loại hiện nay.
\begin{equation}
\label{eq:softmax}
    L=-\frac{1}{N}\sum\limits_{i=1}^{N}{\log }\frac{{{e}^{W_{{{y}_{i}}}^{T}{{x}_{i}}+{{b}_{{{y}_{i}}}}}}}{\sum\limits_{j=1}^{n}{{{e}^{W_{j}^{T}{{x}_{i}}+{{b}_{j}}}}}},
\end{equation}
trong đó $L$ là giá trị suy hao dùng để huấn luyện mô hình, $ {{y}_{i}} $ là nhãn của dữ liệu thứ $i$, $n$ là số lượng nhãn của tập dữ liệu, và $N$ là số lượng mẫu của tập dữ liệu. Tuy nhiên, mô hình sử dụng hàm suy hao Softmax chỉ có thể huấn luyện với một số lượng nhãn cố định, và bắt buộc phải huấn luyện lại mô hình nếu chúng ta muốn thêm nhãn mới. Mạng Siamese~\cite{bromley1993signature} là một kiến trúc có thể làm việc với các nhãn bên ngoài tập huấn luyện thay vì tự huấn luyện lại mô hình với mỗi nhãn mới. Đối với mỗi đầu vào, thay vì phân loại chúng, mô hình trích xuất một vectơ đặc trưng, sau đó so sánh nó với những vectơ đã tồn tại trong cơ sở dữ liệu để đưa ra kết quả, như kiến trúc từ Hình \ref{fig:siamese_network}. Như vậy, khi thêm một nhãn mới, chúng ta chỉ cần thêm vectơ của nhãn đó vào cơ sở dữ liệu.

Lấy ví dụ, $ N = 2, b_j = 0, x_0 = \left[0.1, 0.995 \right], x_1 = \left[ 0.2, 0.9798 \right], W_{y_0} = \left[0, 1 \right], W_{y_1} = \left[ 1, 0 \right] $. Theo công thức \ref{eq:softmax} thì $ L =  -0.5\times \left( \frac{{e}^{0.995}}{{e}^{0.995} + {e}^{0.1}} + \frac{{e}^{0.2}}{{e}^{0.9798} + {e}^{0.2}} \right) \simeq 0.3257$.

Tiếp theo, chúng ta chuyển đổi đầu ra logit~\cite{pereyra2017regularizing} thành:
\begin{equation}
    W_{j}^{T}{{x}_{i}}=\left\| {{W}_{j}} \right\|\left\| {{x}_{i}} \right\|\cos \left( {{\theta }_{ji}} \right)
\end{equation}
trong đó ${{W}_{j}}^{T}$ là góc giữa đặc trưng và trọng số $W$ của mô hình. Theo cách chuẩn hóa $l_{2}$~\cite{wang2017normface} ta có  $\left\| {{W}_{j}} \right\|=\left\| {{x}_{i}} \right\|=1$, ${{b}_{j}}=0$, đặt $s$ là hệ số, khi đó công thức ban đầu \ref{eq:softmax} trở thành:
\begin{equation}
\label{eq:softmax_cos}
    L=-\frac{1}{N}\sum\limits_{i=1}^{N}{\log }\frac{{{e}^{s\cos {{\theta }_{{{y}_{i}}}}}}}{{{e}^{s\cos {{\theta }_{{{y}_{i}}}}}}+\sum\limits_{j=1,j\ne {{y}_{i}}}^{n}{{{e}^{s\cos {{\theta }_{ji}}}}}}.
\end{equation}

Theo ví dụ, ta tính được $\theta_{0y_0} \simeq 0.1, \theta_{0y_1} \simeq 1.47, \theta_{1y_0} \simeq 0.2, \theta_{1y_1} = 1.37$.

Dựa vào biến đổi trên công thức \ref{eq:softmax_cos}, một số phương pháp cải tiến đã ra đời. Các phương pháp sử dụng một hoặc nhiều giá trị lề $m$ nhằm tăng góc giữa dữ liệu và nhãn của nó, tạo thách thức cho mô hình để tăng cường khả năng học dữ liệu.\\
\textbf{Liu và các cộng sự} \cite{liu2017sphereface}
\begin{equation}
\label{eq:SphereFace}
    L_{SphereFace}=-\frac{1}{N}\sum\limits_{i=1}^{N}{\log }\frac{{{e}^{s\cos \left( m*{{\theta }_{{{y}_{i}}}} \right)}}}{{{e}^{s\cos \left( m*{{\theta }_{{{y}_{i}}}} \right)}}+\sum\limits_{j=1,j\ne {{y}_{i}}}^{n}{{{e}^{s\cos {{\theta }_{ji}}}}}},
\end{equation}
Lấy $m = 1.1, s = 2$ khi đó theo ví dụ $L_{SphereFace} \simeq 0.0336 + 0.4302 \simeq 0.4638 $
\textbf{Wang và các cộng sự} \cite{wang2018cosface}
\begin{equation}
\label{eq:CosFace}
    L_{CosFace}=-\frac{1}{N}\sum\limits_{i=1}^{N}{\log }\frac{{{e}^{s\left( \cos \left( {{\theta }_{{{y}_{i}}}} \right)-m \right)}}}{{{e}^{s\left( \cos \left( {{\theta }_{{{y}_{i}}}} \right)-m \right)}}+\sum\limits_{j=1,j\ne {{y}_{i}}}^{n}{{{e}^{s\cos {{\theta }_{ji}}}}}},
\end{equation}
Lấy $m = 0.05, s = 2$ khi đó theo ví dụ $L_{CosFace} \simeq 0.0368 + 0.3985 \simeq 0.4353 $
\textbf{Deng và các cộng sự} \cite{deng2019arcface}
\begin{equation}
\label{eq:ArcFace}
    L_{ArcFace}=-\frac{1}{N}\sum\limits_{i=1}^{N}{\log }\frac{{{e}^{s\left( \cos \left( {{\theta }_{{{y}_{i}}}}+m \right) \right)}}}{{{e}^{s\left( \cos \left( {{\theta }_{{{y}_{i}}}}+m \right) \right)}}+\sum\limits_{j=1,j\ne {{y}_{i}}}^{n}{{{e}^{s\cos {{\theta }_{ji}}}}}},
\end{equation}
Lấy $m = 0.05, s = 2$ khi đó theo ví dụ $L_{ArcFace} \simeq 0.034 + 0.3982 \simeq 0.4322 $.
\textbf{Boutros và các cộng sự} \cite{boutros2022elasticface} $L_{ElasticFace-Arc}$, $L_{ElasticFace-Cos}$. Phương pháp được cải tiến dựa trên công thức \ref{eq:ArcFace} và \ref{eq:CosFace}, với giá trị lề $m$ không có định mà là một giá trị ngẫu nhiên của phân phối chuẩn. ElasticFace-Arc là một phương pháp được kế thừa nhiều từ ArcFace~\cite{deng2019arcface} và hàm suy hao này cũng đã đạt state-of-the-art (SOTA) trên một số tập dữ liệu.
\begin{equation}
\label{eq:ElasticFace-Arc}
L_{EArc}=\frac{1}{N} \sum_{i \in N}-\log \frac{e^{s\left(\cos \left(\theta_{y_{i}}+E(m, \sigma)\right)\right)}}{e^{s\left(\cos \left(\theta_{y_{i}}+E(m, \sigma)\right)\right)}+\sum_{j=1, j \neq y_{i}}^{c} e^{s\left(\cos \left(\theta_{j}\right)\right)}},
\end{equation}

\begin{equation}
\label{eq:ElasticFace-Cos}
L_{ECos}=\frac{1}{N} \sum_{i \in N}-\log \frac{e^{s\left( \cos \left(\theta_{y_{i}}\right)-E(m, \sigma)\right)}}{e^{s\left(\cos \left(\theta_{y_{i}}\right)-E(m, \sigma)\right)}+\sum_{j=1, j \neq y_{i}}^{c} e^{s\left(\cos \left(\theta_{j}\right)\right)}},
\end{equation}
trong đó $E\left( \overline{x},\sigma  \right)$ là hàm trả về giá trị ngẫu nhiên từ phân phối chuẩn Gaussian với giá trị trung bình $\overline{x}$ và độ lệch chuẩn $\sigma$.

Ngoài ra, ta còn một số hàm suy hao khác như \textbf{Pairwise confusion loss}~\cite{dubey2018pairwise}, \textbf{Triplet loss}~\cite{schroff2015facenet} với cách hoạt động là dùng mô hình rút trích vectơ đặc trưng sau đó tối ưu hóa khoảng cách giữa các vectơ đặc trưng.

\begin{figure}
    \centering
    \includegraphics[width=0.6\linewidth]{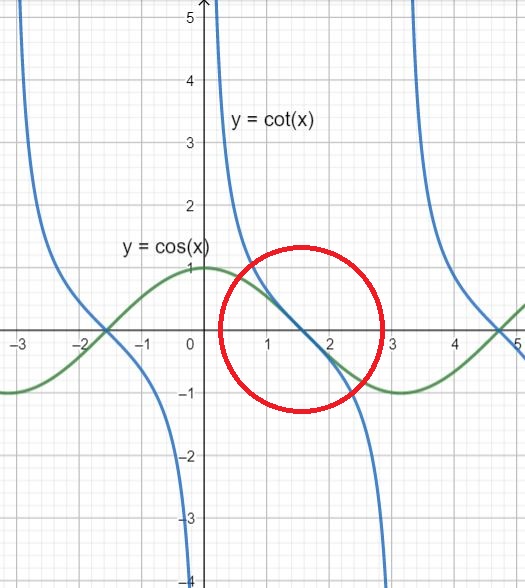}
    \caption{Đồ thị biểu diễn lần lượt hàm $y=\cos \left( x \right)$ và hàm $y=\cot \left( x \right)$.}
    \label{fig:cos_cot}
\end{figure}

Như đã kể trên, có nhiều hàm suy hao để huấn luyện một mạng Siamese. Trong số đó, hàm phổ biến nhất để sử dụng trong các cuộc thi và trong xây dựng ứng dụng là hàm ArcFace theo công thức \ref{eq:ArcFace}, do độ hiệu quả cao và dễ cài đặt. Tuy nhiên, vấn đề của ArcFace là hàm cosin chỉ cho ra các giá trị trong đoạn $\left[ -1,1 \right]$. Vì vậy, hàm cosin không thể minh họa tốt mối quan hệ giữa các góc như hàm cotan, một hàm có miền giá trị $\left( -\infty ,+\infty  \right)$. \\
Ta lấy ví dụ, khi $j\ne y$ góc ${{\theta }_{ij}}$ nhỏ và ${{\theta }_{iy}}$ cũng là một góc nhỏ. Khi đó giá trị cosin của tất cả các góc đều xấp xỉ 1. Khi đó khả năng phân biệt của hàm cosin đối với các góc này là rất kém. Trong các bài toán cần độ chính xác đầu tiên (top-1 accuracy) cao như nhận diện khuôn mặt thì việc phân biệt tốt các góc này là rất cần thiết. Do đó hàm cotan trong trường hợp này làm tốt hơn hàm cosin.\\
Trong trường hợp ${{\theta }_{yj}}$ lớn, lớn ở đây là nhỏ hơn và xấp xỉ $\pi$, do quy ước $arccos$ nên $\theta \in \left[ 0,\pi  \right)$, tức là trọng số ban đầu của mô hình không được tốt. Khi đó, hàm cosin sẽ cho các giá trị xấp xỉ -1 dẫn đến loss khi đó sẽ nhỏ, mô hình sẽ ít tập trung vào dữ liệu huấn luyện dẫn đến quá trình huấn luyện kém hiệu quả. Trong khi cotan sẽ có giá trị âm nhỏ hơn rất nhiều, làm cho loss lớn hơn và quá trình huấn luyện trở nên hiệu quả hơn. Trong các trường góc không quá nhỏ và không quá lớn thì giá trị của hàm cosin và cotan xấp xỉ nhau như Hình \ref{fig:cos_cot}, do đó cotan là một phiên bản nâng cấp hoàn hảo cho hàm cosin trong Large Margin Loss. Kế thừa ArcFace ta thay thế cosin bằng cotan, khi đó LMCot sẽ có dạng:

\begin{equation}
\label{eq:LMCot}
    L_{LMCot}=-\frac{1}{N}\sum\limits_{i=1}^{N}{\log }\frac{{{e}^{s\left( \cot \left( {{\theta }_{yi}}+m \right) \right)}}}{{{e}^{s\left( \cot \left( {{\theta }_{yi}}+m \right) \right)}}+\sum\limits_{j=1,j\ne {{y}_{i}}}^{n}{{{e}^{s\cot {{\theta }_{ji}}}}}}
\end{equation}

trong đó ${{y}_{i}}$, $\theta$, $m$, $s$, $N$ là các tham số như trong công thức \ref{eq:softmax} và \ref{eq:ArcFace}.
Lấy $m = 0.05, s = 2$ khi đó theo ví dụ $L_{LMCot} \simeq {10}^{-6} + 2.0765 \simeq 2.0765 $. 

Như vậy dễ thấy việc dùng hàm cotan sẽ giảm độ suy hao trên các mẫu đã đúng như trường hợp $x_0 = [0.1, 0.995], W_{y_0} = [0, 1]$. Và tăng trên những trường hợp sai nhiều $x_1 = [0.2, 0.9798], W_{y_1} = [1, 0]$, như một bước làm nóng khi mô hình còn hoạt động kém.

\pagebreak

Trong vùng màu đỏ, $\cos \left( x \right)$ có giá trị gần với $\cot \left( x \right)$. Theo quy ước góc giữa hai vectơ đặc trưng $x\in \left[ 0,\pi  \right)$.

Mã giả minh họa các cài đặt hàm LMCot với backbone (phần khung) $B$ và trọng số của mô hình là $W$.
\noindent\rule{\linewidth}{0.4pt}
{\raggedright \textbf{Đầu vào}: dữ liệu $x$, tỉ lệ $s$, giá trị lề $m$, epsilon $\varepsilon $, số lượng nhãn của tập dữ liệu $n$, nhãn đúng của dữ liệu $gt$. \par}
\noindent\rule{\linewidth}{0.4pt}
\begin{enumerate}
\item $f=\left\| B\left( x \right) \right\|$
\item $W=\left\| W \right\|$
\item $\cos \_theta=Wf$
\item $theta=\arccos \left( \cos \_theta \right)$
\item $\cot \_theta=1/\max \left( \tan \left( theta \right),\varepsilon  \right)$
\item $\cot \_theta\_m=1/\tan \left( theta+m \right)$
\item $one\_hot=onehot\left( gt \right)$
\item $\text{logit}=one\_hot*\cot \_theta\_m + \left( 1-one\_hot \right)*\cot \_theta $
\item $\text{logit}=s*\text{logit}$
\end{enumerate}
\noindent\rule{\linewidth}{0.4pt}
\leftline{\textbf{Đầu ra}: Giá trị để phân loại theo lớp $\text{logit}$}
\noindent\rule{\linewidth}{0.4pt}

Trong trường hợp độ phức tạp tính toán $arccos$ là quá cao, chúng ta có thể sử dụng mã giả sau để bỏ qua bước tính toán giá trị góc $\theta$.

\noindent\rule{\linewidth}{0.4pt}
{\raggedright \textbf{Đầu vào}: dữ liệu $x$, tỉ lệ $s$, giá trị lề $m$, epsilon $\varepsilon $, số lượng nhãn của tập dữ liệu $n$, nhãn đúng của dữ liệu $gt$. \par}
\noindent\rule{\linewidth}{0.4pt}
\begin{enumerate}
\item $f=\left\| B\left( x \right) \right\|$
\item $W=\left\| W \right\|$
\item $\cos \_theta=Wf$
\item $\sin \_theta=\max \left( \sqrt{1-\cos \_thet{{a}^{2}}},\varepsilon  \right)$
\item $\cot \_theta=\cos \_theta/\sin \_theta$
\item $\cos \_theta\_m=\cos \_theta*\cos \left( m \right) -\sin \_theta*\sin \left( m \right) $
\item $\sin \_theta\_m=\sin \_theta*\cos \left( m \right)+\cos \_theta*\sin \left( m \right) $ 
\item	$\cot \_theta\_m=\cos \_theta\_m/\sin \_theta\_m$
\item $one\_hot=onehot\left( gt \right)$
\item $\text{logit}=one\_hot*\cot \_theta\_m + \left( 1-one\_hot \right)*\cot \_theta $
\item $\text{logit}=s*\text{logit}$
\end{enumerate}
\noindent\rule{\linewidth}{0.4pt}
\leftline{\textbf{Đầu ra}: Giá trị để phân loại theo lớp $\text{logit}$}
\noindent\rule{\linewidth}{0.4pt}

\subsubsection{Ý tưởng cải tiến}
Trong phần \ref{sec:LMCot} ở trên, ta chỉ xây dựng LMCot dựa trên hàm ArcFace. Như trong bài báo gốc về ArcFace của Deng và các cộng sự \cite{deng2019arcface}, các tác giả có chỉ ra cách kết hợp ý tưởng cách dùng lề $m$ của SphereFace~\cite{liu2017sphereface}, CosFace~\cite{wang2018cosface}, ArcFace~\cite{deng2019arcface} như sau:
\begin{equation}
\label{eq:ArcFace_ensemble}
L=-\frac{1}{N} \sum_{i=1}^{N} \log \frac{e^{s\left(\cos \left(m_{1} \theta_{y_{i}}+m_{2}\right)-m_{3}\right)}}{e^{s\left(\cos \left(m_{1} \theta_{y_{i}}+m_{2}\right)-m_{3}\right)}+\sum_{j=1, j \neq y_{i}}^{n} e^{s \cos \theta_{j}}}.
\end{equation}
Kế thừa đó ta có thể xây dựng một hàm loss ứng dụng cotan như sau:
\begin{equation}
\label{eq:LMCot_ensemble_01}
L=-\frac{1}{N} \sum_{i=1}^{N} \log \frac{e^{s\left(\cot \left(m_{1} \theta_{y_{i}}+m_{2}\right)-m_{3}\right)}}{e^{s\left(\cot \left(m_{1} \theta_{y_{i}}+m_{2}\right)-m_{3}\right)}+\sum_{j=1, j \neq y_{i}}^{n} e^{s \cot \theta_{j}}}.
\end{equation}
Trong đó $m1$, $m2$, $m3$ lần lượt là các giá trị lề khác nhau. Và ${{y}_{i}}$, $\theta$, $m$, $s$, $N$ là các tham số như trong công thức \ref{eq:ArcFace}.

Ta kết hợp thêm ý tưởng của ElasticFace-Arc với hàm cosin thay bằng cotan, ta được:
\begin{equation}
\label{eq:ElasticFace-Arc-Cot}
L_{EArc}=\frac{1}{N} \sum_{i \in N}-\log \frac{e^{o\left(\cot \left(\theta_{y_{i}}+E(m, \sigma)\right)\right)}}{e^{o\left(\cot \left(\theta_{y_{i}}+E(m, \sigma)\right)\right)}+\sum_{j=1, j \neq y_{i}}^{c} e^{a\left(\cot \left(\theta_{j}\right)\right)}}
\end{equation}
với $E\left( \overline{x},\sigma  \right)$  tương tự như công thức \ref{eq:ElasticFace-Arc} và \ref{eq:ElasticFace-Cos}.

Kết hợp công thức \ref{eq:LMCot_ensemble_01} và \ref{eq:ElasticFace-Arc-Cot} ta được:
\begin{equation}
\label{eq:LMCot_ensemble_02}
\begin{split}
    L&=-\frac{1}{N}\sum\limits_{i=1}^{N}{\log }\frac{{{e}^{f \left( \theta \right)}}}{{{e}^{f \left( \theta \right)}}+I}, \\ 
    \quad
    I&=\sum\limits_{j=1,j\ne {{y}_{i}}}^{n}{{{e}^{s\cot {{\theta }_{ji}}}}}, \\
    \quad
    f \left( \theta \right) &= s\left( \cot \left( E\left( {{m}_{1}},{{\sigma }_{1}} \right){{\theta }_{yi}}+E\left( {{m}_{2}},{{\sigma }_{2}} \right) \right)+E\left( {{m}_{3}},{{\sigma }_{3}} \right) \right)
\end{split}
\end{equation}

Ngoài ra, chúng em cũng đề xuất một phương pháp sử dụng đồng thời hai hàm suy hao với cotan và cosin.
\begin{equation}
\label{eq:LMCot_ensemble_03}
    \begin{split}
        L&=\alpha {{L}_{\cot }}+\beta {{L}_{\cos }}, \\
        \quad
        {{L}_{\cot }} &=-\frac{1}{N}\sum\limits_{i=1}^{N}{\log }\frac{{{e}^{f_{cot} \left( \theta \right)}}}{{{e}^{f_{cot} \left( \theta \right)}}+I}, \\ 
        \quad
        {{L}_{\cos }} &=-\frac{1}{N}\sum\limits_{i=1}^{N}{\log }\frac{{{e}^{f_{cos} \left( \theta \right)}}}{{{e}^{f_{cos} \left( \theta \right)}}+I}, \\ 
        \quad
        \text{và } I &=\sum\limits_{j=1,j\ne {{y}_{i}}}^{n}{{{e}^{s\cot {{\theta }_{ji}}}}}, \\
        \quad
        f_{cot} \left( \theta \right) &= s\left( \cot \left( E\left( {{m}_{1}},{{\sigma }_{1}} \right){{\theta }_{yi}}+E\left( {{m}_{2}},{{\sigma }_{2}} \right) \right)+E\left( {{m}_{3}},{{\sigma }_{3}} \right) \right), \\
        \quad
        f_{cos} \left( \theta \right) &= s\left( \cos \left( E\left( {{m}_{1}},{{\sigma }_{1}} \right){{\theta }_{yi}}+E\left( {{m}_{2}},{{\sigma }_{2}} \right) \right)+E\left( {{m}_{3}},{{\sigma }_{3}} \right) \right)
    \end{split}
\end{equation}

Với hai đại lượng $\alpha, \beta$ lần lượt là tỷ lệ phụ thuộc vào $L_{cot}, L_{cos}$. Tùy theo bài toán ta chọn tỷ lệ $\alpha, \beta$ phù hợp. Nhóm em đề xuất với những bài toán cần độ chính xác như nhận diện thì tăng $\alpha$. Và ngược lại với những bài toán như truy vấn thì tăng $\beta$.

Cuối cùng là cách chia quá trình huấn luyện ra thành nhiều giai đoạn và mỗi giai đoạn ta dùng các hàm loss khác nhau để tăng khả năng học dữ liệu của mô hình.

\subsection{Double Loss} \label{sec:DoubleLoss}
Có rất nhiều hàm suy hao phục vụ cho việc huấn luyện một mô hình phân loại nhị phân như hàm cross-entropy thông thường, \textbf{focal loss}~\cite{lin2017focal} để giải quyết vấn đề mất cân bằng dữ liệu, hay \textbf{label smoothing} để làm mịn nhãn tránh hiện tượng overfit. Tuy nhiên, với quy trình chứng thực cần độ chính xác cao thì chưa có hàm suy hao nào thật sự tối ưu. Do đó, nhóm chúng em đề xuất một hàm suy hao \textbf{double loss} có khả năng tách biệt phân phối của hai nhãn về hai phía như Hình \ref{fig:live_spoof_double}. Với việc tách biệt hai phần cần phân loại rõ rệt ta sẽ tránh được việc nhầm lẫn, một yêu cầu rất quan trọng trong các quy trình chứng thực. 

Để hai phần tách biệt nhau cũng có nghĩa là khoảng cách của chúng phải xa nhau. Chính vì vậy hiệu của chúng phải lớn. Để làm được điều đó, mô hình phải nhận vào hai đầu vào, mỗi đầu vào là một nhãn riêng biệt. Ta dùng mô hình để đưa ra điểm số phân loại dựa trên đầu vào, sau đó ta tối ưu hai điểm số đó để chúng cách xa nhau. Double loss được giới thiệu như sau:
\begin{equation}
\label{eq:double_loss}
    L = low\_score-high\_score + 1.0
\end{equation}
Trong đó, low\_score là điểm số phân loại của đầu vào có nhãn 0, và high\_socre là điểm số phân loại của đầu vào có nhãn 1. Do đây là hàm suy hao nên bắt buộc $L>0$, vậy nên ta cộng 1.0 vào để thỏa mãn điều kiện.

\begin{figure}
    \centering
    \includegraphics[width=0.6\linewidth]{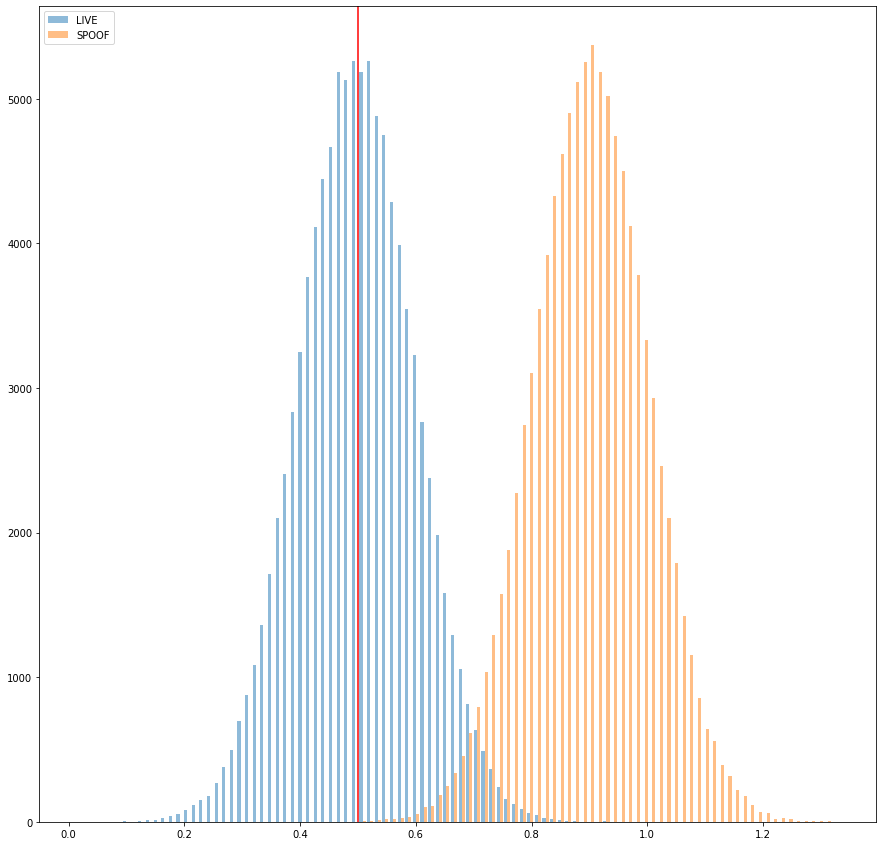}
    \caption{Minh họa trường hợp không tốt của double loss}
    \label{fig:double-only}
\end{figure}

Ta cũng có thể kết hợp huấn luyện đồng thời với các hàm suy hao khác để tăng tốc độ huấn luyện. \textbf{Hàm double loss có khả năng làm cho khoảng cách điểm dự đoán giữa hai nhãn trở nên xa nhau, nhưng khả năng định hướng nhãn nào là 0 và nhãn nào là 1 lại rất kém}. Hàm sẽ nhận hai đầu vào, một chỉ chứa nhãn 0 và một chỉ chứa nhãn 1. Khi đó, theo công thức \ref{eq:double_loss} thì mô hình sẽ làm cho $low\_score$ và $high\_score$ tách biệt có phân phối cách xa nhau. Tuy nhiên, mô hình lại không học được nhãn 0 và 1 cho dữ liệu, dẫn đến có thể bị kẹt lại khi hai phân phối tách biệt nhau như Hình \ref{fig:double-only}. Đường màu đỏ trong hình là ngưỡng phân loại thông thường 0.5, vậy nên trong trường hợp này độ chính xác sẽ rất thấp. Do vậy, huấn luyện đồng thời double loss và một hàm suy hao khác sẽ giúp quá trình huấn luyện trở nên nhanh hơn. Như vậy mô hình cần ba đầu vào: một đầu vào chứa cả hai nhãn, một đầu vào chỉ chứa nhãn 0, và một đầu vào chỉ chứa nhãn một. Hai đầu vào chỉ chứa nhãn 0 và  chỉ chứa nhãn 1 là dùng để huấn luyện với double loss. Đầu vào chứa cả hai nhãn dùng để huấn luyện với hàm suy hao khác như cross-entropy, focal loss, vân vân. Và ta phải thực hiện huấn luyện mô hình với nhiều hàm suy hao cùng lúc.

Ngoài ra, ta cũng có thể tận dụng ý tưởng dùng lề $m$ như các hàm Large Margin Loss như trong phần \ref{sec:LMCot}. Mục đích cũng là để gia tăng khoảng cách giữa hai nhãn. Ý tưởng là, thay vì ở lớp tích chập cuối ta dùng hàm kích hoạt sigmoid thì ta sẽ thêm lề vào đó rồi dùng sigmoid.

\pagebreak

Mã giả của việc dùng lề $m$, với $W$ là trọng số của mô hình, $W_fc(x)$ là đầu ra tại tầng tích chập cuối cùng của mô hình với dữ liệu đầu vào là $x$. 
\noindent\rule{\linewidth}{0.4pt}
{\raggedright \textbf{Đầu vào}: giá trị lề $m$, nhãn đúng của dữ liệu $gt$. \par}
\noindent\rule{\linewidth}{0.4pt}
\begin{enumerate}
\item $score = W_fc \left( x \right)$
\item $score = score + \left( gt - 0.5 \right)*m$
\item $logit = sigmoid(score)$
\end{enumerate}
\noindent\rule{\linewidth}{0.4pt}
\leftline{\textbf{Đầu ra}: Giá trị để phân loại theo lớp $\text{logit}$}
\noindent\rule{\linewidth}{0.4pt}

Như vậy nếu dùng cả hai ý tưởng double loss và lề cho cross-entropy thì huấn luyện mô hình cần bốn dữ liệu đầu vào: một đầu vào chứa cả hai nhãn $inp_1$, một đầu vào chứa nhãn của dữ liệu $inp_1$, một đầu vào chỉ chứa nhãn 0, và một đầu vào chỉ chứa nhãn một. Hai đầu vào chỉ chứa nhãn 0 và  chỉ chứa nhãn 1 là dùng để huấn luyện với double loss. Đầu vào chứa cả hai nhãn và nhãn của nó dùng để huấn luyện với hàm suy hao cross-entropy.

%% file: template/chapter5.tex
\chapter{Nhóm chức năng trong hệ thống}
\label{Chapter5}

\section{Mô hình phát hiện khuôn mặt}
Mô hình phát hiện khuôn mặt sẽ sử dụng mạng MTCNN, mạng này bao gồm ba mạng nơ-ron nhỏ phân biệt, tượng trưng cho ba giai đoạn xử lý khác nhau, lần lượt là: mạng P-Net, mạng R-Net và mạng O-Net.

\subsection{Proposal Network (P-Net)}
\begin{figure}[H]
    \centering
    \includegraphics[width = \textwidth]{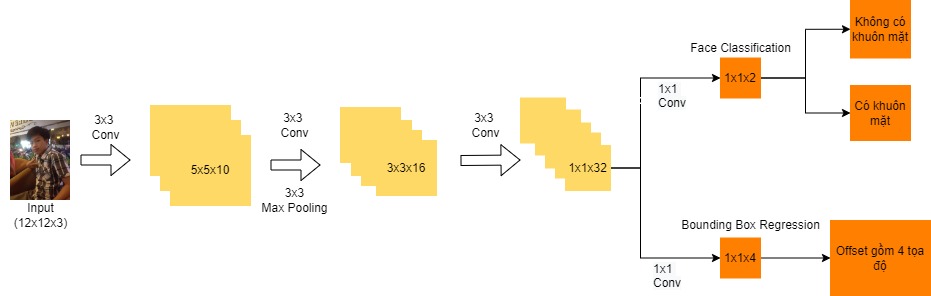}
    \caption{Kiến trúc của mạng P-Net.}
    \label{fig:chart5.1}
\end{figure}

\begin{figure}[H]
    \centering
    \includegraphics[width = 0.5 \textwidth]{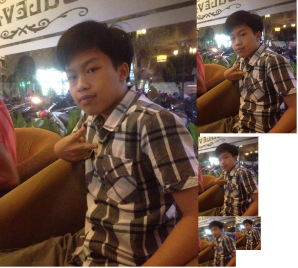}
    \caption{Ví dụ minh họa ảnh được tạo mới và resize về các kích thước khác nhau.}
    \label{fig:chart5.2}
\end{figure}

Trước khi đưa ảnh qua mạng P-Net như Hình \ref{fig:chart5.1}, ta sẽ dùng phương pháp Resize để nhận được các ảnh ở các kích thước khác nhau (như Hình \ref{fig:chart5.2}), điều này sẽ giúp mô hình phát hiện nhiều khuôn mặt ở kích thước đa dạng.

\begin{figure}[H]
    \centering
    \includegraphics[width = 0.7 \textwidth]{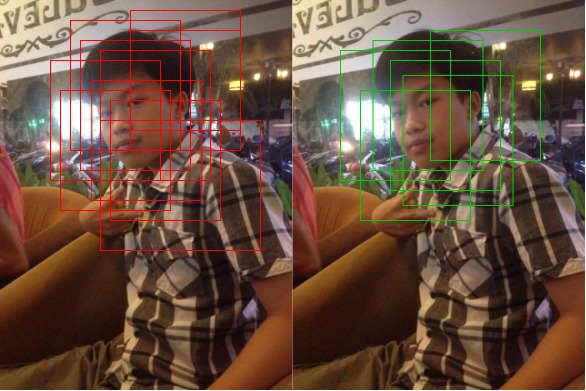}
    \caption{Ví dụ minh họa ảnh kết quả của kỹ thuật NMS tại giai đoạn mạng P-Net.}
    \label{fig:chart5.3}
\end{figure}

Đầu tiên ta sẽ đưa lần lượt các ảnh đã resize qua mạng P-Net. Mạng nơ-ron này khá nông với ba lớp convolution, sau khi ảnh được truyền qua các lớp convolution, các cửa sổ trượt tiềm năng được lựa chọn và tách làm hai nhánh: 
\begin{enumerate}
    \item Bounding Box Regression: Nhánh này giúp tìm được offset (độ lệch) giữa bounding box đang dự đoán và bounding box ground truth (thật).
    \item Face Classification: Nhánh này giúp phân loại được bounding box có Xác suất xuất hiện khuôn mặt có trong bounding box tương ứng (điểm tự tin).
\end{enumerate}
Ngoài ra kỹ thuật Non-Maximum Suppression (NMS) cũng được áp dụng để loại bỏ nhiều bounding box trùng lặp hoặc có điểm tự tin thấp (như Hình \ref{fig:chart5.3}), hơn một ngưỡng cho trước. Cuối cùng sẽ resize các bounding box về cùng một kích thước (24x24).

\subsection{Refine Network (R-Net)}
\begin{figure}[H]
    \centering
    \includegraphics[width = \textwidth]{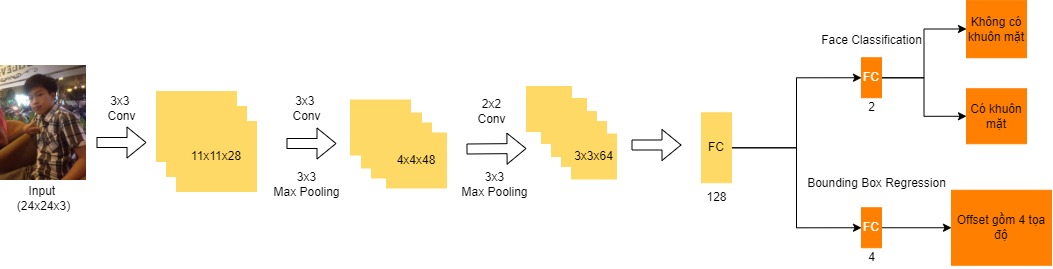}
    \caption{Kiến trúc của mạng R-Net.}
    \label{fig:chart5.4}
\end{figure}

\begin{figure}[H]
    \centering
    \includegraphics[width = 0.7 \textwidth]{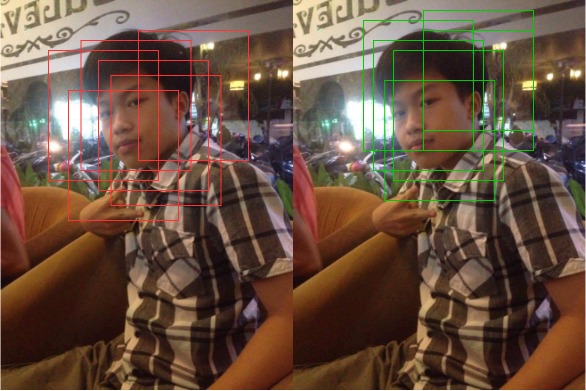}
    \caption{Ví dụ minh họa ảnh kết quả sau khi đưa qua mạng R-Net.}
    \label{fig:RNet}
\end{figure}

Mạng này có công dụng giống mạng P-Net. Ở đây, các bounding box được sử dụng từ kết quả trước sẽ được điều chỉnh về các tọa độ và điểm tự tin. Do đó số lượng bounding box sẽ giảm một phần, như Hình \ref{fig:RNet}.

\begin{figure}[H]
    \centering
    \includegraphics[width = 0.7 \textwidth]{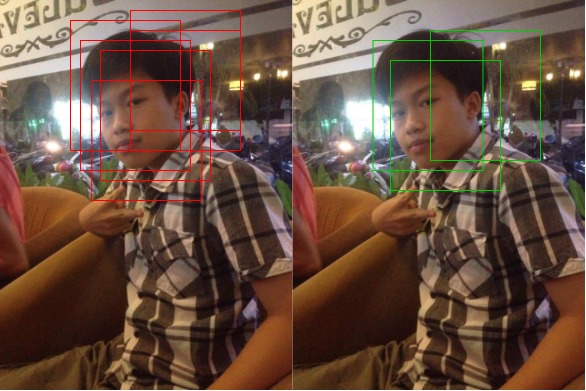}
    \caption{Ví dụ minh họa ảnh kết quả của kỹ thuật NMS tại giai đoạn mạng R-Net.}
    \label{fig:nms-RNet}
\end{figure}

Số lượng cũng sẽ giảm xuống đáng kể vì áp dụng NMS với ngưỡng lớn, như Hình \ref{fig:nms-RNet}.

\subsection{Output Network (O-Net)}
\begin{figure}[H]
    \centering
    \includegraphics[width = \textwidth]{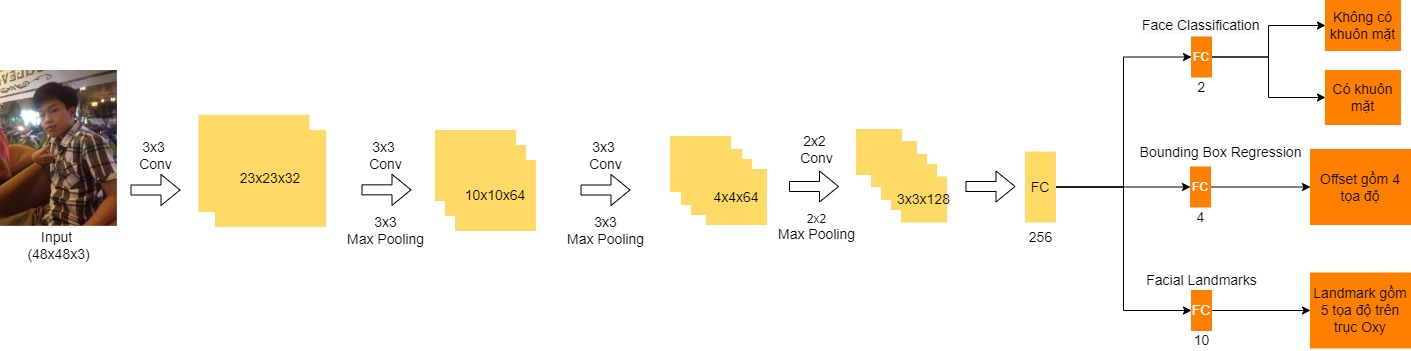}
    \caption{Kiến trúc của mạng O-Net.}
    \label{fig:chart5.5}
\end{figure}
\begin{figure}[H]
    \centering
    \includegraphics[width = 0.7 \textwidth]{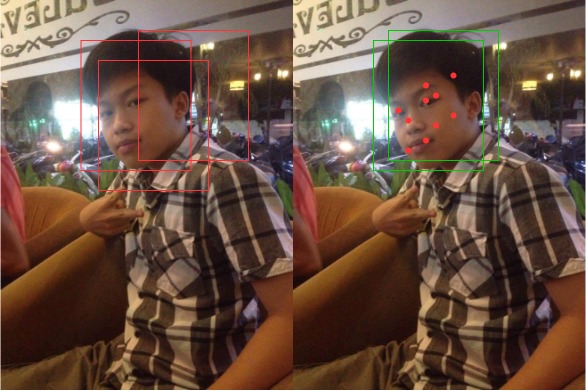}
    \caption{Ví dụ minh họa ảnh kết quả sau khi được đưa qua mạng O-Net.}
    \label{fig:Onet}
\end{figure}

Mạng O-Net lại có công dụng khá giống với mạng R-Net. Ở đây các tọa độ được tinh chỉnh một cách tốt nhất và ta khai thác các thông tin về khuôn mặt một cách chi tiết hơn nữa, đầu ra lúc này sẽ có thêm thông tin về 5 tọa độ landmark trên khuôn mặt, lần lượt là (như Hình \ref{fig:Onet}):
\begin{itemize}
    \item Mí mắt trái, mí mắt phải.
    \item Điểm giữa mũi.
    \item Khóe miệng trái, khóe miệng phải.
\end{itemize}

\begin{figure}[H]
    \centering
    \includegraphics[width = 0.7 \textwidth]{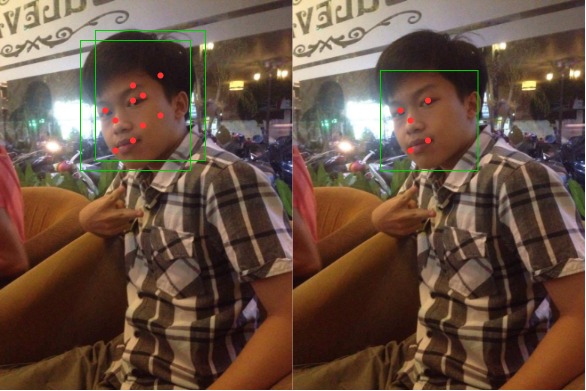}
    \caption{Ví dụ minh họa ảnh kết quả cuối cùng của kỹ thuật NMS tại giai đoạn mạng O-Net.}
    \label{fig:nms-ONet}
\end{figure}

Số lượng bounding box sẽ giảm xuống tối đa vì áp dụng NMS với ngưỡng lớn, như Hình \ref{fig:nms-ONet}.

\section{Kỹ thuật tinh chỉnh khuôn mặt}
\begin{figure}[H]
    \centering
    \includegraphics[width = 0.39 \textwidth]{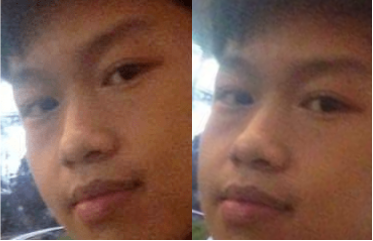}
    \caption{Ví dụ minh họa ảnh đầu vào và ảnh kết quả sau khi áp dụng kỹ thuật tinh chỉnh khuôn mặt.}
    \label{fig:chart5.7}
\end{figure}

Đây là một kỹ thuật luôn được sử dụng trong các bài toán liên quan đến khuôn mặt, thường được áp dụng sau giai đoạn phát hiện khuôn mặt và trước giai đoạn nhận diện khuôn mặt. Đây đóng vai trò như một bước chuẩn hóa dữ liệu đầu vào trong hầu hết các bài toán học máy, học sâu. Việc áp dụng kỹ thuật này nâng cao hiệu quả trong việc học các đặc điểm của khuôn mặt. Tại quá trình này chúng em áp dụng kỹ thuật xoay ảnh, với tiêu chí đoạn thẳng nối giữa tọa độ hai điểm tròng mắt phải nằm ngang với chiều rộng tấm ảnh, như Hình \ref{fig:chart5.7}.

\section{Nhận diện khuôn mặt} \label{sec:Face_rec}
Để nhận diện khuôn mặt, nhóm dùng cách rút trích đặc trưng của khuôn mặt và so sánh các đặc trưng bằng mạng Siamese.

\begin{figure}[htbp]
\label{fig:siamese_network}
    \centering
    \centerline{\includegraphics[width=\linewidth]{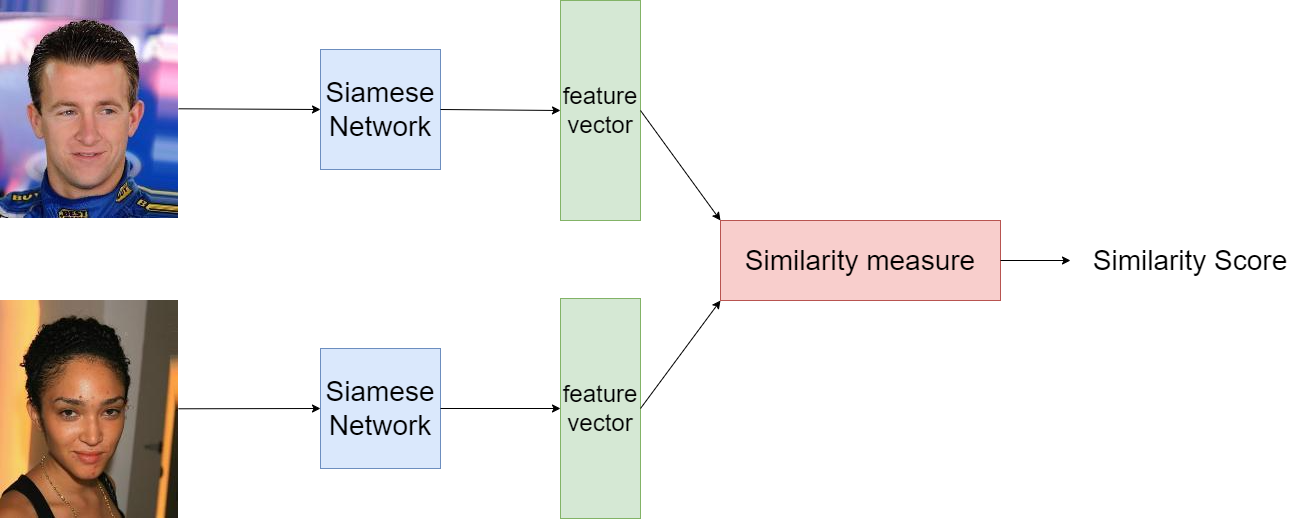}}
    \caption{Hình minh họa cách hoạt động của mạng Siamese.}
\end{figure}
 Hình ảnh được sử dụng được lấy từ tập dữ liệu CelebA \cite{liu2015faceattributes}. Với phép đo tương tự có thể là các hàm khoảng cách như cosine, euclide, manhattan, vân vân.

\begin{figure}[H]
    \centering
    \includegraphics[width=\textwidth]{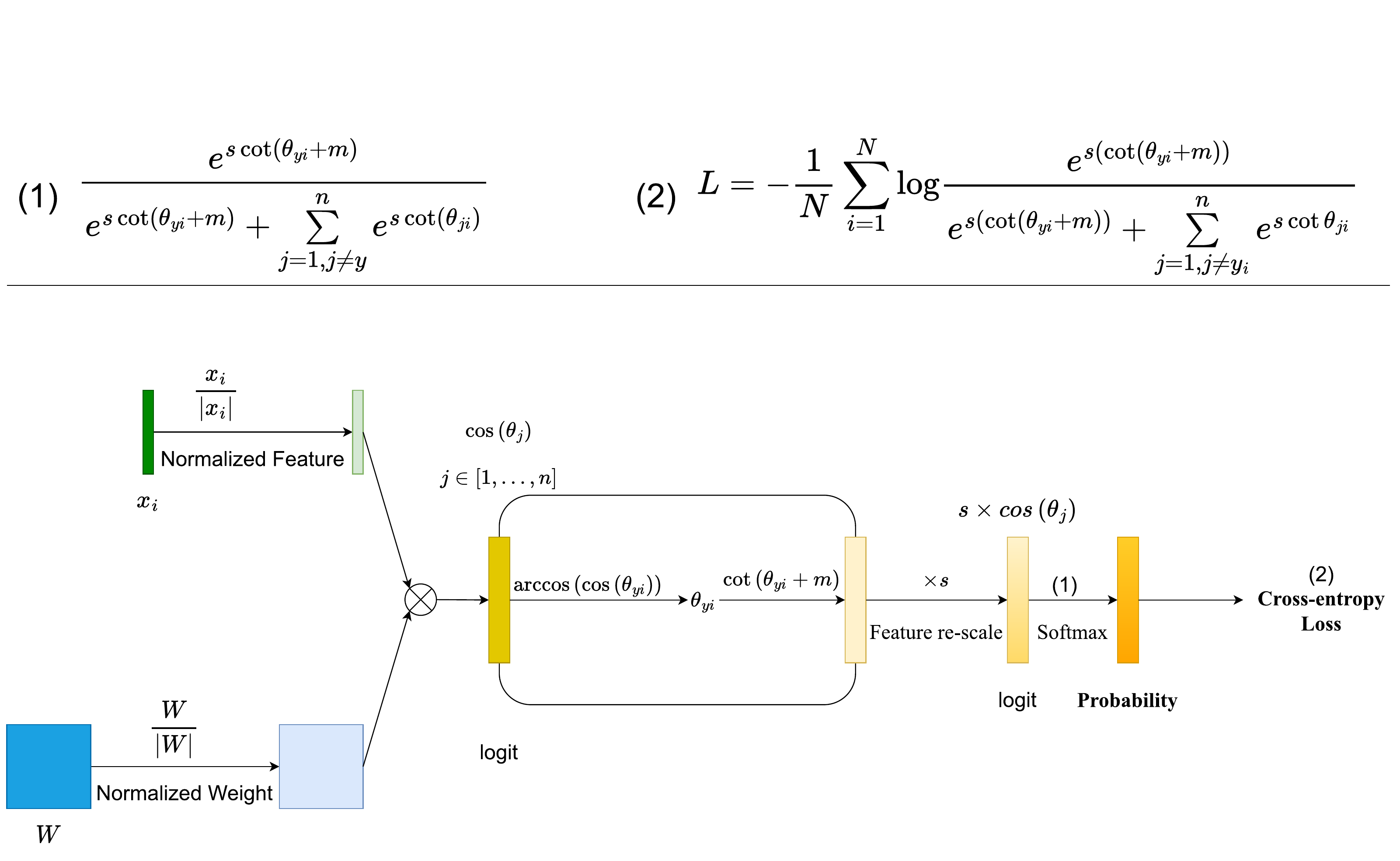}
    \caption{Luồng huấn luyện mô hình bằng cotan loss dựa trên ArcFace\cite{deng2019arcface}.}
    \label{fig:my_label}
\end{figure}
Giá trị $cos \left( \theta_j \right)$ (logit) cho mỗi nhãn được tính từ tích của đặc trưng $x_i$ và trọng số mô hình $W$ cả hai được chuẩn hóa bằng $l2$ \cite{wang2017normface}, công thức là $W_{j}^{T}x_{i}$. Góc giữa đặc trưng $x_i$ và trọng số của mô hình $W$ được tính bởi $arccos \left( \theta_{yi} \right)$. Sau đó ta tính $cot \left( \theta_{yi} + m \right)$, ta nhân kết quả với tỷ lệ $s$. Cuối cùng đưa qua hàm softmax và hàm suy hao cross-entropy.

Trong ứng dụng, chúng em huấn luyện một mô hình để rút trích đặc trưng mặt người bằng hàm suy hao Large Margin Cotangent Loss, cụ thể là hàm \ref{eq:LMCot_ensemble_03}. Tập dữ liệu dùng để huấn luyện là tập CelebA \cite{liu2015faceattributes} với hơn 10 nghìn nhãn và hơn 200 nghìn ảnh. Ngoài ra chúng em còn dùng cả tập CelebA+mask \cite{mare2021realistic} để huấn luyện với ảnh người mang khẩu trang. Với mỗi ảnh trong tập CelebA \cite{liu2015faceattributes} tương ứng ta sẽ có một ảnh người đó mang khẩu trang trong tập  CelebA+mask \cite{mare2021realistic}. Việc huấn luyện với dữ liệu mang khẩu trang sẽ không kém phần quan trọng, vì tình hình dịch bệnh COVID-19 nên mọi người đều phải đeo khẩu trang khi ra ngoài. Ngoài ra, nhóm còn sử dụng các thuật toán chỉnh ảnh để tăng cường dữ liệu ảnh như: lật ảnh trái/phải, xoay ảnh, chỉnh độ sáng và \textbf{điều chỉnh chất lượng hình ảnh}.

\begin{figure}[H]
    \centering
    \includegraphics[width=\linewidth]{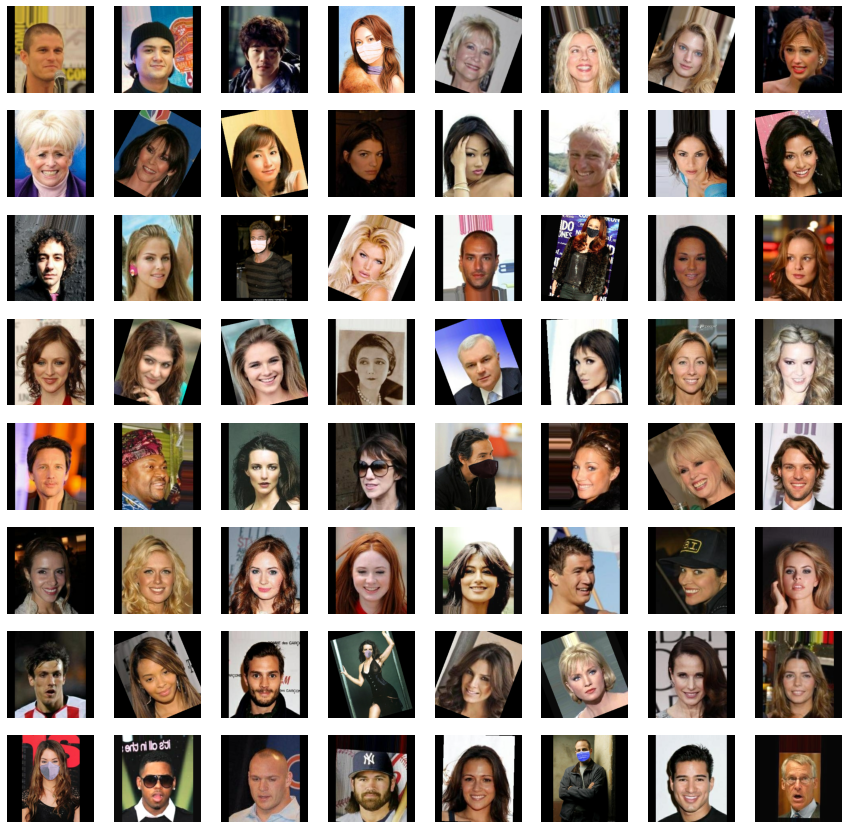}
    \caption{64 hình ảnh ngẫu nhiên được lấy từ tập huấn luyện mô hình nhận diện khuôn mặt.}
    \label{fig:CelebA_random}
\end{figure}
 Các hình được áp dụng các thuật toán chỉnh ảnh để tăng cường dữ liệu ảnh. Các hình ảnh là từ tập CelebA \cite{liu2015faceattributes} và có các hình ảnh người đeo khẩu trang từ tập \cite{mare2021realistic}.
 
\pagebreak

\begin{figure}
\label{fig:Face_reg_graph}
    \centering
    \subfloat[Kiến trúc của mô hình nhận diện khuôn mặt dùng trong huấn luyện. \label{fig:Face_reg_train}]{\includegraphics[width=0.45\linewidth]{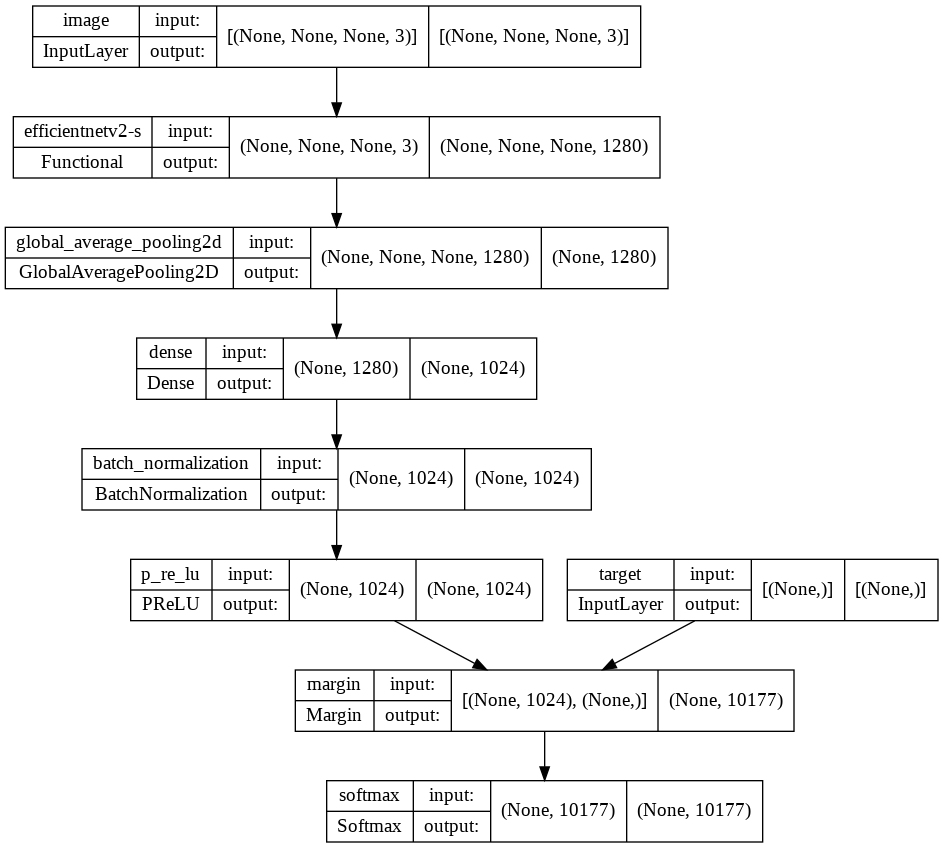}}
    \quad
    \subfloat[Kiến trúc của mô hình nhận diện khuôn mặt dùng trong chạy thực tế. \label{fig:Face_reg_online}]{\includegraphics[width=0.45\linewidth]{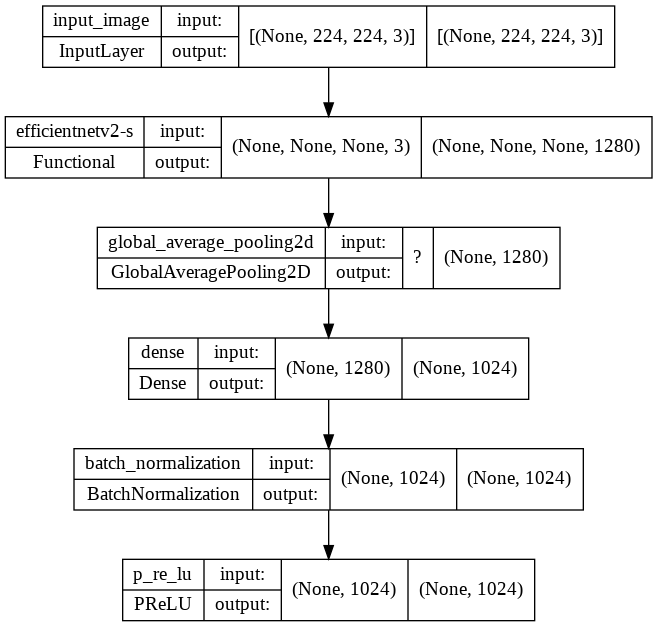}}
    \caption{Kiến trúc của mô hình nhận diện khuôn mặt dùng trong huấn luyện và trong chạy thực tế.}
    \label{fig:face-reg-model}
\end{figure}

Hình \ref{fig:Face_reg_train} là kiến trúc của mô hình dùng để huấn luyện, trong đó lớp margin là lớp LMCot có cách hoạt động như Hình mã giả trong phần \ref{sec:LMCot}. Lớp này cần phải có nhãn của dữ liệu $gt$ nên ở đây ta có thêm một đầu vào chứa nhãn của dữ liệu. Hình \ref{fig:Face_reg_online} là kiến trúc của mô hình dùng để chạy thực tế, ta cắt bỏ lớp margin và lớp softmax vì ta cần là đặc trưng được lấy từ đầu ra của lớp p\_re\_lu. Mô hình có đầu ra là một vectơ 1024 chiều

Mô hình nhóm dùng để huấn luyện có kiến trúc là EfficientNetV2S \cite{tan2021efficientnetv2}. Với kích thước 88MB, mô hình khá thích hợp để triển khai chạy thời gian thực trên các ứng dụng di động, đồng thời mô hình thực thi rất hiệu quả. Tầng cuối cùng của mạng Siamese có đầu ra là một vectơ 1024 chiều. 

\begin{figure}
    \centering
    \includegraphics[width=0.5\linewidth]{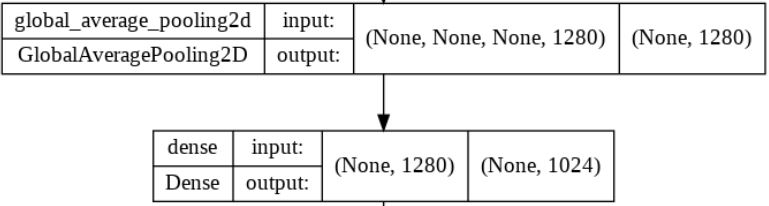}
    \caption{Giải thích kiến trúc mô hình}
    \label{fig:Model_graph_explain}
\end{figure}

Hình \ref{fig:Model_graph_explain} giải là một phần trong hình \ref{fig:Face_reg_train} để giải thích cách đọc kiến trúc. Theo hình \ref{fig:Model_graph_explain} global\_average\_pooling2d là tên của tầng đó trong mô hình. GlovalAveragePoling2D là hàm dùng ở tầng đó. Lần lượt $\left( None, None, None, 1280 \right),\left( None, 1280 \right)$ là chiều đầu vào va đầu ra của tầng đó. Trong đó None là không cố định, có thể thay đổi tùy theo chiều của dữ liệu. Ứng theo chiều mũi tên, đầu ra của global\_average\_pooling2d sẽ là đầu vào cho tầng dense phía dưới. Các tầng khác trong \ref{fig:Face_reg_graph} và \ref{fig:Face_spoof_graph} đọc tương tự.

\section{Chống giả mạo khuôn mặt}
Để chống giả mạo khuôn mặt, nhóm huấn luyện một mô hình phân loại với mục đích phân loại là ảnh thật hay ảnh giả mạo. Mô hình sử dụng kiến trúc MobileNetV2 \cite{sandler2018mobilenetv2} với kích thước chỉ 14MB phù hợp cho các thiết bị điện thoại. Mô hình vẫn hoạt động tốt dù những người trong tập huấn luyện khác với những người trong tập kiểm tra.

Mô hình được huấn luyện bằng double loss kết hợp với margin cross-entropy loss. Vì vậy nên trong quá trình huấn luyện mô hình cần bốn dữ liệu đầu vào: một đầu vào chứa cả hai nhãn $inp_1$, một đầu vào chứa nhãn của dữ liệu $inp_1$, một đầu vào chỉ chứa ảnh thật, và một đầu vào chỉ chứa ảnh giả mạo. Hai đầu vào chỉ chứa nhãn 0 và chỉ chứa nhãn 1 là dùng để huấn luyện với double loss. Đầu vào chứa cả hai nhãn và nhãn của nó dùng để huấn luyện với hàm suy hao margin cross-entropy. Sau đó để dùng mô hình trong thực tế, ta loại bỏ các đầu vào khác chỉ giữ lại đầu vào $inp_1$, đầu ra là đầu ra điểm số phân loại (duy nhất).

Để có thể hoạt động với dữ liệu người mang khẩu trang. Tức là mô hình nhận dạng mặt người có thể làm việc, thì mô hình chống giả mạo phải phân loại được người đang đeo khẩu trang. Tuy nhiên, tập CelebA-Spoof không có ảnh mang khẩu trang. Do đó ta kết hợp CelebA-Spoof và CelebA+mask \cite{mare2021realistic}. Trong đó tập huấn luyện kết hợp với tập huấn luyện, tập kiểm tra kết hợp với tâp kiểm tra. Khi đó dữ liệu từ tập CelebA+mask ta gắn nhãn là ảnh thật. Khi đó, mô hình sẽ nhầm lẫn chỉ cần mang khẩu trang là thật. Vậy nên ta cần tạo các ảnh giả mạo mang khầu trang bằng các hàm  biến đổi ảnh một cách quá nhiều. Như giảm chất lượng ảnh dưới 60\%, tăng giảm độ sáng trên 50\%, cutout ảnh, vân vân.

\begin{figure}[H]
\label{fig:Face_spoof_graph}
    \centering
    \subfloat[Kiến trúc của mô hình chống giả mạo dùng trong huấn luyện. \label{fig:Face_spoof_train}]{\includegraphics[width=1.05\linewidth]{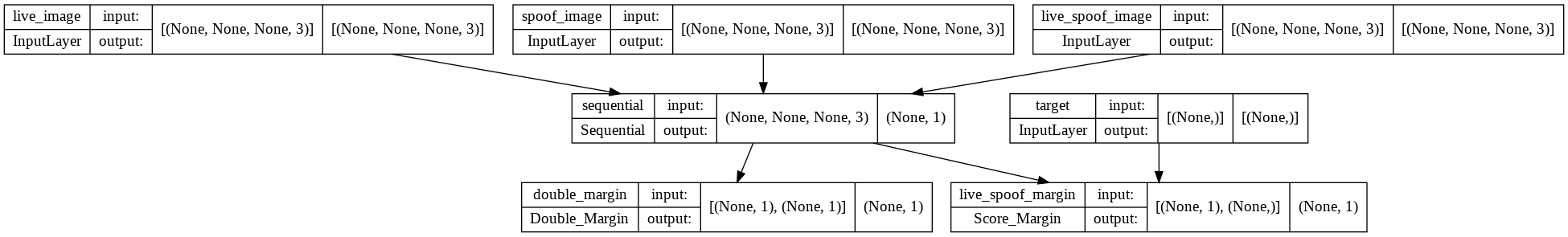}}
    \quad
    \subfloat[Kiến trúc của mô hình chống giả mạo dùng trong chạy thực tế. \label{fig:Face_spoof_online}]{\includegraphics[width=0.6\linewidth]{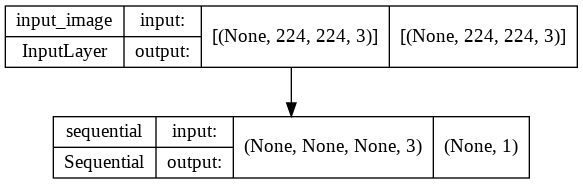}}
    \caption{Kiến trúc của mô hình chống giả mạo khuôn mặt dùng trong huấn luyện và trong chạy thực tế.}
    \label{fig:face-spoof-model}
\end{figure}
Hình \ref{fig:Face_spoof_train} là kiến trúc của mô hình dùng để huấn luyện, trong đó lớp double\_margin là hàm suy hao double loss như \ref{eq:double_loss}. Và live\_spoof\_margin là lớp phục vụ cho hàm suy hao margin cross-entropy, như trong mã giả ở \ref{sec:DoubleLoss}. Lớp này cần phải có nhãn của dữ liệu $gt$ nên ở đây ta có thêm một đầu vào chứa nhãn của dữ liệu. Hình \ref{fig:Face_reg_online} là kiến trúc của mô hình dùng để chạy thực tế, ta cắt bỏ lớp hai lớp double\_margin và lớp live\_spoof\_margin vì ta cần là điểm phân loại tức đầu ra của lớp $sequential$. Trong lớp $sequential$ là các trọng số cần huấn luyện của kiến trúc MobileNetV2, một lớp GlobalAveragePooling2D để giảm chiều dữ liệu và một lớp tích chập với hàm kích hoạt sigmoid để cung cấp đầu ra là điểm số phân loại. 

\section{Phân loại mắt đóng mở}
Để huấn luyện mô hình phân loại mắt đóng mở, chúng em sử dụng bộ dữ liệu MRL Eye \cite{Fusek2018433}. Bộ dữ liệu được tạo ra từ camera trích xuất từ hộp đen trên xe hơi, để phục vụ cho bài toán dự đoán buồn ngủ ở người (Eye Drowsiness Detection). Bộ dữ liệu này bao gồm hai nhãn:
\begin{itemize}
    \item Mắt đóng: 41,966 ảnh.
    \item Mắt mở: 42,957 ảnh.
\end{itemize}

Để đánh giá mô hình một cách tường minh hơn, chúng em đã sử dụng tất cả các ảnh của bộ dữ liệu MRL để huấn luyện và sử dụng bộ dữ liệu CEW \cite{SONG20142825} để đánh giá, bao gồm hai nhãn:
\begin{itemize}
    \item Mắt đóng: 2386 ảnh.
    \item Mắt mở: 2464 ảnh.
\end{itemize}

\begin{figure}[H]
    \centering
    \includegraphics[width = \textwidth]{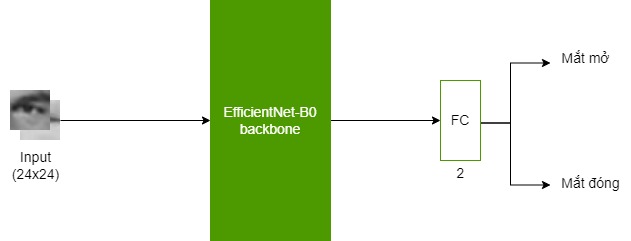}
    \caption{Kiến trúc của mô hình phân loại mắt đóng mở.}
    \label{fig:chart5.8}
\end{figure}

Kiến trúc của mô hình phân loại này khá đơn giản với đầu vào là ảnh 24x24, sau đó được truyền qua phần khung EfficientNetB0. Trước khi qua lớp fully connected để trả ra vectơ 2 chiều đại diện cho hai nhãn \textit{mắt mở} và \textit{mắt đóng}.

Sở dĩ chúng em sử dụng kiến trúc EfficientNetB0 bởi vì chúng khá nhỏ và hoạt động tốt hơn hầu hết các mô hình cùng kích cỡ.

%% file: template/chapter6.tex
\chapter{Hệ thống chứng thực trên điện thoại Android}
\label{Chapter6}
\section{Kiến trúc hệ thống}

\begin{figure}[H]
    \centering
    \includegraphics[width = \textwidth]{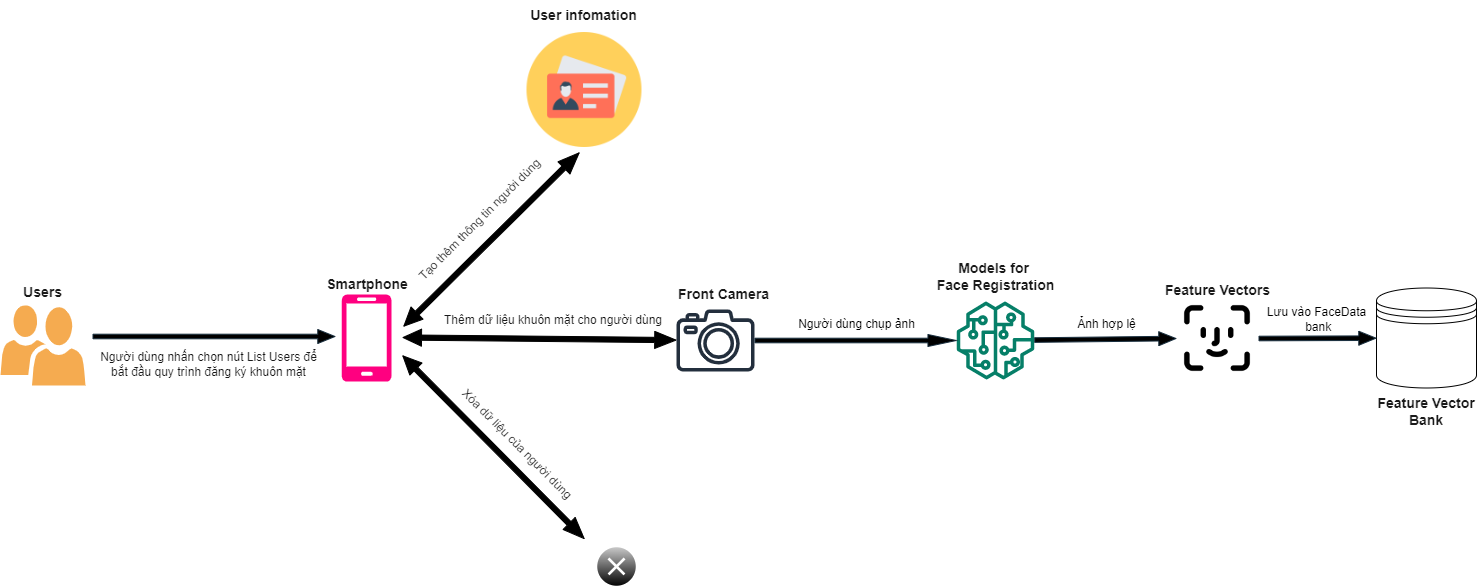}
    \caption{Biểu đồ minh họa luồng người dùng đăng ký khuôn mặt trên hệ thống.}
    \label{fig:chart6.1}
\end{figure}

\begin{figure}[H]
    \centering
    \includegraphics[width = \textwidth]{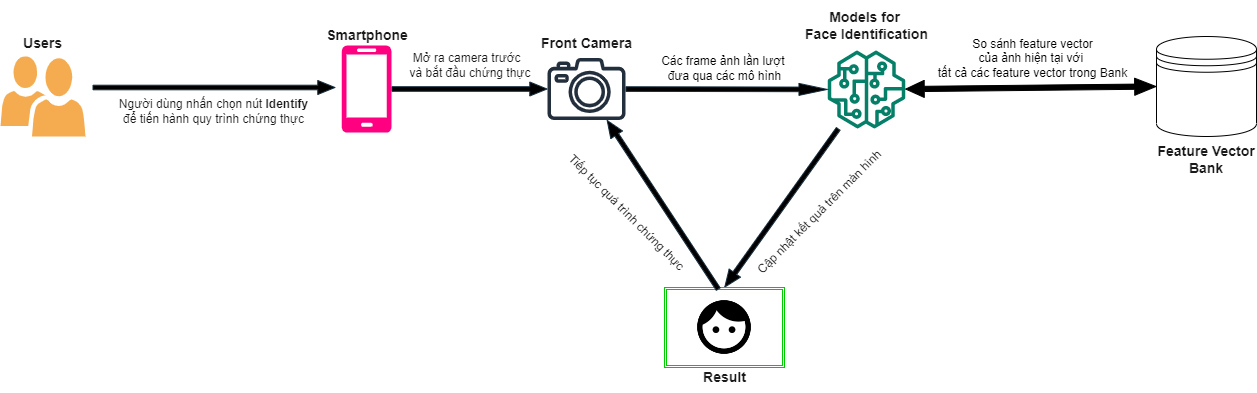}
    \caption{Biểu đồ minh họa luồng người dùng chứng thực khuôn mặt.}
    \label{fig:chart6.2}
\end{figure}

\section{Đặc tả chức năng}
\noindent Chúng em đặc tả chi tiết hai luồng kiến trúc ở trên qua hai bảng sau:

\begin{tabularx}{\linewidth}{|X|X|}
\caption{Đặc tả chức năng cho luồng người dùng đăng ký khuôn mặt}\\

\hline
\textbf{Người dùng}                                                                      & \textbf{Hệ thống} \\ \hline
1. Người dùng tiến hành nhấn nút \textbf{List Users} tại giao diện trang chủ của ứng dụng. & \\ \hline
                                                                                 & 2. Chuyển người dùng đến trang đăng ký khuôn mặt.                                      \\ \hline
3. Người dùng nhấn nút \textbf{Create New} để tạo thêm dữ liệu người đăng ký mới.          &                                                                                            \\ \hline
                                                                                 & 4. Hệ thống tạo thêm một User mới trên hệ thống, hiển thị dữ liệu khuôn mặt chưa được thêm \\ \hline
5. Người dùng nhấn nút \textbf{Add face} để tiến hành tạo dữ liệu khuôn mặt cho họ.        &                                                                                            \\ \hline
                                                                                 & 6. Hệ thống mở camera trước để người dùng chụp ảnh khuôn mặt của mình.                                                                                        \\ \hline
                                        7. Người dùng nhấn nút \textbf{Take} để chụp lần lượt các ảnh khuôn mặt.                                   &                                                                                            \\ \hline
                                                      &       8. Hệ thống ghi nhận lại ảnh được chụp, mỗi ảnh sẽ được đưa lần lượt qua mô hình phát hiện khuôn mặt, ảnh đầu ra được tinh chỉnh bằng phép xoay. Ảnh kết quả sẽ được hiển thị ở góc dưới màn hình. Sau đó, ảnh kết quả được đi qua mô hình nhận diện khuôn mặt để rút trích vectơ đặc trưng. Ảnh kết quả cũng sẽ được đưa qua thuật toán Laplacian để đánh giá độ rõ của ảnh. Nếu điểm đánh giá độ rõ không vượt ngưỡng, hệ thống sẽ không ghi nhận lại bức ảnh hiện tại và yêu cầu người dùng chụp lại. Ngược lại, vectơ đặc trưng sẽ được lưu vào kho dữ liệu.                                                                                   \\ \hline
                    9. Sau khi hệ thống đã ghi nhận nhiều nhất 5 bức ảnh, người dùng có thể chọn nút \textbf{Done} để kết thúc quá trình đăng kí, hoặc chọn nút \textbf{Retry} để chụp lại từ bước 7.                                   &                                                                                            \\ \hline
                                   &    10. Sau khi người dùng kết thúc đăng ký, hệ thống sẽ hiển thị dữ liệu đối với người dùng tương ứng đã được ghi lại.                                                                                  \\ \hline
                                   
\end{tabularx}

\pagebreak
\begin{tabularx}{\linewidth}{|X|X|}
\caption{Đặc tả chức năng cho luồng người dùng chứng thực khuôn mặt}\\
\hline
\textbf{Người dùng}                                                                      & \textbf{Hệ thống}                                                                                  \\ \hline
1. Người dùng chọn nút \textbf{Identify} tại giao diện trang chủ của ứng dụng. &                                                                                            \\ \hline
        &       2. Hệ thống mở camera trước để tiến hành chứng thực khuôn mặt.                                                                                     \\ \hline
    & 3. Luồng dữ liệu từ camera được hệ thống xử lý theo từng frame và lần lượt đưa qua các mô hình. Đầu tiên ảnh được đưa qua mô hình Phát hiện khuôn mặt, nếu phát hiện có ảnh khuôn mặt trong frame, ảnh đầu ra sau đó được áp dụng kỹ thuật xoay để tinh chỉnh lại ảnh khuôn mặt.                                                                                            \\ \hline
        &       4. Ảnh frame được đưa qua mô hình chống giả mạo khuôn mặt, nếu khuôn mặt được xác thực là giả, hệ thống sẽ trả ra kết quả \textbf{Khuôn mặt không hợp lệ}. Nếu ảnh là thật, ảnh kết quả sau đó được đưa qua mô hình nhận diện khuôn mặt.                                                                                  \\ \hline
        &     5. Ảnh kết quả sẽ được rút trích thành vectơ đặc trưng và được so sánh với tất cả các vectơ đặc trưng trong kho dữ liệu đã đăng ký, độ đo tương tự cosine được áp dụng để tìm định danh người dùng, hệ thống sẽ trả ra kết quả là danh tính của người dùng. Nếu hệ thống không tìm được danh tính người dùng thì sẽ trả ra kết quả \textbf{Người lạ}, tức hệ thống nhận diện người dùng chưa từng đăng ký trước đó.                                                                                 \\ \hline
        &       6. Hai ảnh chứa con mắt sẽ được cắt từ ảnh khuôn mặt. Lần lượt cho hai ảnh qua mô hình phân loại mắt đóng mở. Nếu cả hai mắt được phân loại là đóng thì sẽ trả ra kết quả \textbf{Hai mắt đang đóng}, tức là chứng thực không thành công.                                                                                     \\ \hline
                &       7. Hệ thống xử lý frame và trả ra kết quả liên tục khi camera trước vẫn còn hoạt động và kết thúc khi người dùng thoát.                                                                                     \\ \hline
\end{tabularx}

\section{Chi tiết cài đặt các hàm xử lý quan trọng}
\subsection{Hàm xử lý luồng dữ liệu camera}
Các hàm xử lý frame từ luồng dữ liệu của camera và các hàm xử lý về view như để vẽ hoặc theo dõi các kết quả, các bounding box được sử dụng từ ví dụ\footnote{https://github.com/tensorflow/examples/tree/master/lite/examples/image\_classification/android} mà những nhà phát triển TensorFlow cung cấp.

\subsection{Hàm phát hiện mặt người}
Như đã giới thiệu, mô hình phát hiện mặt người được sử dụng chính là mạng MTCNN. Cài đặt hàm xử lý ảnh đưa qua mô hình MTCNN trên ngôn ngữ Java của Android được sử dụng từ tác giả \textbf{vcvycy}\footnote{https://github.com/vcvycy/MTCNN4Android}.

Tại bước này, chúng em giới hạn độ lớn bề ngang tối thiểu của ảnh khuôn mặt phải  $\geq \frac{W}{5}$ với W là bề ngang của frame ảnh. Chúng em nhận xét vùng mặt được phát hiện phải bị ràng buộc, tránh các rủi ro có thể xảy ra trong quy trình chứng thực.

Sau khi gọi hàm \textbf{detectFaces} với hai tham số lân lượt là frame ảnh và kích cỡ mặt nhỏ nhất để ràng buộc. Hàm sẽ đưa qua lần lượt ba mô hình con P-Net, R-Net và O-Net. Cuối cùng, hàm trả ra một mảng các Box object.

\subsection{Hàm tinh chỉnh khuôn mặt}
Nếu đầu ra của hàm phát hiện khuôn mặt không xuất hiện một Box nào cả, ta tiến hành cập nhật kết quả lên màn hình và bắt đầu xử lý frame kế tiếp.

Ngược lại, nếu đầu ra của hàm phát hiện khuôn mặt có một hoặc nhiều Box. Tại bước này, đầu tiên ta sẽ chọn Box có kích thước lớn nhất và thực hiện phép xoay ảnh dựa trên Box đó. Ta chọn giá trị tròng mắt trái và tròng mắt phải từ các giá trị landmark của Box, sau đó tính hiệu trên hai tọa độ O\textsubscript{x} và O\textsubscript{y} và xoay ảnh theo một góc \textbf{arctan}. Với ảnh kết quả đã được xoay đúng vị trí, ta tiếp tục thực hiện hàm phát hiện khuôn mặt trên ảnh kết quả, đầu ra sẽ là một mảng các Box đã được tinh chỉnh.

\subsection{Hàm chống giả mạo khuôn mặt}
Ta thực hiện trích xuất đặc trưng từ ảnh lấy từ frame qua mô hình chống giả mạo khuôn mặt, tại đây ta sử dụng ảnh đầy đủ lấy từ frame và không sử dụng ảnh khuôn mặt do dữ liệu huấn luyện là những ảnh đầy đủ không cắt. Cuối cùng hàm trả kết quả điểm số, khi so sánh với ngưỡng cho trước ta sẽ biết được độ thật giả của khuôn mặt. Dựa trên kết quả huấn luyện từ trước của mô hình chống giả mạo khuôn mặt, ta có được nhận xét:
\begin{itemize}
    \item Nếu điểm số $\geq$ 0.65 thì ảnh được dự đoán là giả.
    \item Ngược lại nếu điểm số < 0.65 thì ảnh được dự đoán là thật.
\end{itemize}

Tại quy trình đăng ký khuôn mặt, chúng em chỉ sử dụng hàm Laplacian để đánh giá độ rõ của ảnh. Sử dụng phép tích chập giữa các điểm pixel trong ảnh và Laplacian kernel, ta có thể tính được các điểm góc cạnh có trong ảnh, tổng số điểm góc cạnh đó chính là kết quả của hàm. Nếu kết quả từ hàm Laplacian với ảnh chụp của người dùng không vượt ngưỡng cho trước, hệ thống sẽ yêu cầu chụp lại. Điều này đánh giá được điều kiện môi trường như ánh sáng, để ảnh chụp có chất lượng cao hơn từ đó tăng phần nào độ hiệu quả khi sử dụng trong bước trích xuất đặc trưng.

Tại quy trình chứng thực khuôn mặt, nếu ảnh khuôn mặt khi đưa qua mô hình chống giả mạo khuôn mặt là thật, ta tiếp tục đưa ảnh qua mô hình nhận diện khuôn mặt để tiếp tục xử lý. Ngược lại, nếu khuôn mặt được dự đoán là giả, ta sẽ cập nhật kết quả \textbf{Khuôn mặt không hợp lệ}.

\subsection{Hàm nhận diện khuôn mặt}
Tại quy trình đăng ký khuôn mặt, ảnh sau khi tinh chỉnh sẽ được trích xuất vectơ đặc trưng 1024 chiều. Nếu kết quả tính toán độ rõ của ảnh khuôn mặt từ hàm Laplacian vượt ngưỡng, ta sẽ lưu vectơ đặc trưng tương ứng với người dùng vào trong kho dữ liệu.

Tại quy trình chứng thực khuôn mặt, ảnh sau khi tinh chỉnh sẽ đưa qua mô hình để trích xuất đặc trưng, đầu ra gồm vectơ 1024 chiều. Sau đó ta so sánh với những vectơ 1024 chiều của những ảnh đăng ký trước đó trên hệ thống với độ tương tự cosine. Ta chọn độ đo có xếp hạng (ranking) cao nhất, nếu độ đo đó vượt ngưỡng cho trước, ta lấy được danh tính của người dùng hiện tại. Nếu điểm độ đo đó không vượt ngưỡng, kết quả trả ra cuối cùng sẽ là \textbf{Người lạ}.

\subsection{Hàm phân loại mắt đóng mở}

Hàm này chỉ được sử dụng trong quy trình chứng thực. Ảnh chứa lần lượt mắt trái và mắt phải được cắt từ tọa độ landmark của đôi mắt trong ảnh đã được tinh chỉnh, với kích thước cố định [$\frac{W}{6}$, $\frac{H}{10}$] theo kích thước của khuôn mặt [W, H], sau đó cả hai ảnh được đưa qua hàm xử lý phân loại.

Ta cũng thực hiện trích xuất đặc trưng với đầu ra là điểm tự tin (confidence score) của hai nhãn \textit{đóng} và \textit{mở}. Ta chọn nhãn có điểm tự tin lớn nhất làm kết quả phân loại. Nếu cả hai ảnh được phân loại là \textit{đóng}, đồng nghĩa với việc hệ thống không chấp nhận khuôn mặt.

\section{Giao diện}

\begin{figure}[H]
    \centering
    \includegraphics[width = 0.3 \textwidth]{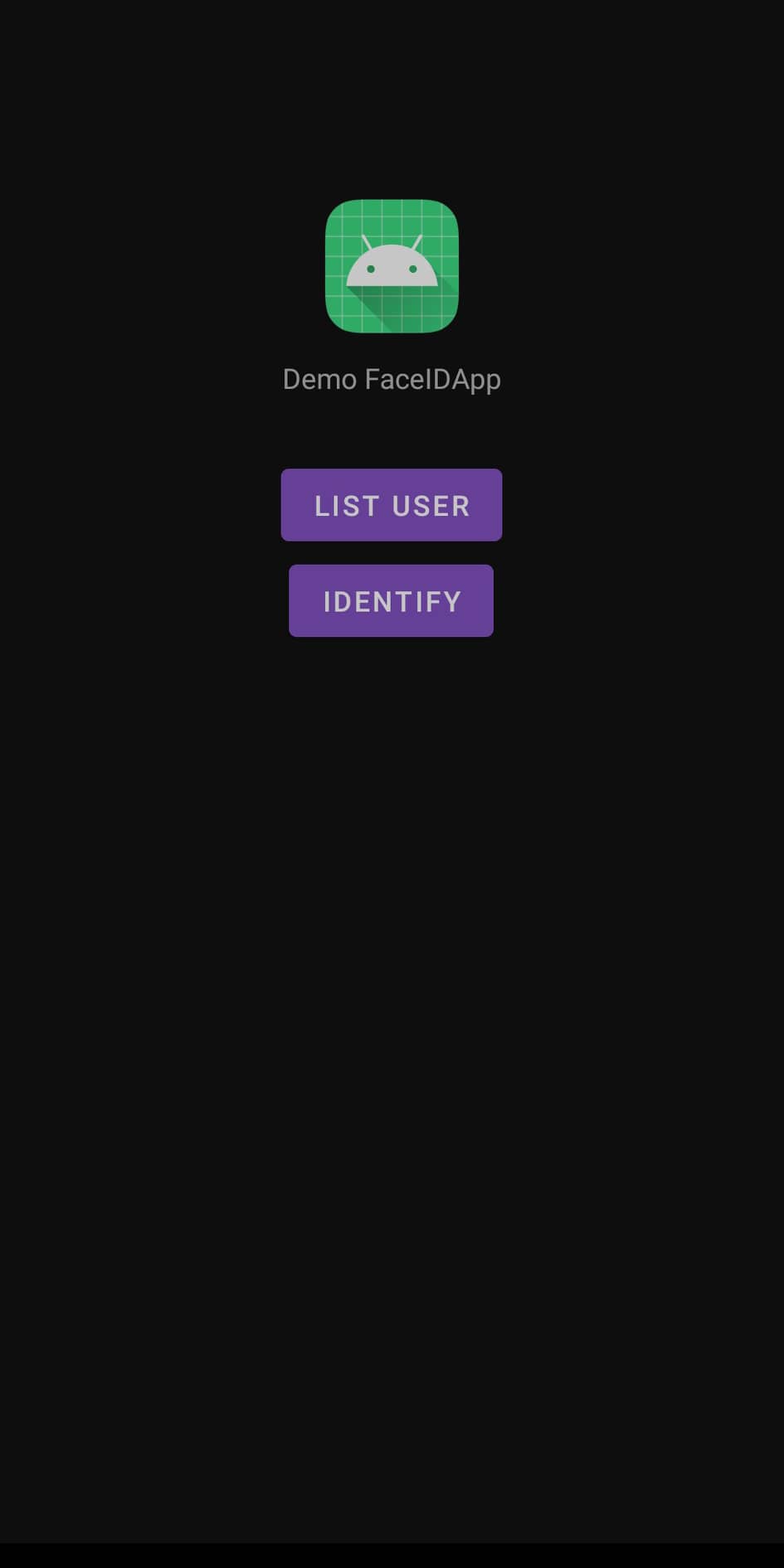} 
    \caption{Giao diện trang chủ.}
    \label{fig:chart6.3}
\end{figure}
Tại đây có hai nút:
\begin{enumerate}
    \item Nút \textbf{List Users} để thao tác các tác vụ liên quan đến đăng ký khuôn mặt.
    \item Nút \textbf{Identify} để người dùng thực hiện chứng thực khuôn mặt.
\end{enumerate}

\begin{figure}[H]
    \centering
    \includegraphics[width = 0.3 \textwidth]{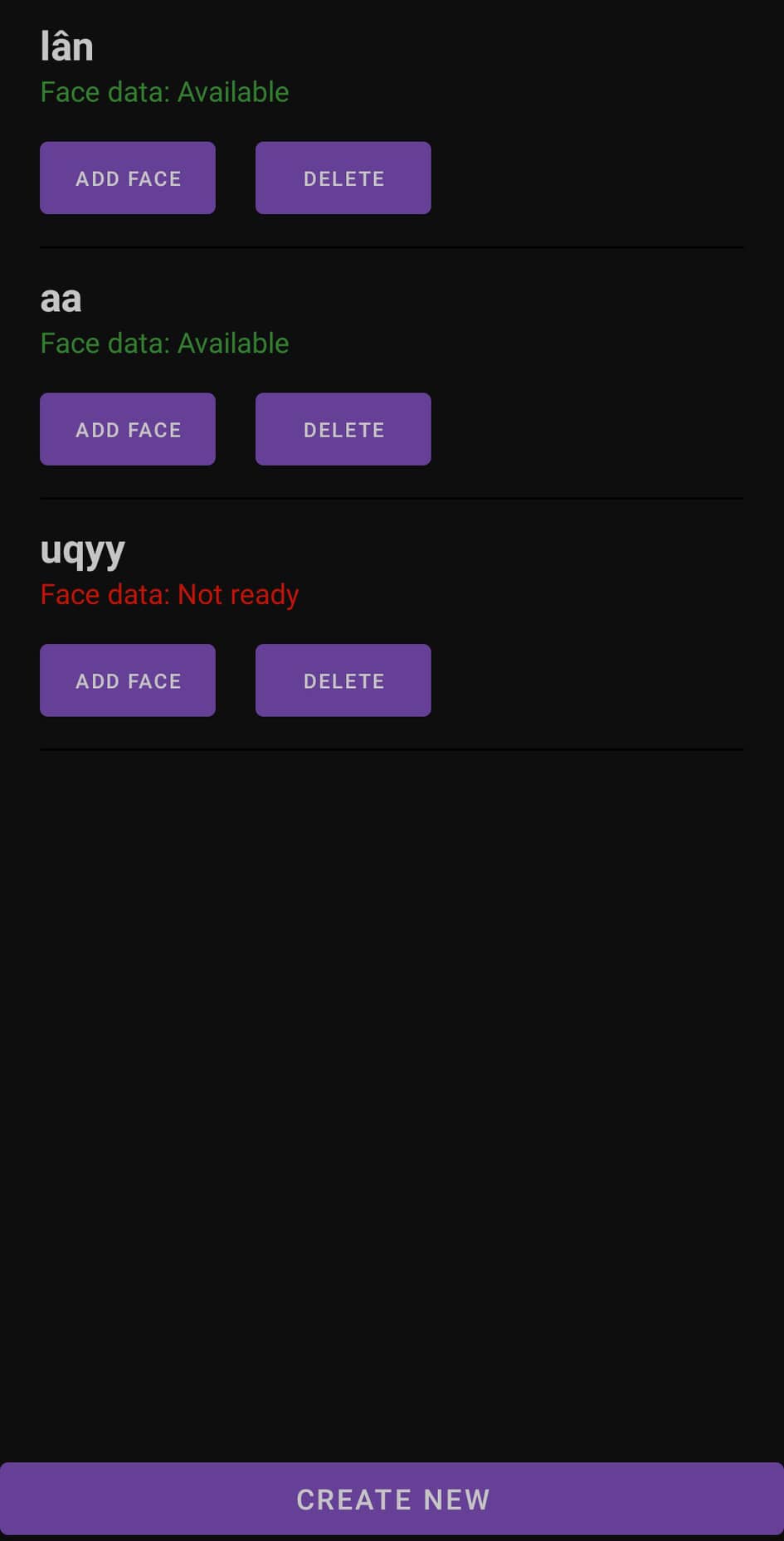} 
    \caption{Giao diện danh sách những người đăng ký khuôn mặt.}
    \label{fig:chart6.4}
\end{figure}

Tại đây ta có thể thực hiện các thao tác như sau:
\begin{itemize}
    \item Chọn nút \textbf{Create New} và nhập tên người dùng để đăng ký tên người dùng trên hệ thống. 
    \item Chọn nút \textbf{Add face} để tiến hành chụp ảnh để tạo dữ liệu khuôn mặt trên hệ thống.
    \item Chọn nút \textbf{Delete} để xóa dữ liệu người dùng trên hệ thống. 
\end{itemize}

\begin{figure}[H]
    \centering
    \includegraphics[width = 0.3 \textwidth]{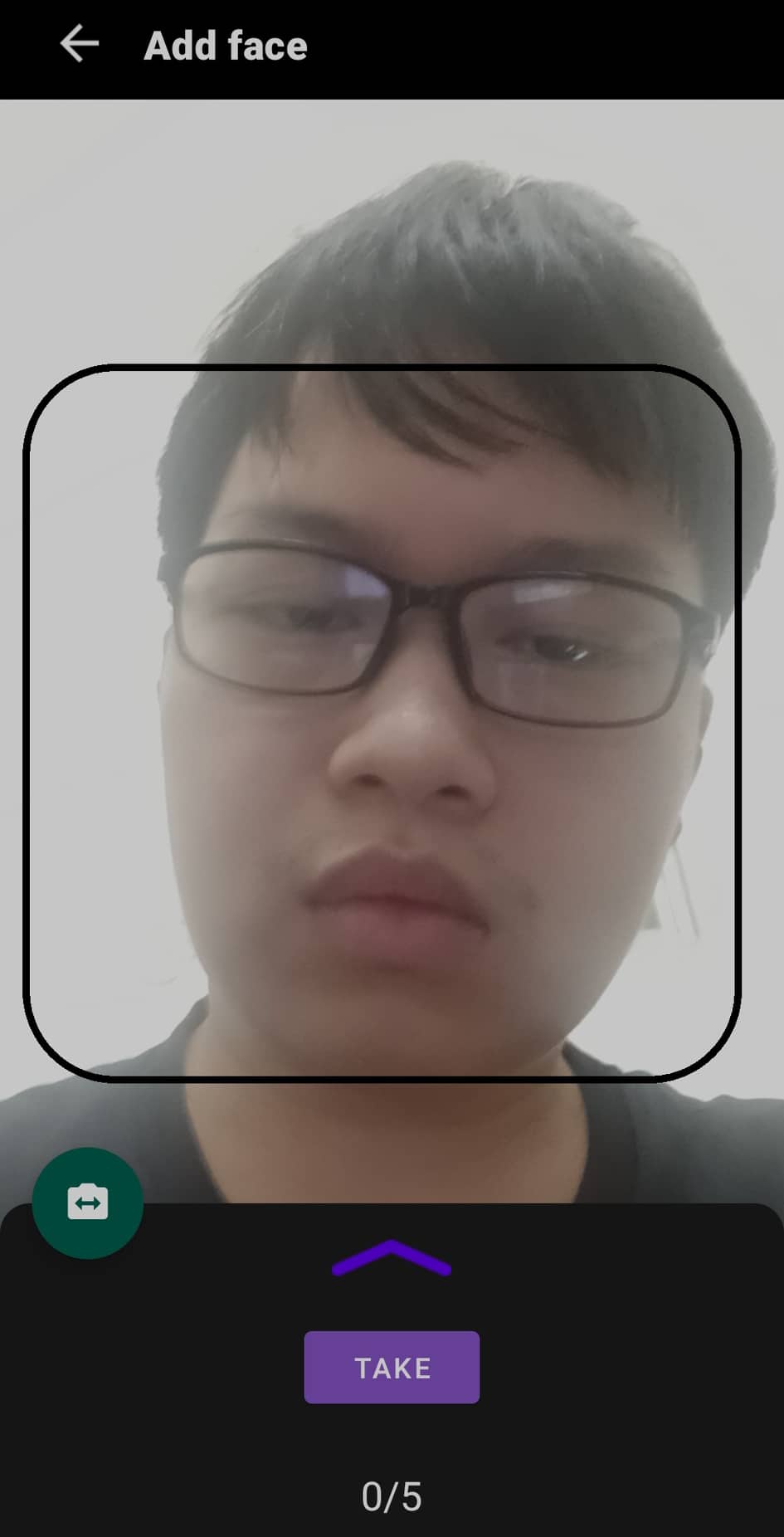} 
    \caption{Giao diện người dùng chụp ảnh để tạo dữ liệu khuôn mặt trên hệ thống.}
    \label{fig:chart6.5}
\end{figure}

Người dùng được chụp tối đa 5 tấm hình để tạo dữ liệu khuôn mặt. Trong quá trình chụp, hệ thống có thể yêu cầu người dùng chụp lại nếu ảnh chụp không hợp lệ đối với hệ thống. Lúc này ảnh sẽ không được ghi nhận trong hệ thống và không hiển thị ở mục preview bên dưới.

\begin{figure}[H]
    \centering
    \includegraphics[width = 0.3 \textwidth]{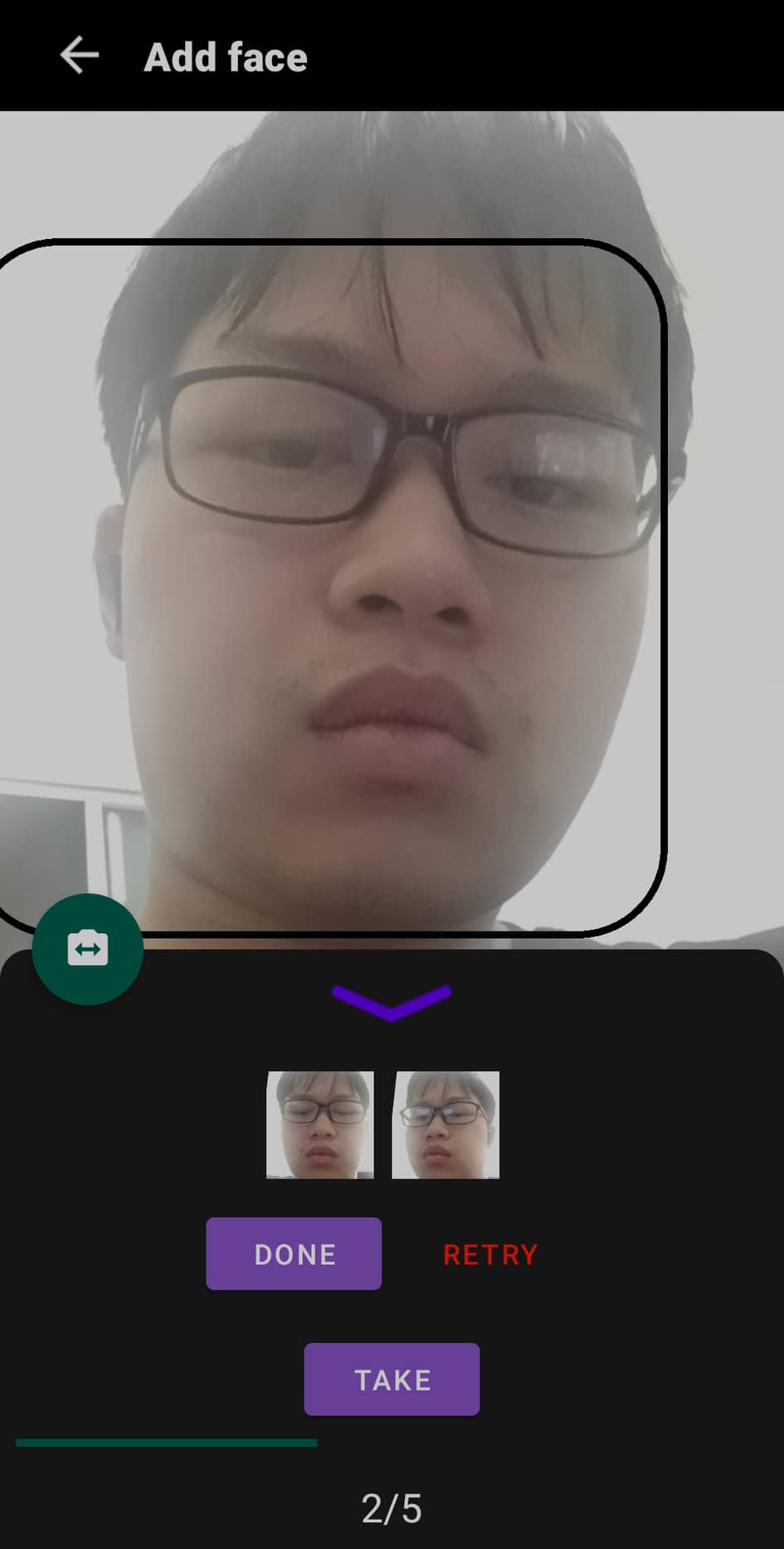} 
    \caption{Giao diện minh họa quá trình đăng ký ảnh khuôn mặt.}
    \label{fig:chart6.6}
\end{figure}

\begin{figure}[H]
    \centering
    \includegraphics[width = 0.3 \textwidth]{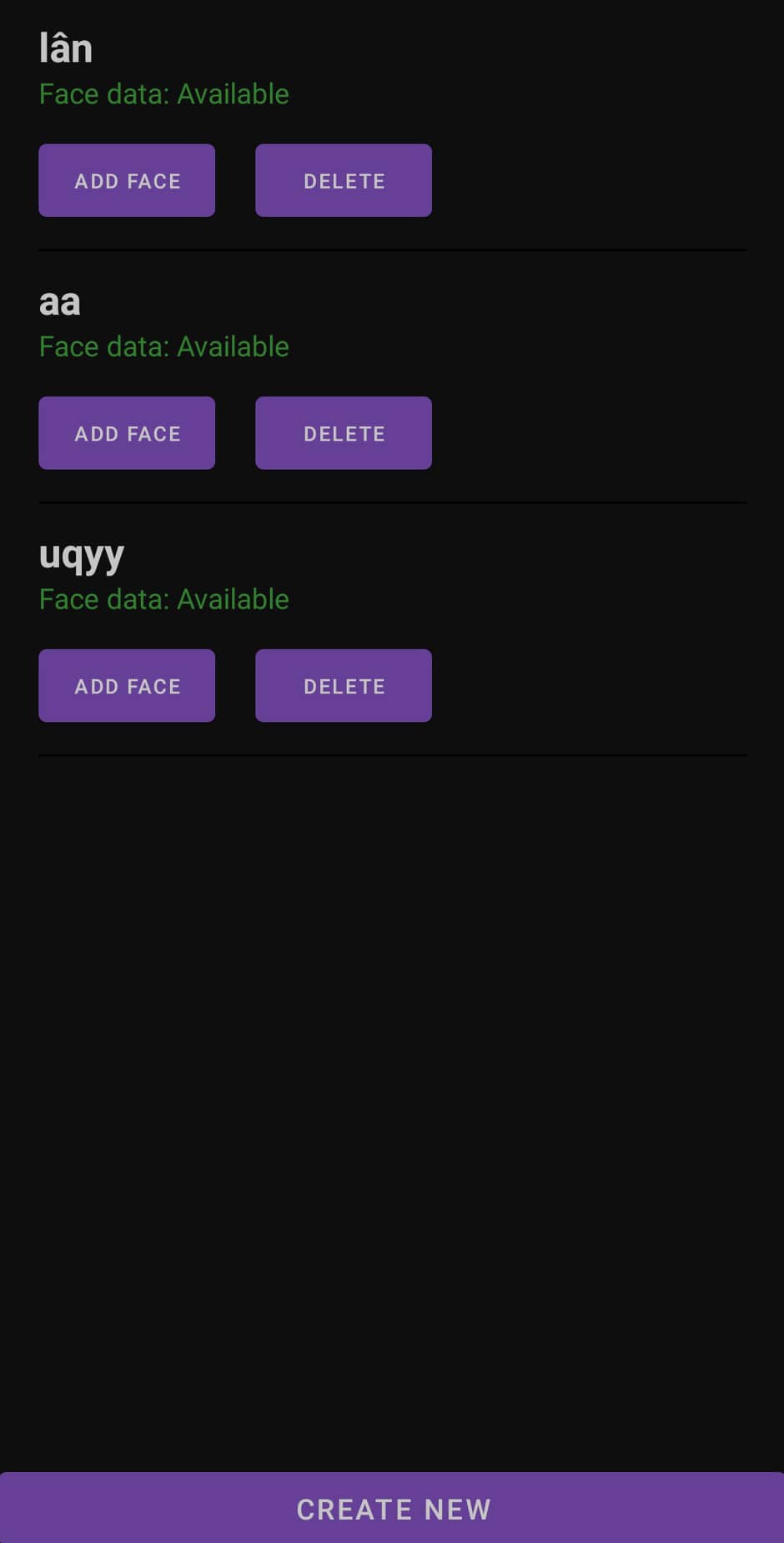} 
    \caption{Giao diện hệ thống cập nhật dữ liệu cho người đã đăng ký khuôn mặt.}
    \label{fig:chart6.7}
\end{figure}

Khi chụp xong người dùng có thể chọn:
\begin{itemize}
    \item Nút \textbf{Retry} để tiến hành chụp lại từ đầu, các ảnh đã chụp sẽ được loại bỏ.
    \item Nút \textbf{Done} để kết thúc quá trình đăng ký. Lúc này người dùng sẽ được chuyển về trang danh sách đăng ký và hệ thống sẽ hiển thị người dùng đã đăng ký khuôn mặt, như Hình \ref{fig:chart6.7}.
\end{itemize}

\begin{figure}[H]
    \centering
    \includegraphics[width = 0.3 \textwidth]{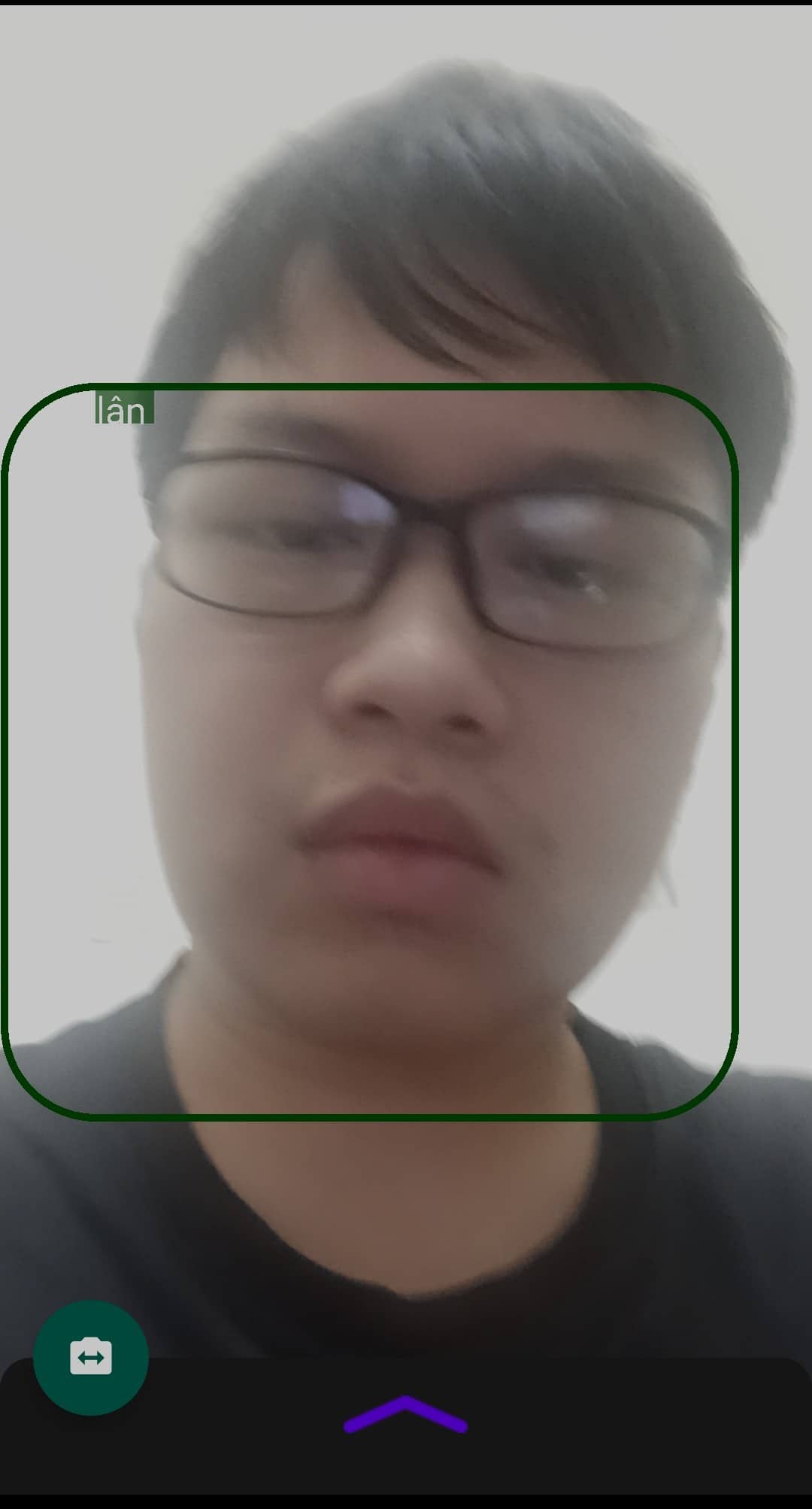} 
    \caption{Giao diện minh họa quy trình chứng thực.}
    \label{fig:chart6.8}
\end{figure}

Khi người dùng chọn nút \textbf{Identify}, hệ thống sẽ mở camera và xử lý ảnh qua các mô hình học sâu. Kết quả sẽ được hiển thị liên tục trên màn hình, có thể được chia ra thành các mục như sau:

\begin{itemize}
    \item \textbf{Khuôn mặt không hợp lệ}: kết quả này có ý nghĩa rằng hệ thống nhận định khuôn mặt đang phát hiện là giả. Bounding box được hiển thị có màu đỏ.
    \item \textbf{Người lạ}: kết quả này có ý nghĩa rằng hệ thống nhận định khuôn mặt đang phát hiện chưa được đăng ký trong hệ thống. Bounding box hiển thị cũng sẽ có màu đỏ.
    \item \textbf{Hai mắt đang đóng}: Kết quả này có ý nghĩa hệ thống đã xác định được danh tính người dùng nhưng người dùng không mở cả hai mắt, trường hợp này vẫn được cho là xác thực không thành công bởi người dùng không chú ý đến thiết bị. Bounding box hiển thị vẫn sẽ có màu đỏ.
    \item Hệ thống hiển thị kết quả là tên của một người dùng nào đó đăng ký trong hệ thống. Bounding box được hiển thị có màu xanh lục.
\end{itemize}

Tổng thời gian xử lý (Inference time) của quy trình chứng thực, với một frame đưa qua các mô hình học sâu sẽ rơi vào khoảng 0.3$s$ đến 0.5$s$, tương đương 2-3 $fps$.

\section{Kết luận}
Nhóm đã thực hiện cài đặt hai quy trình cơ bản trong một hệ thống chứng thực khuôn mặt trên ứng dụng nền tảng Android. Tại quy trình chứng thực, chúng em minh họa quy trình chứng thực như một quy trình nhận diện khuôn mặt, bởi khi đó chúng ta mới có thể quan sát cách hệ thống xử lý ảnh và hiển thị kết quả. Quy trình này được xử lý khác với một quy trình chứng thực thường được thấy trong triển khai thực tế, chẳng hạn như: mở khóa khuôn mặt, thanh toán điện tử, vân vân.

%% file: template/chapter7.tex
\chapter{Kết quả nghiên cứu}
\label{Chapter7}

\section{Đánh giá và so sánh}
\subsection{Nhận diện khuôn mặt}
Như đã giới thiệu trong \ref{sec:Face_rec}, chúng em huấn luyện mô hình nhận diện mặt người trên tập dữ liệu CelebA \cite{liu2015faceattributes}. Sử dụng hàm LMCot \ref{eq:LMCot} mà chúng em đề xuất đã cho kết quả huấn luyện rất tốt trên tập dữ liệu CelebA \cite{liu2015faceattributes}, hơn các phương pháp tốt nhất hiện nay nếu cùng huấn luyện trên cùng một tập dữ liệu. \textbf{Đặc biệt với tập CelebA thì tập huấn luyện, tập xác thực và tập kiểm tra bao gồm những người khác nhau và không trùng nhau}. Tức là với với một người bất kỳ thì ảnh người đó chỉ xuất hiện trong một tập (huấn luyện hoặc xác thực hoặc kiểm tra).
\begin{table}[htbp]
\caption{Kết quả trên tập CelebA \cite{liu2015faceattributes} và tập CelebA+mask \cite{mare2021realistic}}
\label{table02}
\begin{center}
\begin{threeparttable}
    \begin{tabular}{|l|c|c|}
    \hline
                          & EER trên CelebA & EER trên CelebA+mask \\ \hline
    \textbf{LMCot (ArcFace-based)} & \textbf{7.3\%}           & \textbf{7.7\%}                 \\ \hline
    ArcFace               & 7.35\%          & 7.77\%                  \\ \hline
    CosFace               & 7.33\%          & 7.8\%                 \\ \hline
    SphereFace            & 7.42\%          & 7.82\%                  \\ \hline
    \end{tabular}
    
    \begin{tablenotes}
      \small
      \item Tỷ lệ lỗi ngang nhau (ERR) (\%) của các hàm suy hao. Với kiến trúc là EfficientNetv2S và tập huấn luyện cũng như kiểm tra là các phần tương ứng của tập CelebA \cite{liu2015faceattributes} và tập CelebA+mask \cite{mare2021realistic}.
    \end{tablenotes}
\end{threeparttable}
\end{center}
\end{table}

\begin{figure}[H]
    \centering
    \subfloat[Biểu đồ của mô hình gốc khi huấn luyện trên ImageNet \cite{deng2009imagenet}. \label{fig:hist_init}]{\includegraphics[width=0.4\linewidth]{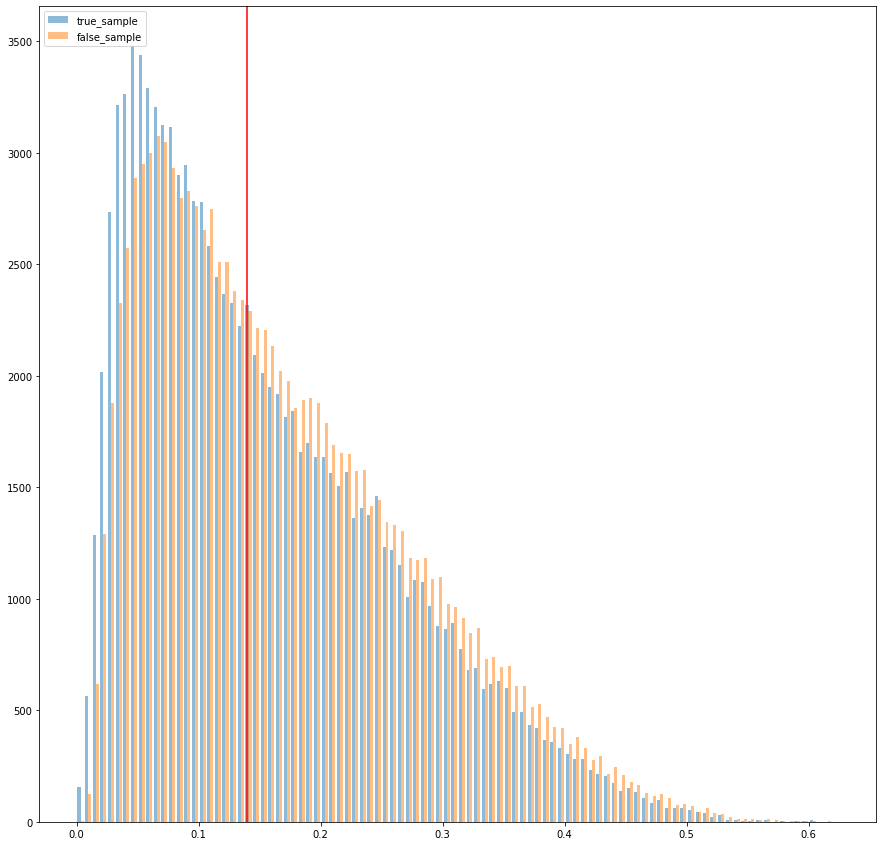}}
    \qquad
    \subfloat[Biểu đồ khi huấn luyện bằng hàm ArcFace. \label{fig:hist_arcface}]{\includegraphics[width=0.4\linewidth]{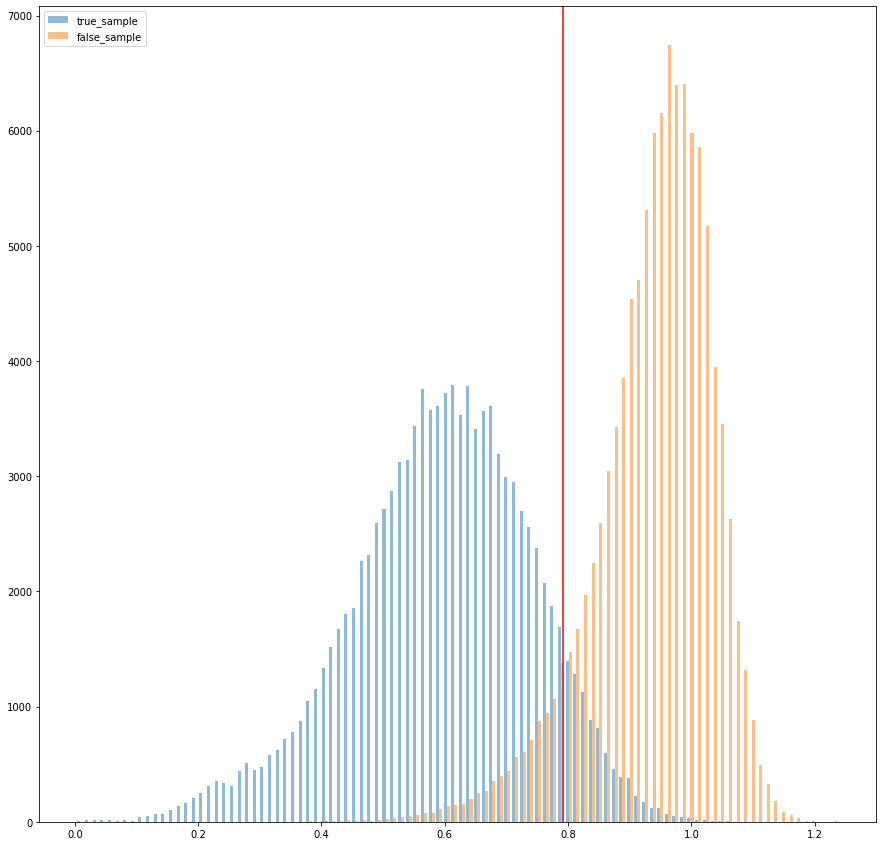}} 
    \quad 
    \subfloat[Biểu đồ khi huấn luyện bằng hàm LMCot. \label{fig:hist_lmcot}]{\includegraphics[width=0.8\linewidth]{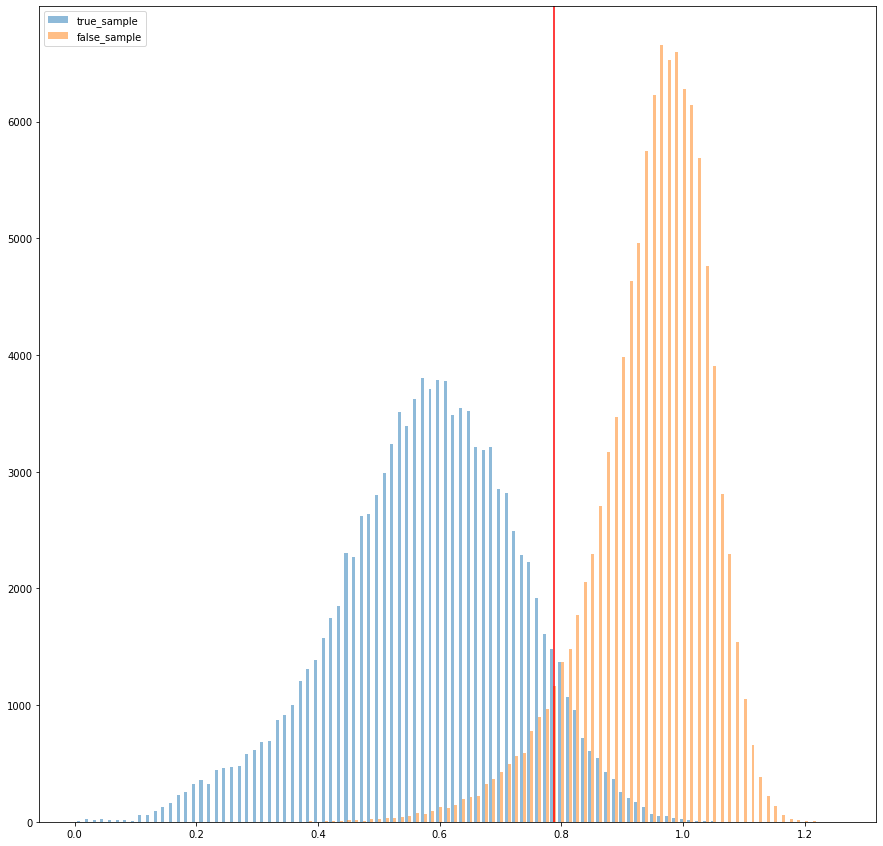}}
    \caption{\textbf{Biểu đồ tần suất} cho khoảng cách cosin của 100 nghìn mẫu ngẫu nhiên các cặp ảnh từ cùng một người và 100 nghìn mẫu ngẫu nhiên các cặp ảnh từ hai người khác nhau, dữ liệu từ bộ thử nghiệm của tập CelebA.}
    \label{fig:LMCot-hist}
\end{figure}
 Biểu đồ tần suất với trục hoành là khoảng cách giữa hai ảnh, trục tung là số lượng mẫu thử. Phần giao nhau là phần còn sai của mô hình. Phần màu xanh là khoảng cách các cặp cùng một người và phần màu cam là khoảng cách các cặp không cùng một người. Đường màu đỏ thể hiện giá trị mà tại đó tỷ lệ chấp nhận sai (FAR) bằng tỷ lệ từ chối sai (FRR). Do nhóm sử dụng mô hình nhỏ như EfficientNetV2S để phù hợp hóa trên các ứng dụng điện thoại, nên độ chính xác chưa đạt mức tuyệt đối.

\begin{figure}[H]
    \centering
    \subfloat[Kết quả khả năng rút trích đặc trưng của mô hình gốc khi huấn luyện trên ImageNet \cite{deng2009imagenet}. \label{fig:pca_init}]{\includegraphics[width=0.57\linewidth]{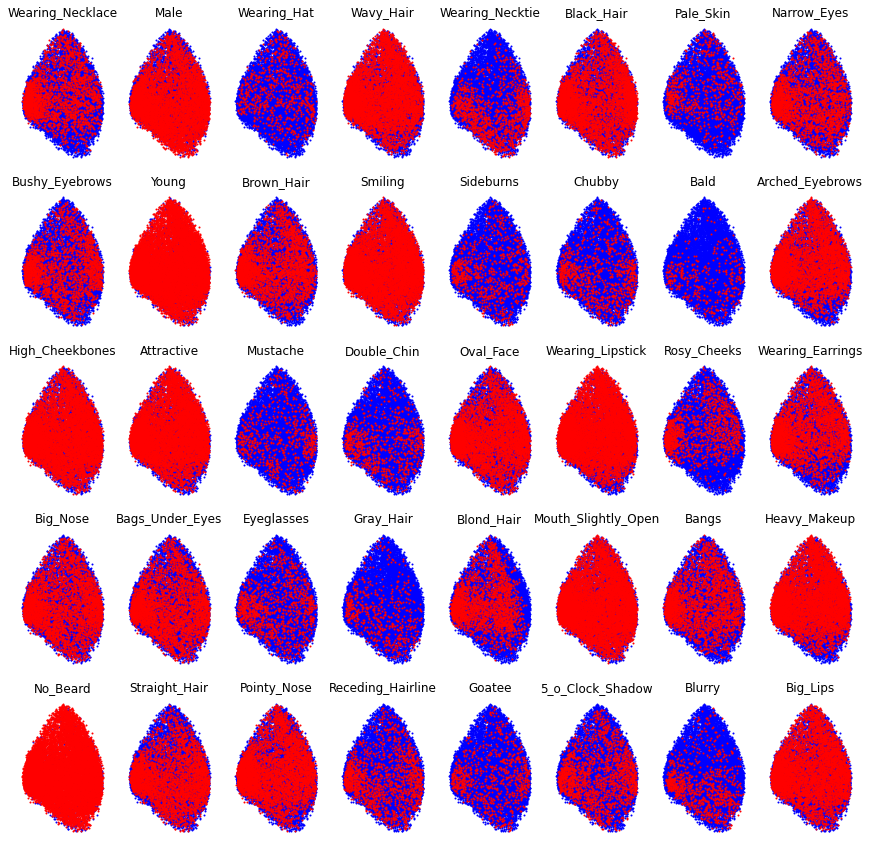}} 
    \qquad
    \subfloat[Kết quả khả năng rút trích đặc trưng của mô hình khi huấn luyện bằng LMCot. \label{fig:pca_lmcot}]{\includegraphics[width=0.57\linewidth]{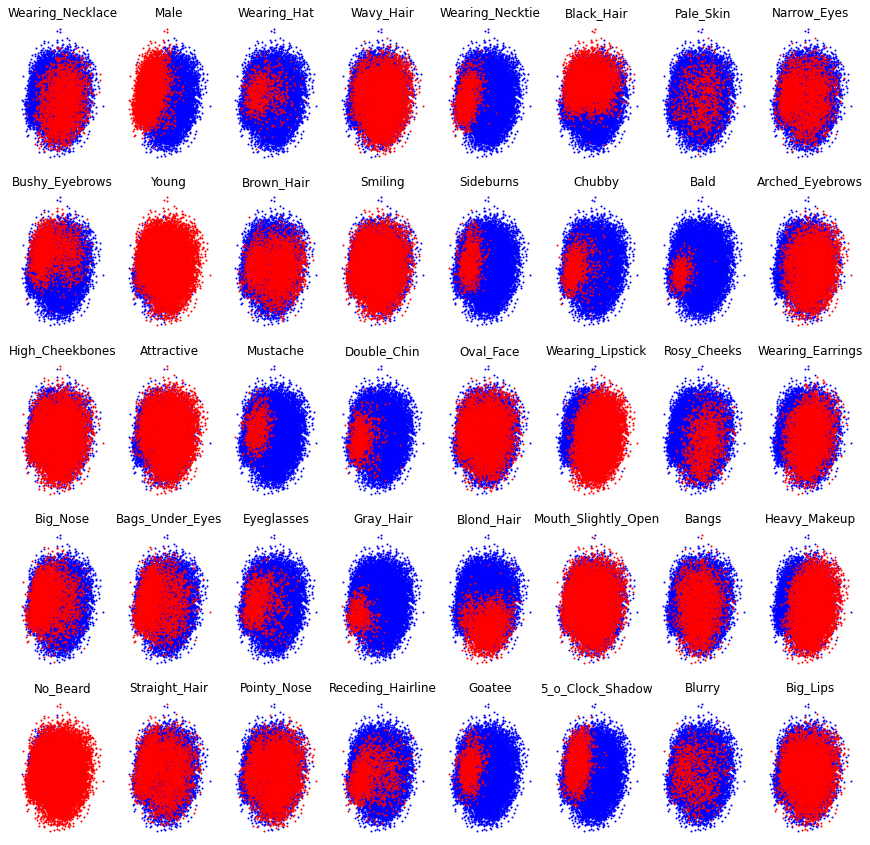}}
    \caption{Hình minh họa khả năng rút trích đặc trưng của mô hình.}
    \label{fig:LMCOt-PCA}
\end{figure}
Sau khi rút trích đặc trưng ở dạng vectơ 1024 chiều, ta dùng thuật toán PCA \cite{wold1987principal} để giảm chiều của vectơ đặc trưng về còn 2 chiều $(x,y)$. Ta vẽ tọa độ đó lên trục tọa độ $Oxy$. Trong tập CelebA ta có 40 đặc trưng theo dạng đúng/sai cho mỗi ảnh như: nam, trẻ, đeo kính, vân vân. Phần màu đỏ là những ảnh có đặc trưng đó, phần màu xanh là những ảnh không có đặc trưng đó. Ta có thể thấy sau khi được huấn luyện mô hình tách biệt 2 phần xanh và đỏ khá rõ ràng. Mặt khác, vì dùng phép tăng cường ảnh thay đổi chất lượng ảnh, nên mô hình bền vững trước độ đo ảnh có bị mờ hay không (blurry). Hay nói cách khác mô hình ít bị ảnh hưởng khi ảnh đầu vào có chất lượng kém.
 
\subsection{Chống giả mạo khuôn mặt}
Trong phần này chúng em huấn luyện hai kiến trúc mạng là MobileNet và MobileNetV2 \cite{sandler2018mobilenetv2}. MobileNetV2 cũng là một mô hình thích hợp dùng trên các thiết bị điện thoại do kích thước nhỏ (14MB), độ chính xác cao và tốc độ xử lý nhanh. 

Lần lượt mỗi mô hình chúng em sẽ dùng hàm cross-entropy so sánh với việc dùng hàm double loss. Các kết quả cho thấy việc dùng hàm cross-entropy kết hợp với dùng double loss sẽ cho kết quả tốt hơn rất nhiều. Trong đó \textbf{cross-entropy} là phương pháp huấn luyện chỉ dùng hàm cross-entropy, \textbf{cross-entropy + double loss} là phương pháp huấn luyện dùng cả cross-entropy và double loss như đã đề cập trong \ref{sec:DoubleLoss}.

AUC là từ viết tắt cho \textbf{Area Under The Curve}. Độ đo AUC có thể diễn giải như sau: Xác suất của một mẫu dương tính được lấy ngẫu nhiên sẽ được xếp hạng cao hơn một mẫu âm tính được lấy ngẫu nhiên. Vậy nên kết quả AUC càng cao thì mô hình càng tốt. $AUC = P\left(score(x^+) > score(x^-)\right)$.

\begin{table}[htpb]
\caption{So sánh giữa các mô hình huấn luyện trên tập huấn luyện của tập CelebA-Spoof/CelebA-Spoof  + mask và đánh giá trên tập kiểm tra tương ứng của cùng bộ dữ liệu.}
\begin{tabularx}{\linewidth}{|l|XX|}
\hline
\multicolumn{1}{|c|}{\multirow{2}{*}{Hàm suy hao}} & \multicolumn{2}{c|}{AUC trên dataset}                       \\ \cline{2-3} 
\multicolumn{1}{|c|}{}                             & \multicolumn{1}{c|}{CelebA-Spoof}  & CelebA-Spoof  + mask   \\ \hline
MobileNet(cross-entropy)                           & \multicolumn{1}{c|}{0.955}          & 0.949                 \\ \hline
MobileNet(cross-entropy + double)                  & \multicolumn{1}{c|}{0.981}          & 0.965                 \\ \hline
MobileNetV2(cross-entropy)                         & \multicolumn{1}{c|}{0.958}          & 0.949                 \\ \hline
MobileNetV2(cross-entropy + double)                & \multicolumn{1}{c|}{0.992}          & 0.982                 \\ \hline
ResNet50(cross-entropy)                            & \multicolumn{1}{c|}{0.962}          & 0.958                 \\ \hline
\textbf{ResNet50(cross-entropy + double)}          & \multicolumn{1}{c|}{\textbf{0.997}} & \textbf{0.985}        \\ \hline
\end{tabularx}
\end{table}

\begin{figure}[H]
    \centering
    \subfloat[Biểu đồ tần suất của mô hình huấn luyện bằng cross-entropy. AUC: 0.958 \label{fig:live_spoof_entropy}]{\includegraphics[width=0.45\linewidth]{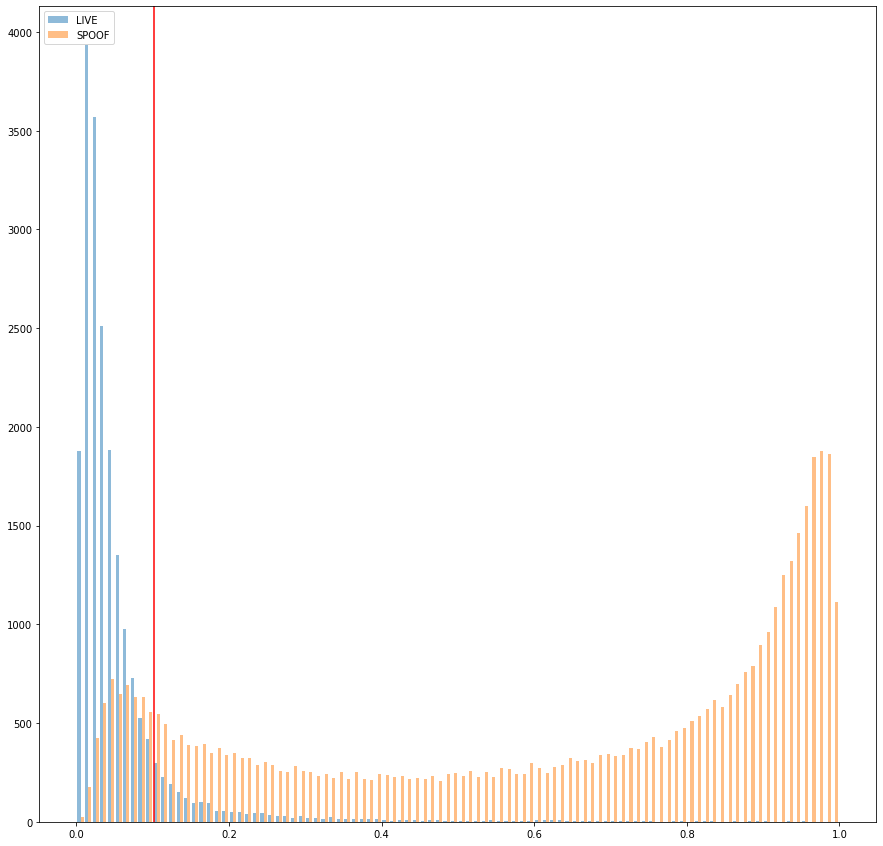}} 
    \qquad
    \subfloat[Biểu đồ tần suất của mô hình huấn luyện bằng double loss. AUC: 0.992 \label{fig:live_spoof_double}]{\includegraphics[width=0.45\linewidth]{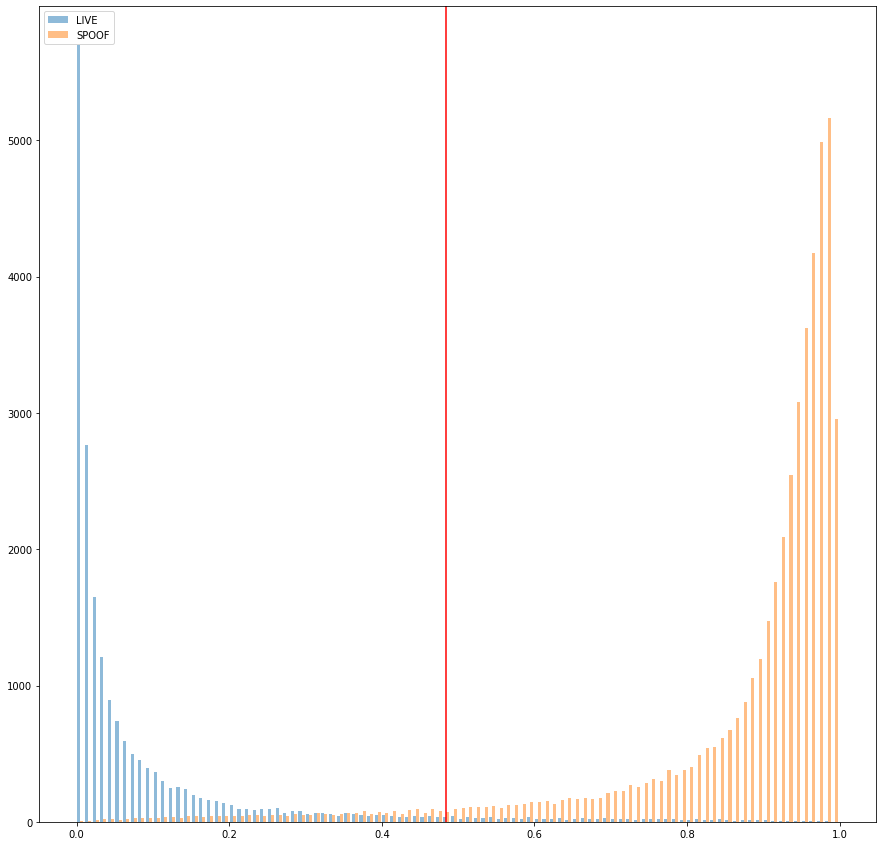}}
    \caption{\textbf{Biểu đồ tần suất} của ảnh thật và ảnh giả.}
    \label{fig:live_spoof_hist}
\end{figure}
Biểu đồ tần suất với trục hoành là mức độ giả mạo, trục tung là số lượng ảnh. Phần giao nhau là phần còn sai của mô hình. Mô hình có kiến trúc là MobileNetV2 \cite{sandler2018mobilenetv2}, huấn luyện bằng hàm suy hao double loss \ref{fig:live_spoof_double} và cross-entropy \ref{fig:live_spoof_entropy}. Mô hình được huấn luyện trên tập huấn luyện của bộ dữ liệu CelebA-Spoof \cite{CelebA-Spoof} và được kiểm tra bằng tập kiểm tra tương ứng của cùng tập dữ liệu. Phần màu xanh là ảnh thật và phần màu cam là ảnh giả mạo. Đường màu đỏ là vị trí mà tại đó tỷ lệ chấp nhận sai (FAR) bằng tỷ lệ từ chối sai (FRR). Mô hình có kiến trúc MobileNetV2 rất nhẹ, khoảng 14MB và thường được dùng cho các ứng dụng trên diện thoại, vậy nên mô hình chưa đạt độ chính xác tuyệt đối.
 
\subsection{Phân loại mắt đóng mở}
Trong phần này, chúng em so sánh độ chính xác giữa các mô hình huấn luyện trên tập MRL Eye và đánh giá trên tập CEW, theo phương pháp đánh giá cross-validation.
\begin{table}[htpb]
\begin{tabular}{|l|c|}
\hline
Phương pháp                                                                                          & Độ chính xác (\%) \\ \hline
EyeNet (Rahman \cite{Rahman2020EyeNetAI})                                           & 90.94             \\ \hline
EfficientNetB0                                                                        & 89.063            \\ \hline
\textbf{\begin{tabular}[c]{@{}l@{}}EfficientNetB0 + RandomResizedCrop\\ (Đề xuất)\end{tabular}} & \textbf{92.117}   \\ \hline
\end{tabular}
\caption{Bảng so sánh độ chính xác giữa các mô hình huấn luyện trên tập MRL Eye và đánh giá trên tập CEW.}
\end{table}

\begin{figure}[H]
    \centering
    \includegraphics[width = 0.6 \textwidth]{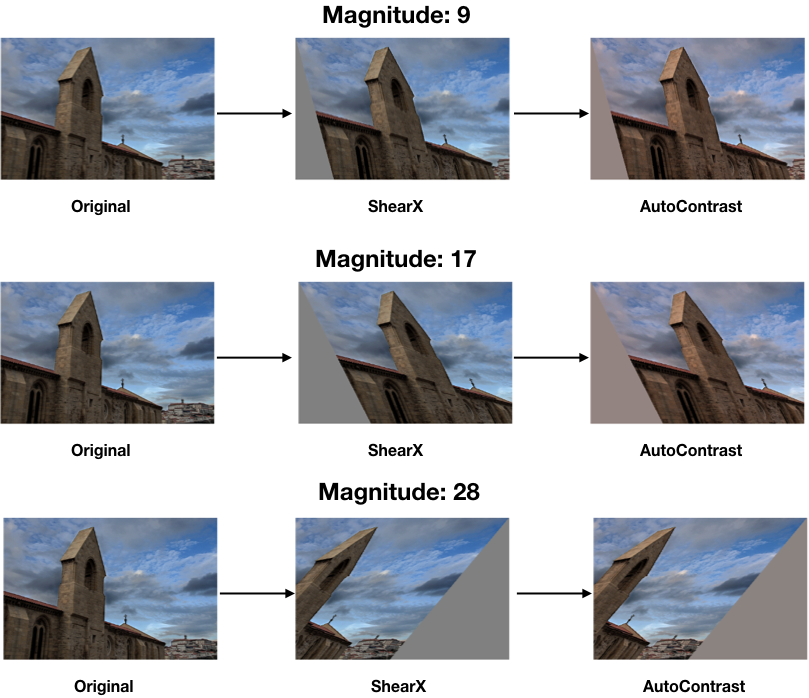}
    \caption{Ví dụ minh họa ảnh gốc sử dụng kỹ thuật tăng cường RandAugment \cite{NEURIPS2020_d85b63ef}.}
    \label{fig:chart7.1}
\end{figure}

Với hai siêu tham số $N$ là số lượng phép biến đổi tăng cường và magnitude $M$ là độ lớn của tất cả các phép biến đổi. Để duy trì tính đa dạng của hình ảnh, các chính sách và xác suất để áp dụng mỗi phép biến đổi với xác suất đồng nhất là $\frac{1}{K}$. Ở đây $K$ là số phương án biến đổi. Vì vậy, với $N$ phép biến đổi cho một hình ảnh huấn luyện, RandAugment có thể thể hiện các chính sách với số lượng $KN$.

Các biến đổi được áp dụng bao gồm auto-contrast, cân bằng, xoay, phân cực, chỉnh màu, thay đổi độ tương phản, thay đổi độ sáng, thay đổi độ sắc nét, shear-x, shear-y, translate-x, translate-y.

Trong các ví dụ như Hình \ref{fig:chart7.1}. $N$ = 2 và ba cường độ được hiển thị tương ứng với các cường độ biến dạng tối ưu. Khi cường độ biến dạng tăng lên, độ mạnh của việc tăng cường tăng lên.

Tại mô hình phân loại mắt đóng mở, chúng em huấn luyện một mô hình CNN đơn giản sử dụng phần khung là kiến trúc EfficientNetB0, áp dụng thêm kỹ thuật tăng cường dữ liệu RandAugment. Kết quả đánh giá khá tốt với độ chính xác 89.063\%. 

\begin{figure}[H]
    \centering
    \includegraphics[width = 0.8 \textwidth]{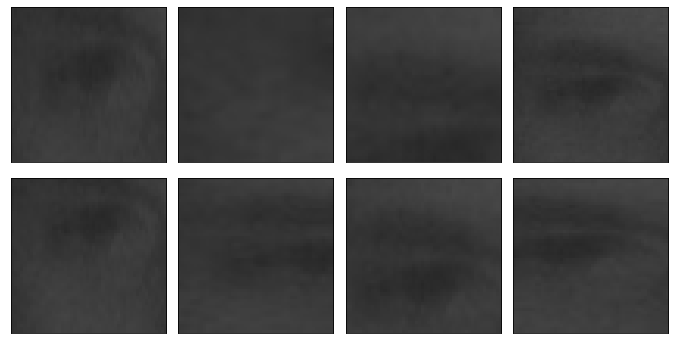}
    \caption{Ví dụ minh họa một ảnh huấn luyện trong tập MRL Eye sử dụng kỹ thuật tăng cường RandomResizedCrop \cite{chen2015mxnet}.}
    \label{fig:chart7.2}
\end{figure}

Sau đó chúng em áp dụng thêm kỹ thuật RandomResizedCrop, như Hình \ref{fig:chart7.2}. 
Ta lấy ví dụ một ảnh gốc có kích thước 56x56, khi áp dụng kỹ thuật này với ảnh mục tiêu có kích thước 24x24, một vùng hình vuông có kích thước 24x24 sẽ được chọn ngẫu nhiên trong vùng 56x56 và được sử dụng cho bước huấn luyện.

Kết quả đánh giá khi vận dụng kỹ thuật tăng cường dữ liệu này tăng khá đáng kể, đạt được độ chính xác 92.117\%. Chúng em đã thử nghiệm nhiều phương pháp tăng cường khác nhưng chúng làm kết quả huấn luyện trở nên kém hơn hẳn, một số phương pháp được liệt kê như sau:
\begin{itemize}
    \item Random Vertical Flip: lật ảnh theo chiều dọc
    \item Random Horizontal Flip: lật ảnh theo chiều ngang.
    \item Color Jitter: thay đổi độ sáng, độ tương phản, độ rực/độ bão hòa của màu.
    
    \item Random Perspective: biến đổi về không gian như độ méo của ảnh.
    
    \item Random Erasing: chọn một vùng hình chữ nhật ở dạng ảnh Tensor và xóa các giá trị pixel trong vùng đó.
    
\end{itemize}

\section{Kết quả đạt được}



\noindent Tổng kết lại, những đóng góp của nhóm được thể hiện qua các mục sau:
\begin{enumerate}
    \item Nhóm đã đề xuất một hệ thống chứng thực sử dụng các mô hình học sâu.
    \item Nhóm đã giới thiệu hai hàm suy hao mới - LMCot và Double loss, được lần lượt sử dụng trong mô hình nhận diện khuôn mặt và mô hình chống giả mạo khuôn mặt. Hơn thế, hàm suy hao LMCot không chỉ ứng dụng trong bài toán nhận diện khuôn mặt mà còn có tiềm năng trong các bài toán đo lường mức độ tương tự hoặc liên quan của hai đối tượng (similarity learning).
    \item Nhóm đã phân tích, cài đặt và cung cấp các số liệu đánh giá về các hàm suy hao mới này trên nhiều tập dữ liệu và nhiều cuộc thi để chứng minh độ hiệu quả của chúng trong đề tài nghiên cứu. Công trình nghiên cứu về hàm Large Margin Cotangent Loss đã được nộp tại hội nghị \textbf{ACOMPA'2022}.
    \item Nhóm cài đặt một ứng dụng trên nền tảng Android để thử nghiệm hệ thống chứng thực đã đề xuất.
\end{enumerate}

%% file: template/chapter8.tex
\chapter{Hướng phát triển của đề tài}
\label{Chapter8}

\noindent Nhóm dự định tiếp tục cải tiến phát triển một số kiến trúc đã được ứng dụng trong hệ thống và giới thiệu thêm một mô hình ước lượng góc nhìn của mắt.

\section{Phát hiện khuôn mặt}
Hiện tại mô hình làm việc rất tốt với khuôn mặt không đeo khẩu trang. Nhưng với khuôn mặt có đeo khẩu trang, độ chính xác của mô hình giảm xuống đáng kể. Hướng phát triển hiện tại đầu tiên của chúng em chính là tiếp tục huấn luyện mạng MTCNN trên các tập dữ liệu khuôn mặt có đeo khẩu trang, hoặc cải tiến mạng MTCNN ở các mạng R-Net, O-Net sao cho chúng hoạt động tốt với các tập dữ liệu khuôn mặt có khẩu trang.

\section{Nhận diện khuôn mặt}
Chúng em đánh giá bài toán nhận diện khuôn mặt vẫn còn rất nhiều hạn chế và cần các nghiên cứu trong tương lai để giải quyết các hạn chế này.
\subsection{Hạn chế của LMCot}
LMCot tuy tốt hơn các hàm suy hao hiện nay tuy nhiên LMCot vẫn còn rất nhiều hạn chế cần được giải quyết như:
\begin{itemize}
    \item Độ phức tạp khi tính \textbf{tan} và \textbf{cotan}.
    \item Khi thực hiện tính \textbf{cotan} ta phải cài đặt thêm một ngưỡng $\varepsilon$ để tránh dữ liệu vượt quá giới hạn về số bit. Đây cũng là một giới hạn rất quan trọng.
\end{itemize}
\subsection{Hạn chế về độ đo}
Việc sử dụng mạng Siamese ảnh hưởng rất lớn bởi độ đo, do đó ta cần một độ đo hợp lý để mô hình hoạt động tốt hơn. Hiện nay, ta có 2 cách truy vấn chính đó là dựa vào giới hạn khoảng cách, tức là nếu khoảng cách nhỏ hơn một ngưỡng nhất định thì ta nói rằng đó là người A. Tuy nhiên có những lúc đặc trưng của một bức ảnh nằm tách biệt với phần còn lại (kể cả ảnh cùng nhãn), khi đó mặc dù khoảng cách xa nhưng nếu chúng ta xét khoảng cách đầu tiên (top-1 distance) thì lại đúng. Vậy nên ta cần một giải pháp linh hoạt giữa việc dùng ngưỡng khoảng cách và khoảng cách ngắn nhất để tránh xảy ra các trường hợp khoảng cách quá xa cũng như nhận diện nhầm.
\subsection{Hạn chế về mô hình}
Hầu hết các mô hình nhận diện khuôn mặt được thiết kế theo kiến trúc mạng Siamese, do đó chúng ta cần nghiên cứu thêm các kiến trúc mới hiệu quả hơn. Bên cạnh đó, mạng nơ-ron thiếu khả năng giải thích, chúng ta chỉ biết có một vectơ đặc trưng nhưng không hiểu cụ thể đó là những đặc trưng nào.

\section{Chống giả mạo khuôn mặt}
Chống giả mạo là một bài toán cần cập nhật liên tục vì các phương pháp giả mạo ngày một nhiều và rất tinh vi. Các phương pháp giả mạo áp dụng công nghệ như dùng ảnh 3D là rất khó để phân biệt.

Bên cạnh đó việc sử dụng mạng nơ-ron cũng có những rủi ro cần phải đề phòng như tấn công đối kháng (adversarial attack), vân vân.

\section{Phân loại mắt đóng mở}
Mô hình này có nhược điểm khá lớn khi không thể phân loại mắt đóng mở một cách chính xác như bình thường đối với người dùng đeo kính. Đây chính là nhược điểm nói chung của các mô hình phân loại, khi chúng chỉ học được dữ liệu hiện tại và hoạt động rất kém khi có dữ liệu ngoài xuất hiện. Do đó, cách cải tiến hiện tại của nhóm chính là thu thập thêm dữ liệu mắt và tiến hành huấn luyện dữ liệu mắt trần đã có với dữ liệu mới đó. Bởi vì dữ liệu về mắt có đeo kính không có sẵn nên nhóm dự định rằng sẽ sử dụng những dữ liệu về khuôn mặt người có đeo kính và tự cắt vùng mắt để tạo ra một tập dữ liệu mới. Với những trường hợp đặc biệt như người dùng đeo kính râm hoặc các loại kính làm che đi phần mắt, ta sẽ xếp chúng vào lớp mắt đóng.

Ngoài ra, với những trường hợp người dùng có đôi mắt híp, mô hình cũng sẽ hoạt động kém hơn vì loại dữ liệu này khá giống với dữ liệu mắt đóng, mô hình sẽ trở nên thiên vị (bias) sang nhãn mắt đóng, điều này cũng có thể giải quyết tạm thời bằng cách thêm dữ liệu huấn luyện. Nhưng tốt hơn hết chúng ta cần phải mở to đôi mắt khi thực hiện chứng thực, như các hệ thống chứng thực hàng đầu đã đề xuất và áp dụng.

\section{Ước lượng góc nhìn của mắt}
\subsection{Giới thiệu}
Đây là một dạng bài toán nâng cao, có nhiệm vụ đo góc nhìn của mắt đối với camera của thiết bị. Hệ thống Face ID của Apple, một trong những hệ thống chứng thực hiện đại nhất. Họ cho rằng ngoài tính năng nhận dạng con mắt, hệ thống còn phải tính toán được mức độ chú ý của người dùng, nếu người dùng thực sự chú ý đến thiết bị, tức là họ muốn hệ thống mở khóa, thì hệ thống sẽ cho phép mở khóa, ngược lại thì không.

Cấu tạo của mắt thật sự rất đặc biệt, góc nhìn của con dao động một phạm vi khá lớn, khoảng 120$^{\circ}$  đến 200$^{\circ}$ tùy theo mỗi người. Mắt có thể nhìn càng rõ một vật khi vật đó có hướng nhìn càng gần với trung tâm con mắt, hoặc tròng mắt. Góc nhìn ở chính giữa, rơi vào khoảng từ 0 đến 40$^{\circ}$ hoặc 60$^{\circ}$, gây ảnh hưởng nhiều nhất tới nhận thức của chúng ta. Với góc nhìn này, chúng ta có thể dễ dàng nhìn rõ một vật nào đó mà không phải di chuyển mắt. 

Trở về với bài toán ước lượng góc nhìn của mắt trong hệ thống chứng thực. Khi huấn luyện, chúng ta sẽ cho mô hình học góc nhìn của khuôn mặt và so sánh góc nhìn thực tế của mắt thông qua một hàm loss L1, từ đó độ lỗi góc nhìn của mắt sẽ càng giảm, đơn vị tính chính là độ. Sau đó trong giai đoạn inference, ta thu được vectơ góc nhìn từ ảnh gốc cùng với thông tin tròng mắt.

\subsection{Kiến trúc}
\begin{figure}[H]
    \centering
    \includegraphics[width = 0.5 \textwidth]{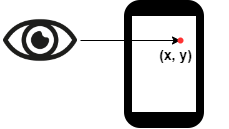}
    \caption{Ví dụ minh họa kết quả của mô hình ước tính góc nhìn của mắt.}
    \label{fig:chart8.1}
\end{figure}

Để huấn luyện mô hình ước lượng góc nhìn của mắt, chúng em sử dụng bộ dữ liệu MPIIFaceGaze thuộc bộ dữ liệu MPIIGaze \cite{DBLP:journals/corr/ZhangSFB15, DBLP:journals/corr/ZhangSFB16, DBLP:journals/corr/abs-1711-09017, DBLP:journals/corr/abs-1901-10906}. Bộ dữ liệu bao gồm 45 nghìn ảnh cho 15 người, mỗi người sẽ có 3 nghìn ảnh, chụp bằng camera trước của laptop với các điều kiện môi trường, ánh sáng khác nhau. Chú thích của bộ dữ liệu được sử dụng bao gồm:
\begin{itemize}
    \item Đường dẫn và tên tệp ảnh.
    \item Vị trí góc nhìn định theo trục pitch và trục yaw, tương ứng với tọa độ trên không gian 2 chiều. Ta chỉ ước tính góc nhìn trên mặt phẳng 2 chiều thay vì 3 chiều để giảm chi phí tính toán, như Hình \ref{fig:chart8.1}.
\end{itemize}

\begin{figure}[H]
    \centering
    \includegraphics[width = \textwidth]{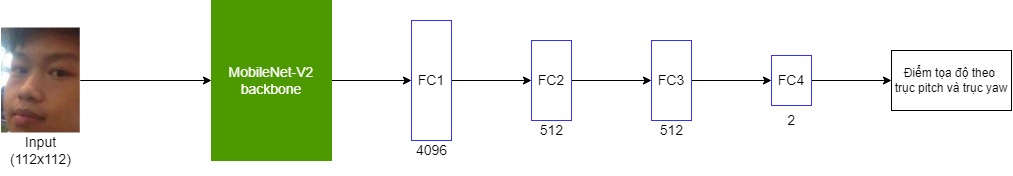}
    \caption{Kiến trúc minh họa cho mô hình ước lượng góc nhìn của mắt.}
    \label{fig:chart8.2}
\end{figure}

Kiến trúc của mô hình khá đơn giản với phần khung MobileNetV2. Tiếp theo sau đó là ba lớp fully connected lần lượt có số nút là 4096, 512, 512 để giảm chiều dữ liệu. Lớp fully connected cuối sẽ đại diện cho hai giá trị theo trục pitch và trục yaw. Mô hình đã được huấn luyện nhưng chúng em gặp trục trặc trong chuyển đổi định dạng (sang TFLite) nên mô hình vẫn chưa được triển khai vào hệ thống ứng dụng. Ngoài ra, nhóm chưa chắc chắn được sự cần thiết và độ hiệu quả của mô hình được đề xuất này cho hệ thống chứng thực. Nguyên nhân chính là mô hình sẽ ảnh hưởng trực tiếp đến kết quả chứng thực của hệ thống ở giai đoạn cuối cùng.

Nếu được áp dụng vào hệ thống, mô hình sẽ được sử dụng ở cuối luồng quy trình chứng thực khuôn mặt, sau mô hình phân loại mắt đóng mở. Nếu mắt được dự đoán là mở thì ta sẽ sử dụng các ảnh khuôn mặt để đưa qua mô hình ước lượng góc nhìn của mắt. Góc nhìn được thể hiện qua hai yếu tố, trục pitch và trục yaw. Ta có thể tìm được góc $\alpha$ hợp bởi tia đọc xuống và đường thẳng nối giữa điểm trung tâm (giữa hai mắt) và giá trị pitch, yaw trên mặt phẳng 2 chiều. Ta chọn một giá trị góc $\delta$ làm ngưỡng, nếu góc $\alpha \leq \delta$ với $\delta$ cho trước (khoảng 30$^{\circ}$ - 45$^{\circ}$), ta đồng ý rằng khuôn mặt nhìn có chủ đích, từ đó chấp nhận danh tính khuôn mặt để thực hiện các tác vụ kế tiếp.

%% file: Appendix/publish.tex
\chapter*{Danh mục công trình của tác giả}
\label{Appendix1}

\begin{enumerate}
\item Bài báo khoa học được nộp tại hội nghị \textbf{ACOMPA'2022}\footnote{http://acomp.tech/}: Large Margin Cotangent Loss for Deep Similarity Learning \cite{duong2022large}.
\item Bài báo khoa học: BiLinear CNNs Model and Test Time Augmentation for Screening Viral and COVID-19 Pneumonia \cite{duong2020bilinear}
\item Bài báo khoa học tại hội nghị \textbf{VNUHCM-US-Conf'20}\footnote{https://conf.hcmus.edu.vn/}: Remove Non-landmark to Improve Landmark Recognition.
\item Bài báo khoa học của nhóm HCMUS tại cuộc thi \textbf{MediaEval 2020}\footnote{https://multimediaeval.github.io/editions/2020}: Image-Text Fusion for Automatic News-Images Re-Matching \cite{Nguyen2020HCMUSAM}.
\item Huy chương Bạc cuộc thi \textbf{Google Landmark Retrieval 2021}\footnote{https://www.kaggle.com/c/landmark-retrieval-2021} 
\item Huy chương Bạc cuộc thi \textbf{Google Landmark Recognition 2021}\footnote{https://www.kaggle.com/competitions/landmark-recognition-2021}. 
\item Huy chương Đồng cuộc thi \textbf{Google Landmark Recognition 2020}\footnote{https://www.kaggle.com/c/landmark-recognition-2020}.

\end{enumerate}

%% file: Appendix/appendix1.tex
\chapter*{Phụ lục}
\label{Appendix1}

\noindent \textbf{Nhận diện khuôn mặt:}
\\
Để so sánh với các hàm suy hao khác, chúng em có huấn luyện thêm mô hình có kiến trúc ResNet50 \cite{He2016DeepRL}, \cite{Han2017DeepPR} trên tập dữ liệu khuôn mặt CASIA \cite{Yi2014LearningFR} sau đó kiểm tra mô hình trên tập dữ liệu khuôn mặt LFW \cite{LFWTech, LFWTechUpdate}.

\begin{table}[htbp]
\caption{Kết quả trên tập LFW}
\label{table01}
\begin{center}
\begin{threeparttable}
    \begin{tabular}{|l|c|}
    \hline
    Hàm suy hao                    & Độ chính xác trên LFW (\%) \\ \hline
    \textbf{LMCot(0.5)\ref{eq:LMCot}} & \textbf{99.58}  \\ \hline
    ArcFace(0.5)                       & 99.53           \\ \hline
    ArcFace(0.4)                       & 99.53           \\ \hline
    CosFace                            & 99.51           \\ \hline
    SphereFace                         & 99.42           \\ \hline
    Softmax                            & 99.08           \\ \hline
    Triplet(0.35)                      & 98.98           \\ \hline
    \end{tabular}
    
    \begin{tablenotes}
      \small
      \item Kiểm tra độ chính xác (\%) của các hàm suy hao khác nhau ([CASIA, ResNet50])
    \end{tablenotes}
\end{threeparttable}
\end{center}
\end{table}

Để so sánh hàm LMCot với các hàm suy hao khác, nhóm có thực hiện huấn luyện và kiểm tra trên cùng một tập dữ liệu và độ đo như trong các bài báo \cite{liu2017sphereface, wang2018cosface, deng2019arcface, boutros2022elasticface} đã dùng để so sánh. Không chỉ nhận diện khuôn mặt, nhóm cũng có khảo sát hàm trên các loại dữ liệu khác như giọng nói, phong cảnh.
\\
\\
\textbf{Nhận diện giọng nói trong cuộc thi VSLP 2021\footnote{https://vlsp.org.vn/vlsp2021}:}
\\
Đây là cuộc thi nằm trong khuôn khổ hội nghị quốc tế lần thứ 8 về xử lý giọng nói và ngôn ngữ Việt. Tập trung vào việc phát triển các mô hình nhận diện giọng nói với dữ liệu hạn chế, bộ huấn luyện được cung cấp có hơn 1000 danh tính người nói. Cuộc thi bao gồm hai nhiệm vụ:
\begin{itemize}
    \item \textbf{SV-T1}: Bộ kiểm tra bao gồm người có trong tập huấn luyện và những người không có trong tập huấn luyện.
    \item \textbf{SV-T2}: Tập kiểm tra chỉ gồm những người không có trong tập huấn luyện.
\end{itemize}

Hiệu suất của các mô hình sẽ được đánh giá bằng Tỷ lệ lỗi ngang nhau (EER) trong đó Tỷ lệ chấp nhận sai (FAR) bằng Tỷ lệ từ chối sai (FRR). Chỉ số đánh giá là độ tương đồng cosin của câu nói đăng ký và câu nói của bài kiểm tra.

Trong cuộc thi này, mỗi người tham gia có 5 bài nộp cho mỗi nhiệm vụ. Chúng em đã thử nghiệm ResNet50 \cite{He2016DeepRL}, EfficientNetB0 \cite{tan2019efficientnet} làm kiến trúc của mô hình. Sau đó thêm phần đệm vào các bản ghi âm để phù hợp với độ dài của bản ghi dài nhất. Chúng em đã sử dụng librosa \cite{mcfee2015librosa} để vẽ đồ thị Mel Spectrograms của những âm thanh này với tham số mặc định, sau đó đưa nó vào mô hình. Dữ liệu đào tạo được chia thành 5 phần, 4 phần cho đào tạo và một lần để xác nhận lần lượt. Với mỗi lượt chúng em chọn trọng số của mô hình có ERR tốt nhất trên tập xác thực. Cuối cùng, chúng em tổng hợp tất cả các mô hình này bằng điểm cách lấy tung bình khoản cách cosin mỗi mẫu trên bộ thử nghiệm. Bằng cách sử dụng phương pháp đơn giản này, chúng em đã đạt được các kết quả sau:
\begin{table}[htbp]
\caption{Kết quả VLSP2021}
\label{table03}
\centering
\begin{threeparttable}
    \begin{tabular}{|l|l|l|}
    \hline
                        & SV-T1          & SV-T2           \\ \hline
    R50 + ArcFace       & 10.520         & 19.850          \\ \hline
    B0 + ArcFace        & 10.365         & 12.665          \\ \hline
    R50 + LMCot \ref{eq:LMCot}        & 10.270         & 12.840          \\ \hline
    \textbf{B0 + LMCot \ref{eq:LMCot}} & \textbf{9.215} & \textbf{12.535} \\ \hline
    Kết hợp tất cả        & 8.805          & 11.605          \\ \hline
    \end{tabular}
    
    \begin{tablenotes}
      \small
      \item Tỷ lệ lỗi ngang nhau (\%) trong cuộc thi nhận diện người nói (VLSP2021). Trong đó R50 là ResNet50 và B0 là  EfficientNetB0. Trong hàng cuối cùng, chúng em lấy trung bình kết quả đầu ra của cả bốn phương pháp trên.
    \end{tablenotes}
\end{threeparttable}
\end{table}

\pagebreak

\noindent \textbf{Dữ liệu phong cảnh:}
\\
Hai cuộc thi Google Landmark Retrieval 2021 và Google Landmark Recognition 2021 là các cuộc thi nằm trong hội thảo về nhận diện (ILR) tại Hội nghị Quốc tế về Thị giác Máy tính 2021 (ICCV 2021). Google Landmark Retrieval 2021 là bài toán truy vấn mục đích truy vấn các ảnh cùng địa điểm với ảnh đầu vào. Google Landmark Recognition 2021 là bài toán nhận diện mục đích là tìm ra nhãn của ảnh đầu vào. Trong hai cuộc thi này, thí sinh được cung cấp tập dữ liệu Google Landmark v2 (gldv2) \cite{weyand2020google}. Những tên sau được sử dụng cho các tập hợp con của gldv2 \cite{henkel2021efficient}:
\begin{itemize}
    \item \textbf{gldv2}: 5 triệu ảnh và 200 nghìn nhãn.
    \item \textbf{gldv2c}: tập dữ liệu sạch (1.6M ảnh của 81 nghìn nhãn).
    \item \textbf{gldv2x}: tập dữ không liệu sạch được lấy theo 81 nghìn nhãn của gldv2c (3.2 triệu ảnh).
\end{itemize}

Trong Google Landmark Retrieval 2021 kết quả được đánh giá theo độ đo mean Average Precision ở mức 100 (mAP@100):
\begin{equation}
\label{eq:metric_glretrieval}
    mAP@100=\frac{1}{Q} \sum_{q=1}^{Q} \frac{1}{\min \left(m_{q}, 100\right)} \sum_{k=1}^{\min \left(n_{q}, 100\right)} P_{q}(k) \operatorname{rel}_{q}(k)
\end{equation}
còn Google Landmark Recognition 2021 được đánh giá bằng độ đo Global Average Precision (GAP) tại $(k = 1)$
\begin{equation}
\label{eq:metric_glrecognition}
    GAP=\frac{1}{M} \sum_{i=1}^{N} P(i) \operatorname{rel}(i)
\end{equation}
trong đó $Q$ là số lượng hình ảnh truy vấn, ${{m}_{q}}$ là số lượng hình ảnh có chứa địa điểm chung với hình ảnh truy vấn $q$. ${{n}_{q}}$ là số dự đoán giải pháp thực hiện cho truy vấn $q$, ${{P}_{q}}\left( k \right)$ là độ chính xác ở hạng $k$ của truy vấn thứ $q$ còn $re{{l}_{q}}\left( k \right)$ là 1 nếu dự đoán thứ k đúng và là 0 nếu ngược lại. $N$ là tổng số dự đoán được giải pháp trả về trên tất cả các truy vấn. Tổng số truy vấn có ít nhất một địa điểm trong tập huấn luyện hiển thị trong đó là $M$, một số truy vấn có thể nằm ngoài các địa điểm trong tập huấn luyện.

\begin{table}[htpb]
\centering
\caption{Kết quả trong cuộc thi Google Landmark Retrieval 2021}
\begin{threeparttable}
\begin{tabular}{|l|l|l|}
\hline
                                                                                                                        & Giải pháp chia sẻ           & Private          \\ \hline
\rowcolor[HTML]{C5E0B3} 
DELG (Giải pháp của ban tổ chức)                                                                                                         & 0.21480          & 0.22152          \\ \hline
4*B0-ArcFace   (Giải pháp chia sẻ)                                                                                                 & 0.24890          & 0.26164          \\ \hline
\rowcolor[HTML]{B4C6E7} 
\textbf{4*B0-LMCot (Đề xuất)}                                                                                              & \textbf{0.26146} & \textbf{0.28338} \\ \hline
4*B7-ArcFace   (Giải pháp chia sẻ)                                                                                                 & 0.31940          & 0.32955          \\ \hline
\rowcolor[HTML]{B4C6E7} 
\textbf{4*B7-LMCot (Đề xuất)}                                                                                              & \textbf{0.32476} & \textbf{0.34691} \\ \hline
4*V2M-ArcFace   (Đề xuất)                                                                                                  & 0.32554          & 0.33973          \\ \hline
\rowcolor[HTML]{B4C6E7} 
\textbf{4*V2M-LMCot (Đề xuất)}                                                                                             & \textbf{0.34933} & \textbf{0.36649} \\ \hline
ViT-ArcFace   (Đề xuất)                                                                                                    & 0.35127          & 0.36569          \\ \hline
\rowcolor[HTML]{B4C6E7} 
\textbf{ViT-LMCot (Đề xuất)}                                                                                               & \textbf{0.35720} & \textbf{0.36723} \\ \hline
\begin{tabular}[c]{@{}l@{}}ViT-ArcFace + ViT-LMCot \\ + B7-ArcFace + B7-LMCot \\ + V2M-LMCot + 2*V2S-LMCot\end{tabular} & 0.37283          & 0.38798          \\ \hline
\end{tabular}
    
    \begin{tablenotes}
      \small
      \item Kết quả trong cuộc thi Google Landmark Retreval 2021. B0 là EfficientNetB0\cite{tan2019efficientnet}, B7 là EfficientNetB7\cite{tan2019efficientnet}, ViT là Vision Transformer \cite{dosovitskiy2020image}, V2S là EfficientNetV2S\cite{tan2021efficientnetv2}, V2M là EfficientNetV2M\cite{tan2021efficientnetv2}. Kết quả đươc đánh giá theo công thức \ref{eq:metric_glretrieval}. Tất cả giải pháp dùng LMCot được xây dựng dựa trên ArcFace \ref{eq:LMCot}.
    \end{tablenotes}
\end{threeparttable}
\end{table}

\begin{table}[htpb]
\centering
\caption{Kết quả trong cuộc thi Google Landmark Recognition 2021}
\begin{threeparttable}
\begin{tabular}{|l|l|l|}
\hline
                                                                                                                          & Giải pháp chia sẻ           & Private          \\ \hline
\rowcolor[HTML]{C5E0B3} 
DELG (Giải pháp của ban tổ chức)                                                                                                           & 0.20245          & 0.19975          \\ \hline
4*B0-ArcFace   (Giải pháp chia sẻ)                                                                                                   & 0.21898          & 0.21468          \\ \hline
\rowcolor[HTML]{B4C6E7} 
\textbf{4*B0-LMCot (Đề xuất)}                                                                                                & \textbf{0.22460} & \textbf{0.22242} \\ \hline
4*B7-ArcFace   (Giải pháp chia sẻ)                                                                                                   & 0.27760          & 0.26292          \\ \hline
\rowcolor[HTML]{B4C6E7} 
\textbf{4*B7-LMCot (Đề xuất)}                                                                                                & \textbf{0.29926} & \textbf{0.29038} \\ \hline
4*V2M-ArcFace   (Đề xuất)                                                                                                    & 0.28593          & 0.29134          \\ \hline
\rowcolor[HTML]{B4C6E7} 
\textbf{4*V2M-LMCot (Đề xuất)}                                                                                               & \textbf{0.29963} & \textbf{0.29353} \\ \hline
ViT-ArcFace   (Đề xuất)                                                                                                      & 0.30309          & 0.30480          \\ \hline
\rowcolor[HTML]{B4C6E7} 
\textbf{ViT-LMCot (Đề xuất)}                                                                                                 & \textbf{0.30225} & \textbf{0.30605} \\ \hline
\begin{tabular}[c]{@{}l@{}}ViT-ArcFace +   ViT-LMCot \\ + B7-ArcFace + B7-LMCot \\ + V2M-LMCot + 2*V2S-LMCot\end{tabular} & 0.30900          & 0.31503          \\ \hline
\end{tabular}

    \begin{tablenotes}
      \small
      \item Kết quả trong cuộc thi Google Landmark Recognition 2021. B0 là EfficientNetB0\cite{tan2019efficientnet}, B7 là EfficientNetB7\cite{tan2019efficientnet}, ViT là Vision Transformer \cite{dosovitskiy2020image}, V2S là EfficientNetV2S\cite{tan2021efficientnetv2}, V2M là EfficientNetV2M\cite{tan2021efficientnetv2}. Kết quả đươc đánh giá theo công thức \ref{eq:metric_glrecognition}. Tất cả giải pháp dùng LMCot được xây dựng dựa trên ArcFace \ref{eq:LMCot}
    \end{tablenotes}
\end{threeparttable}
\end{table}